\documentclass[journal,comsoc]{IEEEtran}
\usepackage[T1]{fontenc}
\usepackage{cite}
\usepackage[pdftex]{graphicx}
\graphicspath{{./images/}}
\usepackage{array}
\usepackage{amsthm}
\usepackage{multicol}
\usepackage{graphicx}
\usepackage{amssymb, amsmath}
\usepackage{xcolor}
\usepackage{cite}
\usepackage{hyperref}
\usepackage{enumerate}
\usepackage{esint}
\usepackage{balance}
\usepackage{textcomp}
\usepackage{tikz}
\usepackage{multirow}

\ifCLASSINFOpdf
	\usepackage[pdftex]{graphicx}
	\graphicspath{{./images/}}
	\DeclareGraphicsExtensions{.pdf,.jpg,.png}
\else
	\usepackage[dvips]{graphicx}
	\graphicspath{{./images/}}
	\DeclareGraphicsExtensions{.eps}
\fi

\ifCLASSOPTIONcompsoc
	\usepackage[caption=false,font=normalsize,labelfont=sf,textfont=sf]{subfig}
\else
	\usepackage[caption=false,font=footnotesize]{subfig}
\fi

\begin{document}
\title{Machine Learning in Nano-Scale \\ Biomedical Engineering}

\author{{Alexandros--Apostolos~A.~Boulogeorgos,~\IEEEmembership{Senior~Member,~IEEE,}
        Stylianos~E.~Trevlakis,~\IEEEmembership{Student~Member,~IEEE,}
        Sotiris~A.~Tegos,~\IEEEmembership{Student~Member,~IEEE,}
        ~Vasilis~K.~Papanikolaou,~\IEEEmembership{Student~Member,~IEEE,} and ~George~K.~Karagiannidis,~\IEEEmembership{Fellow,~IEEE}}
\thanks{The authors are with the Wireless Communications Systems Group (WCSG), Department of Electrical and Computer Engineering, Aristotle University of Thessaloniki, Thessaloniki, 54124 Greece. e-mails: \{trevlakis, geokarag, tegosoti, vpapanikk\} @auth.gr, al.boulogeorgos@ieee.org.}%
\thanks{Alexandros--Apostolos~A.~Boulogeorgos is also with the Department of Digital Systems, University of Piraeus, Piraeus 18534, Greece.}%
\thanks{Manuscript received -, 2020; revised -, 2020.}}

\markboth{IEEE Transactions on Molecular, Biological, and Multi-Scale Communications,~Vol.~-, No.~-, -~2020}%
{Shell \MakeLowercase{\textit{et al.}}: Bare Demo of IEEEtran.cls for IEEE Communications Society Journals}


\maketitle

\begin{abstract}
	Machine learning (ML) empowers biomedical systems with the capability to optimize their performance through modeling of the available data extremely well, without using strong assumptions about the modeled system. Especially in nano-scale biosystems, where the generated data sets are too vast and complex to mentally parse without computational assist, ML is instrumental in analyzing and extracting new insights, accelerating material and structure discoveries and designing experience as well as supporting nano-scale communications and networks. However, despite these efforts, the use of ML in nano-scale biomedical engineering remains still under-explored in certain areas and research challenges are still open in fields such as structure and material design and simulations, communications and signal processing, and bio-medicine applications. In this article, we review the existing research regarding the use of ML in nano-scale biomedical engineering. In more detail, we first identify and discuss the main challenges that can be formulated as ML problems. These challenges are classified in three main categories: structure and material design and simulation, communications and signal processing and biomedicine applications. Next, we discuss the state of the art ML methodologies that are used to countermeasure the aforementioned challenges. For each of the presented methodologies, special emphasis is given to its principles, applications and limitations. Finally, we conclude the article with insightful discussions, that reveal research gaps and highlight possible future research~directions.    
\end{abstract}

\begin{IEEEkeywords}
	Biomedical engineering, Machine learning, Molecular communications, Nano-structure design, Nano-scale networks.
\end{IEEEkeywords}

\IEEEpeerreviewmaketitle

\section*{Nomenclature}
\addcontentsline{toc}{section}{Nomenclature}
\begin{IEEEdescription}
	[\IEEEsetlabelwidth{$V_1,V_2,V_3,V_4$}]
	\item[2D] Two dimensional
	\item[3D] Three dimensional 
	\item[ANI] Accurate neural network engine for molecular energies
	\item[AL] Active Learning
	\item[AdaBoost] Adaptive Boosting
	\item[AEV] Atomic Environments Vector
	\item[ANN] Artificial Neural Network
	\item[ANOVA] Analysis of Variance
	\item[ARES] Autonomous Research System
	\item[Bagging] Bootstrap Aggregating
	\item[BER] Bit Error Rate 
	\item[BPN] Behler-Parrinello Network
	\item[BSS] Blind Source Separation
	\item[CG] Coarse Graining
	\item[CGN] Coarse Graining Network
	\item[CMOS] Complementary\hspace{-0.1cm} Metal-Oxide-Semiconductor
	\item[CNN] Convolution Neural Network
	\item[DCF] Discrete Convolution Filter
	\item[DNN] Deep Neural Network
	\item[D${}^2$NN] Diffractive Deep Neural Network
	\item[DPN] Deep Potential Network
	\item[DT] Decision Table
	\item[DTL] Decision Tree Learning
	\item[DTNB] Decision Table Naive Bayes
	\item[DTNN]	Deep Tensor Neural Network
	\item[EEG]  Electroencephalography
	\item[FS]	Feature Selection
	\item[FSC]	Feedback System Control
	\item[GAN]	Generative Adversarial Network
	\item[GD]	Gradient Descent
	\item[GRNN]	Generalized Regression Neural Network
	\item[ICA]	Independent Component Analysis
	\item[ISI]	Inter-Symbol Interference
	\item[KNN]	k-Nearest Neighbor
	\item[LDA]	Linear Discriminant Analysis
	\item[LR]	Logistic Regression
	\item[LWL]	Local Weighted Learning
	\item[MAN]  Molecular Absorption Noise
	\item[MC]	Molecular Communications
	\item[MIMO] Multiple-Input Multiple-Output 
	\item[ML]	Machine Learning
	\item[MLP]	Multi-layer Perceptron
	\item[ML-SF]	Machine Learning Scoring Function
	\item[MvLR]	Multivariate linear regression
	\item[NBTree]	Naive Bayes Tree
	\item[NN]	Neural Network
	\item[NNP]	Neural Network Potential
	\item[NP]	Nano-Particles
	\item[PAMAM] Polyamidoamine
	\item[PCA]	Principal Component Analysis
	\item[PDF]	Probability Density Function
	\item[PES]	Potential Energy Surface
	\item[PSO] Particle Swarm Optimization
	\item[QM]	Quantum Mechanic
	\item[QP]	Quadratic Programming
	\item[QPOP]	Quadratic Phenotype Optimization Platform
	\item[QSAR]	Quantitative Structure-activity relationships
	\item[RELU]	REctified Linear Unit
	\item[RForest]	Random Forest
	\item[RNAi] Ribonucleic acid interference
	\item[RNN]	Recurrent Neural Network
	\item[SDR]	Standard Deviation Reduction
	\item[SF]	Scoring Functions
	\item[SiC]	Silicon Carbide
	\item[SmF]	Symmetry Function 
	\item[SMO]	Sequential Minimal Optimization
	\item[SOTA] State Of The Art
	\item[SVM]	Support Vector Machine
	\item[TEM]	Transmission Electron Microscope
	\item[THz]	Terahertz 
	\item[ZnO]	Zinc Oxide
\end{IEEEdescription}

\section{Introduction}
In~1959, Richard P. Feynman articulated ``It would be interesting if you could shallow the surgeon. You put the mechanical surgeon inside the blood vessel and it goes into the heart and looks around... other small machines might be permanently incorporated in the body to assist some inadequately-functioning organ.'' More than half a century later, this quote is still state-of-the-art (SOTA). Currently, nanotechnology revisits the conventional therapeutic approaches by producing more than $100$ nano-material based drugs. These have already been approved or they are under clinical trial~\cite{Bobo2016}, while discussing the utilization of nano-scale communication networks for real time monitoring and precision drug delivery~\cite{Akyildiz2015,Farsad2016}. However, these developments come with the need of analyzing vast and complicated, as well as rich in relations, data sets.    

Fortunately, in the last couple of decades, we have witnessed a revolutionary development of new tools from the field of machine learning (ML), which enables the analysis of large data sets through training models. These models can be utilized for observations classification or predictions and have been considered in several engineering fields, including  computer vision, speech and image recognition, natural language processing, etc. This frontier is continuing its expansion into several other scientific domains, such as quantum physics, chemistry and biology, and is expected to make a significant impact on the design of novel nano-materials and structures, nano-scale communication systems and networks, while simultaneously presenting new data-driven biomedicine applications~\cite{Cleophas2015}.

In the field of nano-materials and structure design, experimental and computational simulating methodologies have traditionally been the two fundamental pillars in exploring and discovering properties of novel constructions as well as optimizing their performance~\cite{Molesky2018}. However, these methodologies are constrained by experimental conditions and limitation of the existing theoretical knowledge. Meanwhile, as the chemical complexity of nano-scale heterogeneous structures increases, the two traditional methodologies are rendered incapable of predicting their properties. In this context, the development of data-driven techniques, like ML, becomes very attractive. Similarly, in nano-scale communications and signal processing, the computational resources are limited and the major challenge is the development of low-complexity and accurate system models and data detection techniques, that do not require channel knowledge and equalization, while taking into account the environmental conditions (e.g., specific enzyme composition). To address these challenges the development of novel ML methods is deemed necessary~\cite{Qian2019}. Last but not least, ML can aid in devising novel, more accurate methods for disease detection and therapy development, by enabling genome classification~\cite{Bao2017} and  selection of the optimum combination of drugs~\cite{Duan2012}.         

Motivated from above, the present contribution provides an interdisciplinary review of the existing research from the areas of nano-engineering, biomedical engineering and ML. To the best of the authors knowledge no such review exists in the technical literature, that focuses on the ML-related methodologies that are employed in nano-scale biomedical engineering. In more detail, the contribution of this paper is as follows:
\begin{itemize}
	\item  The main challenges-problems in nano-scale biomedical engineering, which can be tackled with ML techniques, are identified and classified in three main categories, namely: structure and material design and simulations, communications and signal processing, and bio-medicine~applications.
	
	\item SOTA ML methodologies, which are used in the field of nano-scale biomedical engineering, are reviewed, and their architectures are described. For each one of the presented ML methods, we report its principles and building blocks. Finally, their compelling applications in nano-scale biomedicine systems are surveyed for aiding the readers in refining the motivation of ML in these systems, all the way from analyzing and designing new nano-materials and structures to holistic therapy~development.
	
	\item Finally, the advantages and limitations of each ML approach are highlighted, and  future research directions are~provided.
\end{itemize}

\begin{figure}
	\centering
	\includegraphics[width=1\linewidth]{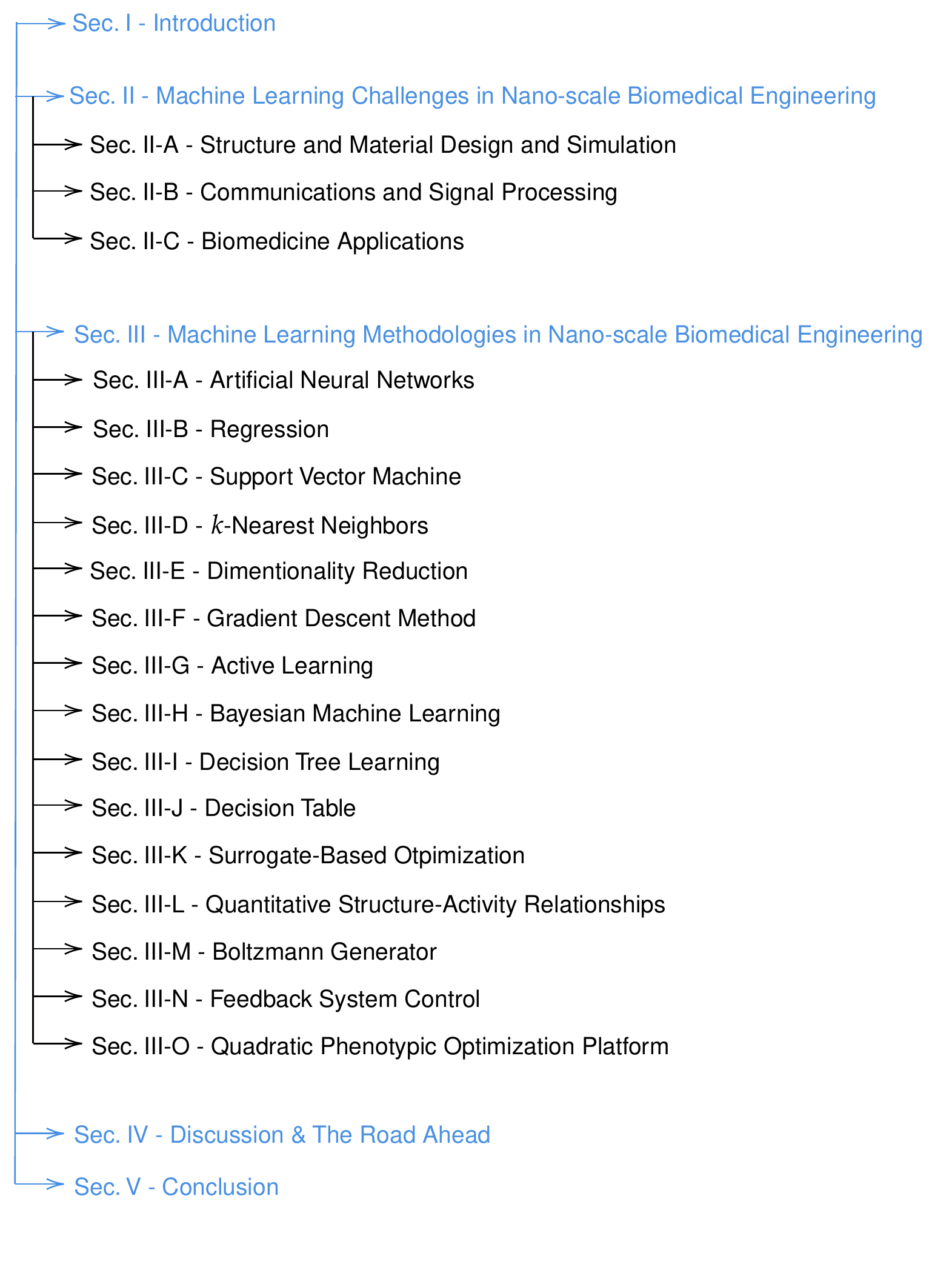}
	\caption{The structure of this treatise.}
	\label{Fig:Strucure}
\end{figure}

The rest of the paper is organized as follows: Section~\ref{S:Problems} identifies the nano-scale biomedical engineering problems that can be solved with ML techniques. Section~\ref{S:Approaches} presents the most common ML approaches related to the field of nano-scale biomedical engineering. Section~\ref{S:Discussion} explains the advantages and limitations of the ML approaches alongside their applications and extracts future directions. Section~\ref{S:Conclusion} concludes this paper and summarizes its contribution. The structure of this treatise is summarized at a glance in Fig.~\ref{Fig:Strucure}.

\section{Machine Learning Challenges in Nano-scale Biomedical Engineering} \label{S:Problems}

\begin{figure}[!t]
	\centering
	\includegraphics[width=1.0\linewidth]{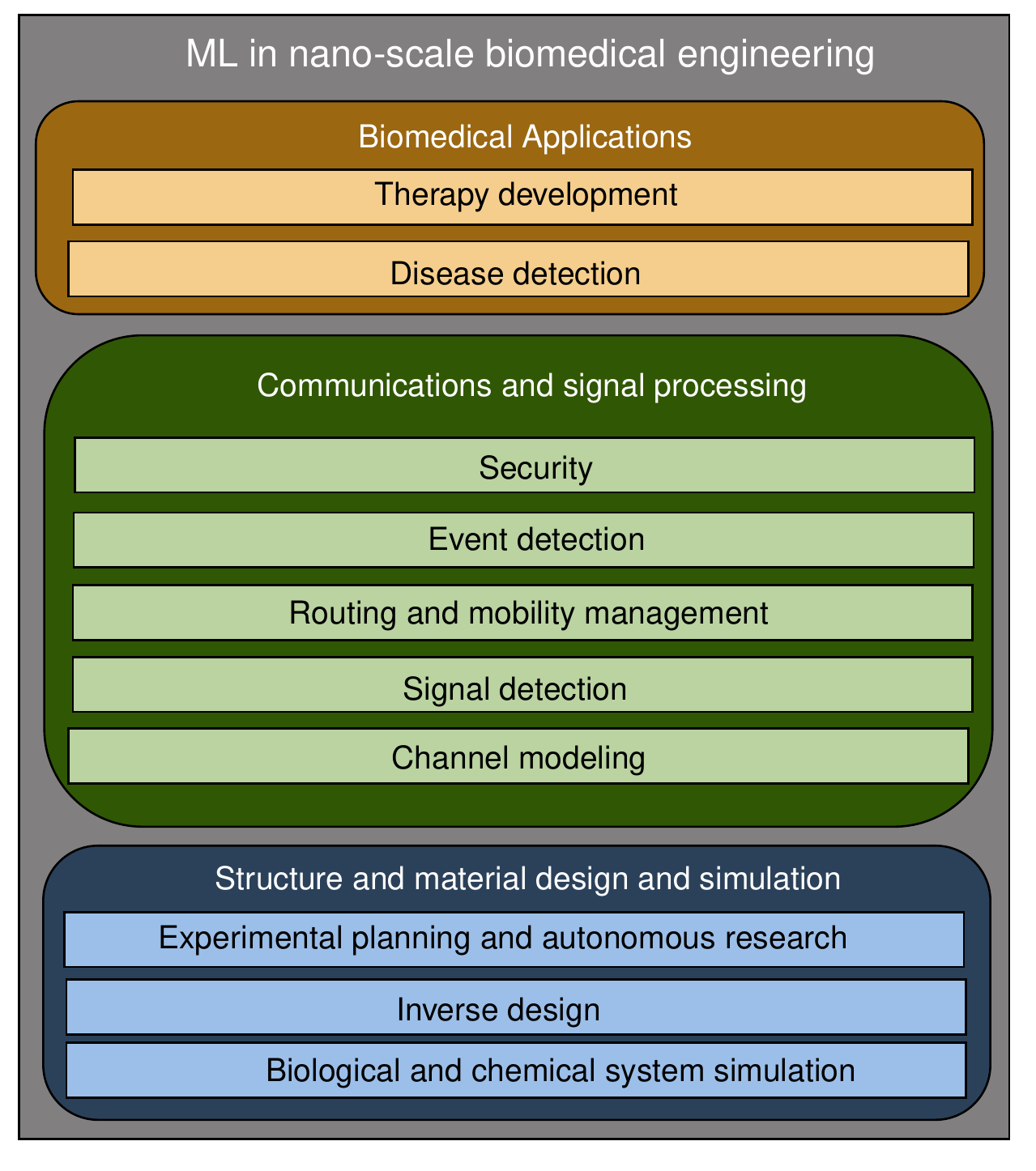}
	\caption{ML challenges in nano-scale biomedical engineering.}
	\label{fig:taxonomy}
\end{figure}

In this section, we report how several of the open challenges in nano-scale biomedical engineering has already been and can be formulated to ML problems. As mentioned in the previous section,  in order to provide a better understanding of the nature of these challenges, we classify them into three categories, i.e. i) structure and material design and simulation, ii) communications and signal processing,  and iii) biomedicine applications. Following this classification, which is illustrated in Fig.~\ref{fig:taxonomy},  the rest of this section is organized as follows: Section~\ref{SS:Structure_and_material_challenges} focuses on presenting the challenges on designing and simulating nano-scale structures, materials and systems, whereas, Section~\ref{SS:Comm_challenges} discusses the necessity of employing ML in nano-scale communications. Similarly, Section~\ref{SS:Apps_challenges} emphasizes in the possible applications of ML in several applications, such as therapy development, drug delivery and data analysis.     

\subsection{Structure and Material Design and Simulation}\label{SS:Structure_and_material_challenges}
One of the fundamental challenges in material science and chemistry is the understanding of the structure properties ~\cite{Butler2018}. The complexity of this problem grows dramatically  in the case of \textit{nanomaterials} because: i) they adopt different properties from their bulk components; and ii) they are usually hetero-structures, consisting of multiple materials.  As a result, the design and optimization of novel structures and materials, by discovering their properties and behavior through simulations and  experiments, lead to multi-parameter and multi-objective problems, which in most cases are extremely difficult or impossible to be solved through conventional approaches; ML can be an efficient alternative choice to this challenge.

\subsubsection{Biological and chemical systems simulation}
In  atomic and molecular systems, there exist complex relationships between the atomistic configuration and the chemical properties, which, in general, cannot be described by explicit forms. In these cases, ML aims to the development of  associate configurations by means of acquiring knowledge from experimental data. Specifically, in order to incorporate quantum effects on molecular dynamics simulations, ML can be employed for the derivation of potential energy surfaces (PESs) from quantum mechanic (QM) evaluations~\cite{Behler2007,Rupp2012,Brockherde2017,Bereau2018,Chmiela2018,Smith2019}. Another use of ML lies in the simulation of molecular dynamic trajectories. For example, in~\cite{John2017,Zhang2018,Wang2019}, the authors formulated ML problems for discovering the optimum reaction coordinates in molecular dynamics, whereas, in~\cite{Stecher2014,Mones2016,Schneider2017,Ribeiro2018,Cendagorta2020}, the problem of estimating free energy surfaces was reported. Furthermore, in~\cite{Warfield2017,Mardt2018,Wu2018,Chen2019}, the ML problem of creating Markov state models, which take into account the molecular kinetics, was investigated. Finally, the ML use in generating samples from equilibrium distributions, that describe molecular systems, was studied in~\cite{Noe2019}.     

\subsubsection{Inverse design} 
The availability of several high-resolution lithographic techniques opened the door to devising complex structures with unprecedented properties. However, the vast choices space, which is created due to the large number of spatial degrees of freedom complemented by the wide choice of materials, makes extremely difficult or even impossible for conventional inverse design methodologies to ensure the existence or uniqueness of acceptable utilizations. To address this challenge, nanoscience community turned their eyes to ML. In more detail, several researchers identified three possible methods, which are based on \textit{artificial neural networks (ANNs)}, \textit{deep neural networks (DNNs)}, and \textit{generative adversarial networks (GANs)}. ANNs follow a trail-and-error approach in order to design multilayer nanoparticles (NP)~\cite{Peurifoy2018}. Meanwhile, DNNs are used in the metasurface design~\cite{Liu2018a}. Finally, GANs can be used to design nanophotonics structures with precise user-define spectral responses~\cite{Liu2018b}.        

\subsubsection{Experiments planning and autonomous research}
ML has been widely employed, in order to efficiently explore the vast parameter space created by different  combinations of nano-materials and experimental conditions and to reduce the number of experiments needed to optimize hetero-structures (see e.g.,~\cite{Cao2018} and references therein). Towards this direction, fully autonomous research can be conducted, in which experiments can be designed based on insights extracted from data processing through ML, without human in the loop ~\cite{King2004}.   

\subsection{Communications and Signal Processing}\label{SS:Comm_challenges}
In biomedical applications, nano-sensors can be utilized for a variety of tasks such as monitoring, detection and treatment~\cite{Boulogeorgos2020,Akyildiz2010}. The size of such nano-sensors ranges between $1-100\text{ }\mathrm{nm}$, which refers to both macro-molecules and bio-cells~\cite{Akyildiz2010}. The proper selection of size and materials is critical for the system performance, while it is constrainted by the target area, their purpose, and safety concerns. Such nano-networks are inspired by living organisms and, when they are injected into the human body, they interact with biological processes in order to collect the necessary information~\cite{Agoulmine2012}. However, they are characterized by limited communication range and processing power, that allow only short-range transmission techniques to be used~\cite{Ali2015}. As a consequence, conventional electromagnetic-based transmission schemes may not be appropriate for communications among molecules~\cite{Hiyama2006,Farsad2016}, since, in molecular communications the information is usually encoded in the number of released particles. The simplest approach for the receiver to demodulate the symbol is to compare the number of received particles with predetermined thresholds.  In the absence of inter-symbol interference (ISI), finding the optimal thresholds is a straightforward process. However, in the presence of ISI the threshold needs to be extracted as a solution of the error probability minimization (or performance maximization) problem~\cite{Jamali2016,Rouzegar2017,Abdallah2020}. The aforementioned approaches require knowledge of the channel model. However, in several practical scenarios, where the molecular communications (MC) system complexity is high, this may not be possible. To countermeasure this issue, ML methods can be employed to accurately model the channel or perform data sequence detection.  

An alternative to MCs that has been used to support nano-networks is communications in the terahertz (THz) band. For these networks, apart from their specifications, an accurate model for the THz communication between nano-sensors is imperative for their simulation and performance assessment. In addition, another problem that is entangled with novel nano-sensor networks is their resilience against attacks, which is of high importance since not only the system reliability is threatened, but also the safety of the patients is at stake. Thus, it is imperative for any possible threats to be recognized and for effective countermeasures to be developed. A solution to the above problems appears to be relatively complex for conventional computational methods. On the other hand, ML can provide the tools to model the space-time trajectories of nano-sensors in the complex environments of the human body as well as to draw strategies that mitigate the security risks of the novel network architectures.   

\subsubsection{Channel modeling}
One of the fundamental problems in MCs is to accurately model the channel in different environments and conditions. Most of the MC models assume that a molecule is removed from the environment after hitting the receiver~\cite{Srinivas2012,Nakano2012,Yilmaz2014,Ahmadzadeh2015,Li2019}; hence, each molecule can contribute to the received signal once. To model this phenomenon, a first-passage process is employed. Another approach was created from the assumption that molecules can pass through the receiver~\cite{Pierobon2010,Kilinc2013,Noel2016,Dinc2018}. In this case, a molecule contributes multiple times to the received signal.   However, neither of the aforementioned approaches are capable of modeling perfectly absorbing receivers, when the transmitters  reflect spherical bodies. Interistingly, such models accommodate practical scenarios where the emitter cells do not have receptors at the emission site and they cannot absorb the emitted molecules. An indicative example lies in hormonal secretion in the synapses and pancreatic $\beta-$cell islets~\cite{Kuran2013}. To fill this gap, ML was employed in~\cite{Yilmaz2017,Lee2017} to model  molecular channels in realistic scenarios, with the aid of ANNs. Similarly, in THz nano-scale networks, where the in-body environment is characterized by high path-loss and molecular absorption noise (MAN), ML methods can be used in order to accurately model MAN. This opens the road to a better understanding of the MAN's nature and the design of new transmission schemes and waveforms.      

\subsubsection{Signal detection}
To avoid channel estimation in MC, Farsal et al. proposed in ~\cite{Farsal2018} a sequence detection scheme, based on \textit{recurrent neural networks (RNNs)}. Compared with previously presented ISI mitigation schemes, ML-based data sequence detection is less complex, since they do not require to perform channel estimation and data equalization. Following a similar approach, in~\cite{Qian2019}, the authors  presented an ANN capable of achieving the same performance as conventional detection techniques, that require perfect knowledge of the channel. 

In THz nano-scale networks, an energy detector is usually used to estimate the received data~\cite{Jornet2014}. In more detail, if the received signal power is below a predefined threshold, the detector decides that the bit $0$ has been sent, otherwise, it decides that $1$ is sent. However, the transmission of $1$ causes a MAN power increase, usually capable of affecting the detection of the next symbols. To counterbalance this, without increasing the symbol duration, a possible approach is to design ML algorithms that are trained to detect the next symbol and take into account the already estimated ones. Another ML challenge in signal detection at THz nano-scale networks, lies with detecting the modulation mode of the transmission signal by a receiver, when no prior synchronization between transmitter and receiver has occurred. The solution to this problem will provide scalability to these networks. Motivated by this, in~\cite{Iqbal2019}, the authors provided a ML algorithm for modulation recognition and classification. 

\subsubsection{Routing and mobility management}
In THz nano-scale networks, the design of routing protocols capable of proactively countermeasuring congestion has been identified as the next step for their utilization~\cite{Zhang2018a}. These protocols need to take into account the extremely constrained computational resources, the stochastic nature of nano-nodes movements as well as the existence of obstacles that may interrupt the line-of-sight transmission. The aforementioned challenges can be faced by employing SOTA ML techniques for analyzing collected data and modeling the nano-sensors' movements, discovering neighbors that can be used as intermediate nodes, identifying possible blockers, and proactively determining the message root from the source to the final destination. In this context, in~\cite{Wang2020}, the authors presented a multi-hop deflection routing algorithm based on reinforcement learning and analyzed its performance in comparison to different neural networks (NNs) and decision tree updating policies.  


\subsubsection{Event detection}
Nano-sensor biomedicine networks can provide continuous monitoring solutions, that can be used as compact, accurate, and portable diagnostic systems. Each nano-sensor obtains a biological signal linked to a specific disease and is used for detecting physiological change or various biological materials~\cite{Nakano2012a}. Successful applications in event detection include monitoring of DNA interactions, antibody, and enzymatic interactions, or cellular communication processes, and are able to detect viruses, asthma attacks and lung cancer~\cite{Nakano2014}. For example, in~\cite{Mannoor2012}, the authors developed a bio-transferrable graphene wireless nano-sensor that is able to sense extremely sensitive chemicals and biological compounds up to single bacterium. Furthermore, in~\cite{Kosaka2014}, a Sandwich Assay was developed that combines mechanical and optoplasmonic transduction in order to detect cancer biomarkers at extremely low concentrations. Also, in~\cite{Mai2017}, a molecular communication-based event detection network was proposed, that is able to cope with scenarios where the molecules propagate according to anomalous diffusion instead of the conventional Brownian motion.

\subsubsection{Security}
Although, the emergence of nano-scale networks based on both electromagnetic and MCs opened opportunities for the development of novel healthcare applications, it also generated new problems concerning the patients' safety. In particular, two types of security risks have been observed, namely blackhole and sentry attacks~\cite{Giaretta2016}. In the former, malicious nano-sensors emit chemicals to attract the legitimate ones and prevent them from searching for their target. On the contrary, in the latter, the malicious nano-sensors repel the legitimate ones for the same reason. Such security risks can be counterbalanced with the use of threshold-based and bayesian ML techniques that have been proven to counter the threats with minimal requirements.  


\subsection{Biomedicine Applications}\label{SS:Apps_challenges}
Timely detection and intervention are tied with successful treatment for many diseases. This is the so-called \textit{proactive treatment} and is one of the main objectives of the next-generation healthcare systems, in order to detect and predict diseases and offer treatment services seamlessly. Data analysis and nanotechnology progress simultaneously toward the realization of these systems.
Recent breakthroughs in nanotechnology-enabled healthcare systems allow for the exploitation of not only the data that already exist in medical databases throughout the world, but also of the data gathered from millions of nano-sensors.

\subsubsection{Disease detection}
One of the most common problems in healthcare systems is genome classification, with cancer detection being the most popular. Various classification algorithms are suitable for tackling this problem, such as Naive Bayes, k-Nearest Neighbors, Decision tree, ANNs and support vector machine (SVM)~\cite{Rizwan2018}. For example, the authors in~\cite{Chen2017}, predicted the risk of cerebral infarction in patients by using demographic and cerebral infarction data. In addition, in~\cite{Bao2017} a unique coarse-to-fine learning method was applied  on genome data to identify gastric cancer. Another example is the research  presented in~\cite{Bardou2018}, where SVM and convolution NNs (CNNs) were used to classify breast cancer subcategory by performing analysis on microscopic images of biopsy.

\subsubsection{Therapy development}
Therapy development and optimization can improve clinical efficacy of treatment for various diseases, without generating unwanted outcomes. Optimization still remains a challenging task, due to its requirement for selecting the right combination of drugs, dose and dosing frequency~\cite{Wilson2020}. For instance, a quadratic phenotype optimization platform (QPOP) was proposed in~\cite{Rashid2018} to determine the optimal combination from 114 drugs to treat bortezomib-resistant multiple myeloma. Since its creation, QPOP has been used to surpass the problems related to drug designing and optimization, as well as drug combinations and dosing strategies. Also, in~\cite{Zarrinpar2016}, the authors presented a platform called CURATE.AI, which was validated clinically and was used to standardize therapy of tuberculosis patients with liver transplant-related immunosuppression. Furthermore, CURATE.AI was used for treatment development and patient guidance that resulted in halted progression of metastatic castration resistant prostate cancer ~\cite{Pantuck2018}.

\section{Machine Learning Methods in Nano-scale Biomedical Engineering}\label{S:Approaches}

\begin{figure}
	\centering
	\includegraphics[width=0.85\linewidth]{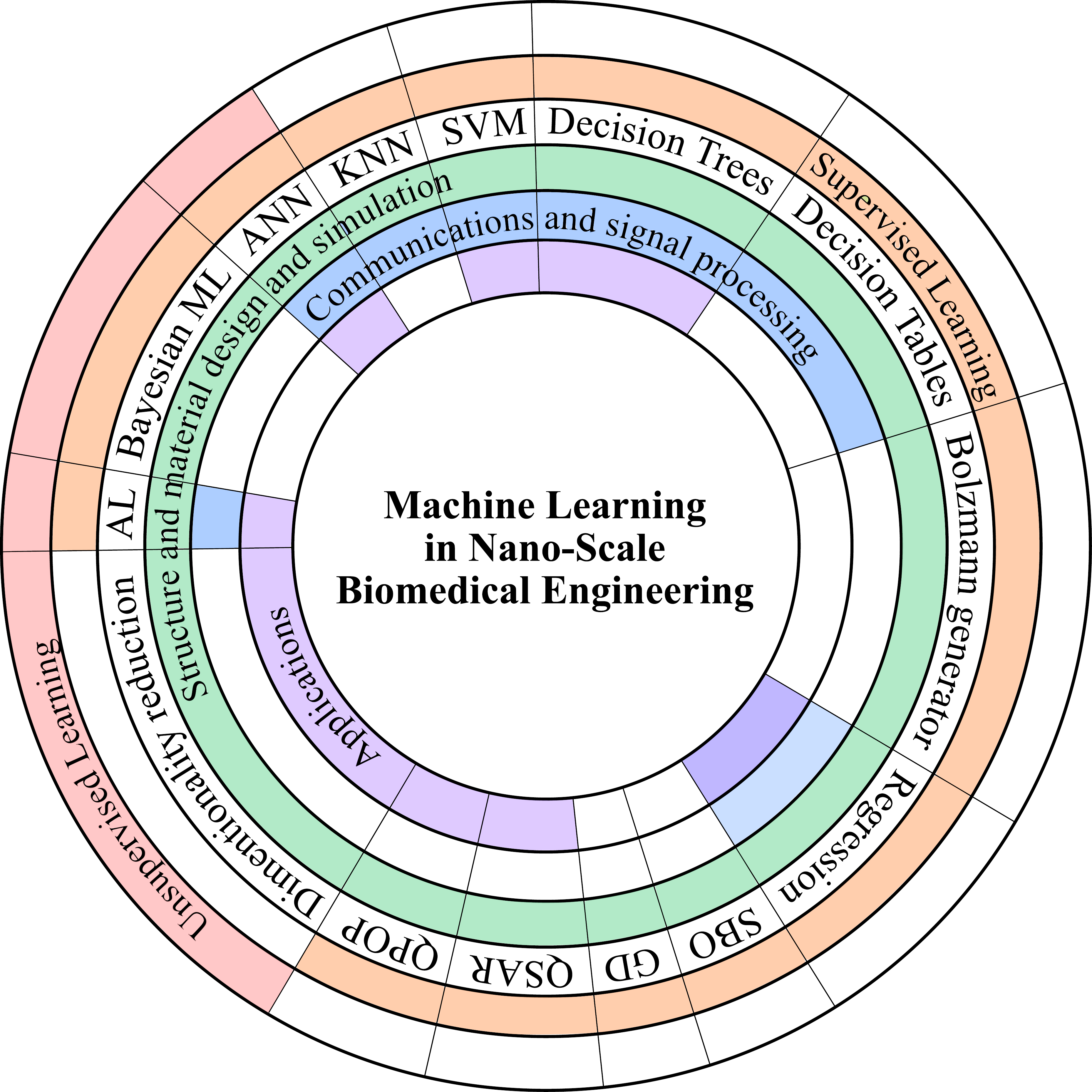}
	\caption{ML methodologies for nano-scale biomedical engineering.}
	\label{Fig:Sup}
\end{figure} 

This section presents the fundamental ML methodologies that are used in nano-scale biomedical engineering. As illustrated in Fig.~\ref{Fig:Sup}, in nano-scale biomedical engineering, depending on how training data are used, we can identify two groups of ML methodologies, namely \textit{supervised}, and~\textit{unsupervised} learning.

\textit{Supervised learning} methodologies require a certain amount of labeled data for training~\cite{Suthaharan2015}. Their objective is to create a function that maps the input data to the output labels relying on the initial training. In more detail, supervised learning return a mapping function $g(x)$ that maximizes the scoring function $f(x_n,y_n)$ for each $n\in[1, N]$, with $x_n$ being the $n-$th sample of the input training data, $y_n$ representing the label of $x_n$, and $N$ being the size of the training set. Of note, in most realistic scenarios, the aforementioned sets are independent and identical distributed.

On the other hand, \textit{unsupervised learning} methodologies aim at exploring the hidden features or structure of data without relying on training sets~\cite{hastie2001the}. Therefore, they have extensively been used for chemical and biological properties discovery in nano-scale structures and materials. The disadvantage of unsupervised learning methodologies lies to the fact that no standard accuracy evaluation method for their output, due to the lack of training data sets. 

\begin{figure}
	\centering
	\includegraphics[width=1\linewidth]{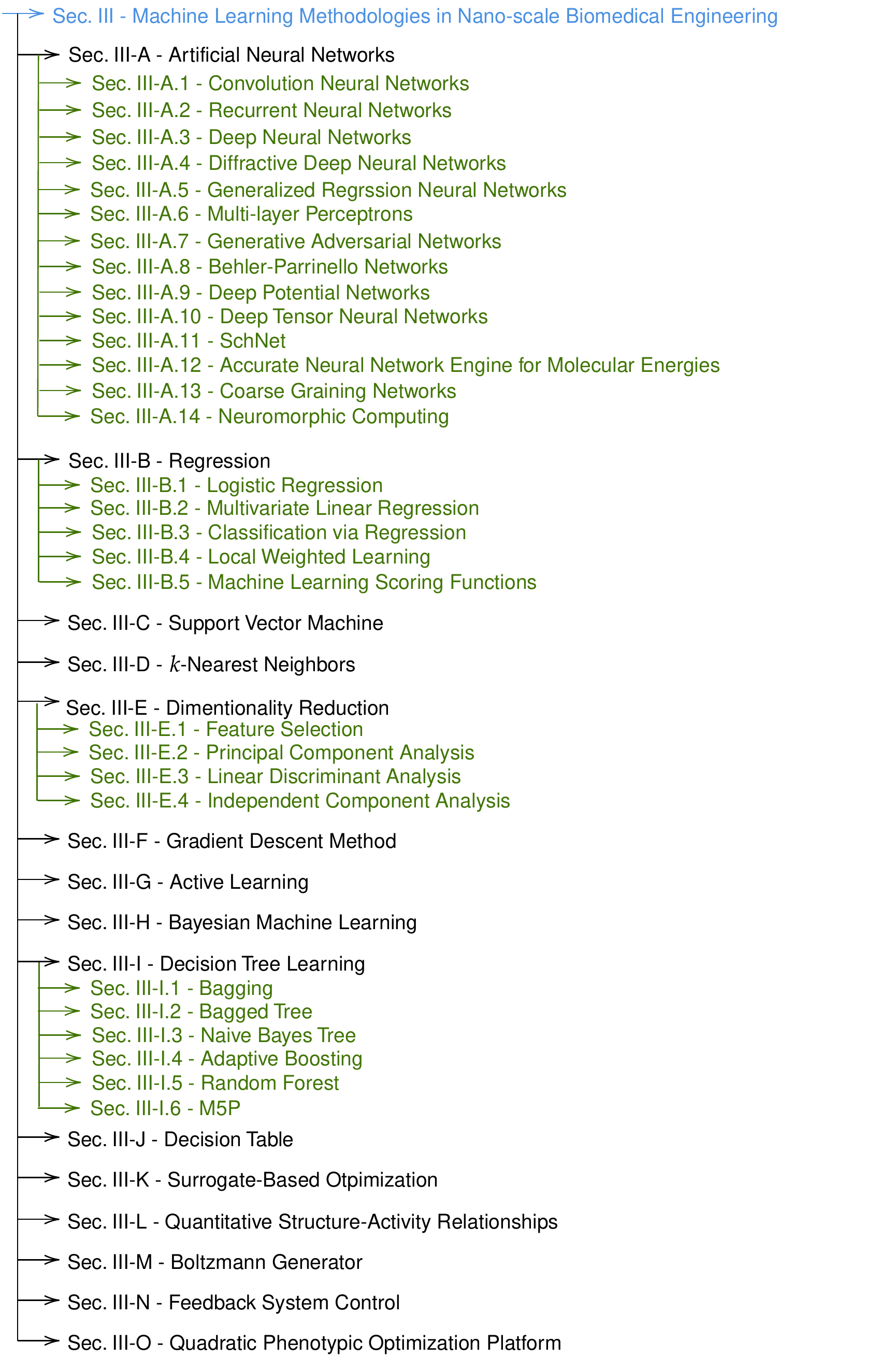}
	\caption{The organization of Section~\ref{S:Approaches}.}
	\label{Fig:OrgIII}
\end{figure}

The rest of this section is  organized as follows: Section~\ref{SS:ANN} provides a survey of the ANNs, which are employed in this field, while Section~\ref{SS:Regression} presents regression methodologies. Meanwhile, the applications, architecture and building blocks of SVMs and $k-$nearest neighbors (KNNs) are respectively described in Sections~\ref{SS:SVM} and~\ref{SS:KNN}, whereas dimentionality reduction methods are given in Section~\ref{SS:DR}. A  brief review of gradient descent (GD) and active learning (AL) methods are respectively delivered in Sections~\ref{SS:GD} and~\ref{SS:AL}. Furthermore, Bayesian ML is discussed in Section~\ref{SS:BML}, whereas decision tree learning (DTL) and decision table (DT) based algorithms are respectively reported in Sections~\ref{SS:DTL} and~\ref{SS:DT}. Section~\ref{SS:SBO} revisits the operating principles of surrogate-based optimization, while Section~\ref{SS:QSAR} describes the use of quantitative structure-activity relationships (QSARs) in ML. Finally, the Boltzmann generator is presented in Section~\ref{SS:BG}, while Sections~\ref{SS:FSC} and~\ref{SS:QPOP} respectively discuss feedback system control (FSC) methods and the quadratic phenotypic optimization platform. The organization of this section is summarized at a glance in Fig.~\ref{Fig:OrgIII}. 

\subsection{Artificial Neural Networks}\label{SS:ANN}
ANNs can be used for both classification and regression. Their operation principle is based on the linear and/or non-linear manipulation of the input-data in several intermediate (hidden) layers. The output of each layer is subjected to by some non-linear functions, namely activation functions. This can be formulated~as
	\begin{align}
		y_k = g\left(v_k + c_k\right),
	\end{align}
where
\begin{align}
	v_k = \sum_{i=1}^{m} w_{ki} x_i,
\end{align}
with $x_i$ and $y_k$ respectively being the input and the output signals of the $k$-th layer, while $w_{ki}$ and $c_k$ respectively standing for the associated weights and bias. Finally, $g(\cdot)$ stands for the activation function. 
This process allows us to model complex relationships of the processed data. 

The reminder of this Section is focused on presenting the ANNs that are commonly used in nano-scale biomedical engineering and is organized as follows: Section~\ref{SSS:CNN} reports the applications of CNNs in this field, presents a typical CNN architecture and discusses its building blocks functionalities.  Similarly, Section~\ref{SSS:RNN} presents the operation of RNNs, while deep NNs (DNNs) are discussed in Section~\ref{SSS:DNN}. Diffractive  DNNs (D$^2$NN) and generalized regression NNs (GRNNs) are respectively described in Section~\ref{SSS:DDNN} and~\ref{SSS:GRNN}, while Sections~\ref{SSS:MLP} and~\ref{SSS:GAN} respectively revisit the multi-layer perceptrons (MLPs) and GANs. Moreover, the applications, architecture and limitations of  Behler-Parrinello networks (BPNs) are reported in Section~\ref{SSS:BPN}, whereas, Sections~\ref{SSS:DPN},~\ref{SSS:DTNN}, and~\ref{SSS:SchNet} respectively present the ones of deep potential networks (DPNs), deep tensor NNs (DTNNs), and SchNets. Likewise, the usability and building blocks of accurate NN engine for molecular energies, or as is widely-known ANI, are provided in Section~\ref{SSS:ANI}. Finally, comprehensive descriptions of coarse graining networks (CGNs) and neurophormic computing are respectively given in Sections~\ref{SSS:CGN} and~\ref{SSS:NC}. Table~\ref{T:ANN} summarizes some of the typical applications of ANNs in nano-scale biomedical engineering. 

\begin{table*}[]
	\centering
	\caption{ANN applications in nano-scale biomedical engineering. }
	\label{T:ML_approaches_ML_challenges}
	\begin{tabular}{|c||l|c|r|}
		\hline
		\textbf{Paper} & \textbf{Application} & \textbf{Method} & \textbf{Description} \\
			\hline 	\hline
		\cite{Wang2019} & Chemical properties discovery & CGN & Prediction of the rototranslationally invariant energy in QM \\
		\hline
		\cite{Liu2018b} & Nano-material inverse design & GAN & Metasurfaces inverse design\\
			\hline
		\cite{Lee2017} & Channel modelling & DNN & MIMO channel modeling in MC\\
		\hline
		\cite{Farsal2018} &  Sequence detection & RNN & Data sequence detection in MC\\
		\hline
		\cite{Shibata2018} & Image analysis & CNN & Skyrmions analysis in labeled Lorentz TEM images \\
		\hline
		\cite{Carrasquilla2017} & Image analysis & CNN & Matter phases identification \\
		\hline
		\cite{Rashidi2018} & ARES & CNN & State-of-the-tip identification in  tunneling microscopy scanning \\
		\hline
		\cite{Hegde2020} & Image analysis & RNN & Nano-structure design \\
		\hline
		\cite{Farsad2017} &  Sequence detection & RNN & Data sequence detection in MC\\
		\hline
		\cite{Lin2018} & Feature detection and object classification & D$^2$NN & Classification of images and creation of imaging lens at THz spectrum \\
		\hline	
		 \cite{Park2014} & Data analysis & GRNN, MLP, BPN & Characterization of psychological wellness from survey results \\
		 \hline
		 \cite{Seela2018} & Nano-structure properties discovery & GRNN & Study of the impact of ZnO NPs suspensions in diesel and Mahua \\ & & &  biodiesel 
		 blended fuel\\
		 \hline
		 \cite{Zarei2019} & Nano-structure properties discovery & GRNN & Prediction of the pool boiling heat transfer coefficient of refrigerant-based \\ & & & nano-fluids \\
		 \hline
		 \cite{Uddin2012} & Nano-structure analysis & MLP & Analysis of the crystalline structure of magnesium oxide films grown over \\ & & & 6H SiC substrates \\
		 \hline
		 \cite{So2019} & Nano-structure design & GAN & Nano-photonic structure design\\
		 \hline
		 \cite{Han2018} & Chemical properties discovery & BPN & Energy surfaces prediction from QM data\\
		 \hline
		 \cite{Nagai2020} & Complex structure simulation & BPN &  Self-learning Monte Carlo creation for many-body interactions\\
		 \hline
		 \cite{Liu2020a} & Complex structure simulation & BPN & Atomic energy prediction \\
		 \hline
		 \cite{Zhang2018b} & Chemical properties discovery &  DPN & PES prediction that use atomic configuration directly at the   input data
		 \\
		 \hline
		 \cite{Schuett2017} & Molecules and nano-material properties & DTNN & General QM molecular potential modeling \\ & discovery &  &  \\
		 \hline
		 \cite{KristofT.Schuett2017} & Chemical properties discovery & SchNet  &   PES prediction that takes into account rototranslationally invariant\\  &  &  & inter-atomic distances \\
		 \hline
		 \cite{Gao2020} & Chemical properties modeling & ANI & Prediction of molecules energies in complex nano-structures \\
		 \hline
		 \cite{Davtyan2016} & Chemical properties modeling & CGN & Theormodynamics prediction in chemical systems\\
		 \hline
		 \cite{Nueske2019} & Chemical properties modeling & CGN & Theormodynamics prediction in chemical systems\\
		 \hline
	\end{tabular} 
\label{T:ANN} 
\end{table*}  

\subsubsection{Convolution Neural Networks}\label{SSS:CNN}
\textit{CNNs} have been extensively used for analyzing images with some degrees of spatial correlation~\cite{Chua1993,EgmontPetersen2002,Tajbakhsh2016,Fang2019}.  The aim of CNNs is to extract fundamental local correlations within the data, and thus, they are suitable for identifying image features that depend on these correlations. In this sense, in~\cite{Shibata2018}, the author employed CNNs to analyze skyrmions in labeled Lorentz transmission electron microscope (TEM) images, while, in~\cite{Carrasquilla2017}, CNNs were used to identify matter phases from data extracted via Monte Carlo~simulations.  Another application of CNNs in nano-scale biomedical systems lies in the utilization of autonomous research systems (ARES)~\cite{Rashidi2018}. Specifically, in~\cite{Rashidi2018}, the authors  presented a learning method that determines the state-of-the-tip in scanning tunneling microscopy. 

\begin{figure}
	
	\centering
	\tikzset{every picture/.style={line width=0.75pt}} 
	
	\begin{tikzpicture}[x=0.55pt,y=0.55pt,yscale=-1,xscale=1]
	
	\draw  [draw opacity=0][fill={rgb, 255:red, 208; green, 2; blue, 27 }  ,fill opacity=1 ] (79,34) -- (239.5,34) -- (239.5,155) -- (79,155) -- cycle ; \draw   (79,34) -- (79,155)(99,34) -- (99,155)(119,34) -- (119,155)(139,34) -- (139,155)(159,34) -- (159,155)(179,34) -- (179,155)(199,34) -- (199,155)(219,34) -- (219,155)(239,34) -- (239,155) ; \draw   (79,34) -- (239.5,34)(79,54) -- (239.5,54)(79,74) -- (239.5,74)(79,94) -- (239.5,94)(79,114) -- (239.5,114)(79,134) -- (239.5,134)(79,154) -- (239.5,154) ; \draw    ;
	\draw  [draw opacity=0][fill={rgb, 255:red, 126; green, 211; blue, 33 }  ,fill opacity=1 ] (99,54) -- (259.5,54) -- (259.5,175) -- (99,175) -- cycle ; \draw   (99,54) -- (99,175)(119,54) -- (119,175)(139,54) -- (139,175)(159,54) -- (159,175)(179,54) -- (179,175)(199,54) -- (199,175)(219,54) -- (219,175)(239,54) -- (239,175)(259,54) -- (259,175) ; \draw   (99,54) -- (259.5,54)(99,74) -- (259.5,74)(99,94) -- (259.5,94)(99,114) -- (259.5,114)(99,134) -- (259.5,134)(99,154) -- (259.5,154)(99,174) -- (259.5,174) ; \draw    ;
	\draw  [draw opacity=0][fill={rgb, 255:red, 74; green, 144; blue, 226 }  ,fill opacity=1 ] (119,74) -- (279.5,74) -- (279.5,195) -- (119,195) -- cycle ; \draw   (119,74) -- (119,195)(139,74) -- (139,195)(159,74) -- (159,195)(179,74) -- (179,195)(199,74) -- (199,195)(219,74) -- (219,195)(239,74) -- (239,195)(259,74) -- (259,195)(279,74) -- (279,195) ; \draw   (119,74) -- (279.5,74)(119,94) -- (279.5,94)(119,114) -- (279.5,114)(119,134) -- (279.5,134)(119,154) -- (279.5,154)(119,174) -- (279.5,174)(119,194) -- (279.5,194) ; \draw    ;
	\draw  [dash pattern={on 3.38pt off 3.27pt}][line width=3]  (173.25,129) -- (224.75,129) -- (224.75,179) -- (173.25,179) -- cycle ;
	\draw    (173.25,179) -- (201.5,253) ;
	\draw    (224.75,179) -- (215.5,253) ;
	\draw    (173.25,129) -- (201.5,239) ;
	\draw    (222.25,131) -- (215.5,239) ;
	\draw  [dash pattern={on 3.38pt off 3.27pt}][line width=3]  (201.5,239) -- (215.5,239) -- (215.5,253) -- (201.5,253) -- cycle ;
	\draw  [fill={rgb, 255:red, 207; green, 206; blue, 205 }  ,fill opacity=1 ] (64.5,260.7) -- (88.2,237) -- (286,237) -- (286,292.3) -- (262.3,316) -- (64.5,316) -- cycle ; \draw   (286,237) -- (262.3,260.7) -- (64.5,260.7) ; \draw   (262.3,260.7) -- (262.3,316) ;
	\draw  [dash pattern={on 3.38pt off 3.27pt}][line width=3]  (200.25,275) -- (221.5,275) -- (221.5,298) -- (200.25,298) -- cycle ;
	\draw  [fill={rgb, 255:red, 184; green, 233; blue, 134 }  ,fill opacity=1 ] (170.5,365.7) -- (194.2,342) -- (266,342) -- (266,397.3) -- (242.3,421) -- (170.5,421) -- cycle ; \draw   (266,342) -- (242.3,365.7) -- (170.5,365.7) ; \draw   (242.3,365.7) -- (242.3,421) ;
	\draw  [dash pattern={on 3.38pt off 3.27pt}][line width=3]  (204.25,377) -- (225.5,377) -- (225.5,400) -- (204.25,400) -- cycle ;
	\draw    (200.25,275) -- (214.88,388.5) ;
	\draw    (221.5,275) -- (214.88,388.5) ;
	\draw    (200.25,298) -- (214.88,388.5) ;
	\draw    (221.5,298) -- (214.88,388.5) ;
	\draw  [fill={rgb, 255:red, 207; green, 206; blue, 205 }  ,fill opacity=1 ] (167.5,466.7) -- (191.2,443) -- (253,443) -- (253,498.3) -- (229.3,522) -- (167.5,522) -- cycle ; \draw   (253,443) -- (229.3,466.7) -- (167.5,466.7) ; \draw   (229.3,466.7) -- (229.3,522) ;
	\draw  [dash pattern={on 3.38pt off 3.27pt}][line width=3]  (197.25,483) -- (218.5,483) -- (218.5,506) -- (197.25,506) -- cycle ;
	\draw    (204.25,377) -- (207.88,494.5) ;
	\draw    (225.5,377) -- (207.88,494.5) ;
	\draw    (204.25,400) -- (207.88,494.5) ;
	\draw    (225.5,400) -- (207.88,494.5) ;
	\draw  [fill={rgb, 255:red, 207; green, 206; blue, 205 }  ,fill opacity=1 ] (163.5,597.7) -- (187.2,574) -- (249,574) -- (249,629.3) -- (225.3,653) -- (163.5,653) -- cycle ; \draw   (249,574) -- (225.3,597.7) -- (163.5,597.7) ; \draw   (225.3,597.7) -- (225.3,653) ;
	\draw  [draw opacity=0][fill={rgb, 255:red, 211; green, 165; blue, 125 }  ,fill opacity=1 ] (90,691) -- (211.5,691) -- (211.5,712) -- (90,712) -- cycle ; \draw   (90,691) -- (90,712)(110,691) -- (110,712)(130,691) -- (130,712)(150,691) -- (150,712)(170,691) -- (170,712)(190,691) -- (190,712)(210,691) -- (210,712) ; \draw   (90,691) -- (211.5,691)(90,711) -- (211.5,711) ; \draw    ;
	\draw  [draw opacity=0][fill={rgb, 255:red, 211; green, 165; blue, 125 }  ,fill opacity=1 ] (244.5,689) -- (265.5,689) -- (265.5,710) -- (244.5,710) -- cycle ; \draw   (244.5,689) -- (244.5,710)(264.5,689) -- (264.5,710) ; \draw   (244.5,689) -- (265.5,689)(244.5,709) -- (265.5,709) ; \draw    ;
	\draw  [dash pattern={on 3.38pt off 3.27pt}][line width=3]  (171.25,632.82) -- (182.5,632.82) -- (182.5,645) -- (171.25,645) -- cycle ;
	\draw    (176.88,638.91) -- (98.5,696) ;
	\draw  [fill={rgb, 255:red, 214; green, 213; blue, 213 }  ,fill opacity=1 ] (108,762.75) .. controls (108,756.26) and (113.26,751) .. (119.75,751) .. controls (126.24,751) and (131.5,756.26) .. (131.5,762.75) .. controls (131.5,769.24) and (126.24,774.5) .. (119.75,774.5) .. controls (113.26,774.5) and (108,769.24) .. (108,762.75) -- cycle ;
	\draw  [fill={rgb, 255:red, 214; green, 213; blue, 213 }  ,fill opacity=1 ] (148,762.75) .. controls (148,756.26) and (153.26,751) .. (159.75,751) .. controls (166.24,751) and (171.5,756.26) .. (171.5,762.75) .. controls (171.5,769.24) and (166.24,774.5) .. (159.75,774.5) .. controls (153.26,774.5) and (148,769.24) .. (148,762.75) -- cycle ;
	\draw  [fill={rgb, 255:red, 214; green, 213; blue, 213 }  ,fill opacity=1 ] (188,762.75) .. controls (188,756.26) and (193.26,751) .. (199.75,751) .. controls (206.24,751) and (211.5,756.26) .. (211.5,762.75) .. controls (211.5,769.24) and (206.24,774.5) .. (199.75,774.5) .. controls (193.26,774.5) and (188,769.24) .. (188,762.75) -- cycle ;
	\draw  [fill={rgb, 255:red, 214; green, 213; blue, 213 }  ,fill opacity=1 ] (228,762.75) .. controls (228,756.26) and (233.26,751) .. (239.75,751) .. controls (246.24,751) and (251.5,756.26) .. (251.5,762.75) .. controls (251.5,769.24) and (246.24,774.5) .. (239.75,774.5) .. controls (233.26,774.5) and (228,769.24) .. (228,762.75) -- cycle ;
	\draw  [fill={rgb, 255:red, 214; green, 213; blue, 213 }  ,fill opacity=1 ] (108,812.75) .. controls (108,806.26) and (113.26,801) .. (119.75,801) .. controls (126.24,801) and (131.5,806.26) .. (131.5,812.75) .. controls (131.5,819.24) and (126.24,824.5) .. (119.75,824.5) .. controls (113.26,824.5) and (108,819.24) .. (108,812.75) -- cycle ;
	\draw  [fill={rgb, 255:red, 214; green, 213; blue, 213 }  ,fill opacity=1 ] (148,812.75) .. controls (148,806.26) and (153.26,801) .. (159.75,801) .. controls (166.24,801) and (171.5,806.26) .. (171.5,812.75) .. controls (171.5,819.24) and (166.24,824.5) .. (159.75,824.5) .. controls (153.26,824.5) and (148,819.24) .. (148,812.75) -- cycle ;
	\draw  [fill={rgb, 255:red, 214; green, 213; blue, 213 }  ,fill opacity=1 ] (188,812.75) .. controls (188,806.26) and (193.26,801) .. (199.75,801) .. controls (206.24,801) and (211.5,806.26) .. (211.5,812.75) .. controls (211.5,819.24) and (206.24,824.5) .. (199.75,824.5) .. controls (193.26,824.5) and (188,819.24) .. (188,812.75) -- cycle ;
	\draw  [fill={rgb, 255:red, 214; green, 213; blue, 213 }  ,fill opacity=1 ] (228,812.75) .. controls (228,806.26) and (233.26,801) .. (239.75,801) .. controls (246.24,801) and (251.5,806.26) .. (251.5,812.75) .. controls (251.5,819.24) and (246.24,824.5) .. (239.75,824.5) .. controls (233.26,824.5) and (228,819.24) .. (228,812.75) -- cycle ;
	\draw    (100.5,706) -- (119.75,751) ;
	\draw    (100.5,706) -- (159.75,751) ;
	\draw    (100.5,706) -- (199.75,751) ;
	\draw    (100.5,706) -- (239.75,751) ;
	\draw    (122.5,706) -- (119.75,751) ;
	\draw    (122.5,706) -- (159.75,751) ;
	\draw    (122.5,706) -- (199.75,751) ;
	\draw    (122.5,706) -- (239.75,751) ;
	\draw    (142.5,707) -- (119.75,751) ;
	\draw    (142.5,707) -- (159.75,751) ;
	\draw    (142.5,707) -- (199.75,751) ;
	\draw    (142.5,707) -- (239.75,751) ;
	\draw    (159.5,705) -- (119.75,751) ;
	\draw    (159.5,705) -- (159.75,751) ;
	\draw    (159.5,705) -- (199.75,751) ;
	\draw    (159.5,705) -- (239.75,751) ;
	\draw    (180.5,706) -- (119.75,751) ;
	\draw    (180.5,706) -- (159.75,751) ;
	\draw    (180.5,706) -- (199.75,751) ;
	\draw    (180.5,706) -- (239.75,751) ;
	\draw    (201.5,707) -- (119.75,751) ;
	\draw    (201.5,707) -- (159.75,751) ;
	\draw    (201.5,707) -- (199.75,751) ;
	\draw    (201.5,707) -- (239.75,751) ;
	\draw    (251.5,703) -- (119.75,751) ;
	\draw    (251.5,703) -- (159.75,751) ;
	\draw    (251.5,703) -- (199.75,751) ;
	\draw    (251.5,703) -- (239.75,751) ;
	\draw    (119.75,774.5) -- (119.75,801) ;
	\draw    (119.75,774.5) -- (159.75,801) ;
	\draw    (119.75,774.5) -- (199.75,801) ;
	\draw    (119.75,774.5) -- (239.75,801) ;
	\draw    (159.75,774.5) -- (119.75,801) ;
	\draw    (159.75,774.5) -- (159.75,801) ;
	\draw    (159.75,774.5) -- (199.75,801) ;
	\draw    (159.75,774.5) -- (239.75,801) ;
	\draw  [fill={rgb, 255:red, 245; green, 166; blue, 35 }  ,fill opacity=1 ] (139,860.75) .. controls (139,854.26) and (144.26,849) .. (150.75,849) .. controls (157.24,849) and (162.5,854.26) .. (162.5,860.75) .. controls (162.5,867.24) and (157.24,872.5) .. (150.75,872.5) .. controls (144.26,872.5) and (139,867.24) .. (139,860.75) -- cycle ;
	\draw  [fill={rgb, 255:red, 245; green, 166; blue, 35 }  ,fill opacity=1 ] (199,860.75) .. controls (199,854.26) and (204.26,849) .. (210.75,849) .. controls (217.24,849) and (222.5,854.26) .. (222.5,860.75) .. controls (222.5,867.24) and (217.24,872.5) .. (210.75,872.5) .. controls (204.26,872.5) and (199,867.24) .. (199,860.75) -- cycle ;
	\draw    (199.75,774.5) -- (199.75,801) ;
	\draw    (199.75,774.5) -- (119.75,801) ;
	\draw    (199.75,774.5) -- (159.75,801) ;
	\draw    (199.75,774.5) -- (239.75,801) ;
	\draw    (239.75,774.5) -- (199.75,801) ;
	\draw    (239.75,774.5) -- (239.75,801) ;
	\draw    (239.75,774.5) -- (159.75,801) ;
	\draw    (239.75,774.5) -- (119.75,801) ;
	\draw    (150.75,849) -- (119.75,824.5) ;
	\draw    (150.75,849) -- (159.75,824.5) ;
	\draw    (150.75,849) -- (199.75,824.5) ;
	\draw    (150.75,849) -- (239.75,824.5) ;
	\draw    (210.75,849) -- (119.75,824.5) ;
	\draw    (210.75,849) -- (159.75,824.5) ;
	\draw    (210.75,849) -- (199.75,824.5) ;
	\draw    (210.75,849) -- (239.75,824.5) ;
	
	\draw (247,26) node [anchor=north west][inner sep=0.75pt]   [align=left] {Red};
	\draw (266,48) node [anchor=north west][inner sep=0.75pt]   [align=left] {Green};
	\draw (285,71) node [anchor=north west][inner sep=0.75pt]   [align=left] {Blue};
	\draw (43,213) node [anchor=north west][inner sep=0.75pt]   [align=left] {Convolution + RELU};
	\draw (101,382) node [anchor=north west][inner sep=0.75pt]   [align=left] {Pooling};
	\draw (17,482) node [anchor=north west][inner sep=0.75pt]   [align=left] {Convolution + RELU};
	\draw (199,537.4) node [anchor=north west][inner sep=0.75pt]    {$\vdots $};
	\draw (213,692.4) node [anchor=north west][inner sep=0.75pt]    {$\cdots $};
	\draw (24,659) node [anchor=north west][inner sep=0.75pt]   [align=left] {Flattening};
	\draw (42,693) node [anchor=north west][inner sep=0.75pt]   [align=left] {Inputs};
	\draw (70,853) node [anchor=north west][inner sep=0.75pt]   [align=left] {Outputs};
	\draw (165,852.4) node [anchor=north west][inner sep=0.75pt]    {$\cdots $};
	\end{tikzpicture}
	\caption{CNN architecture.}	
	\label{Fig:CNN}
\end{figure}
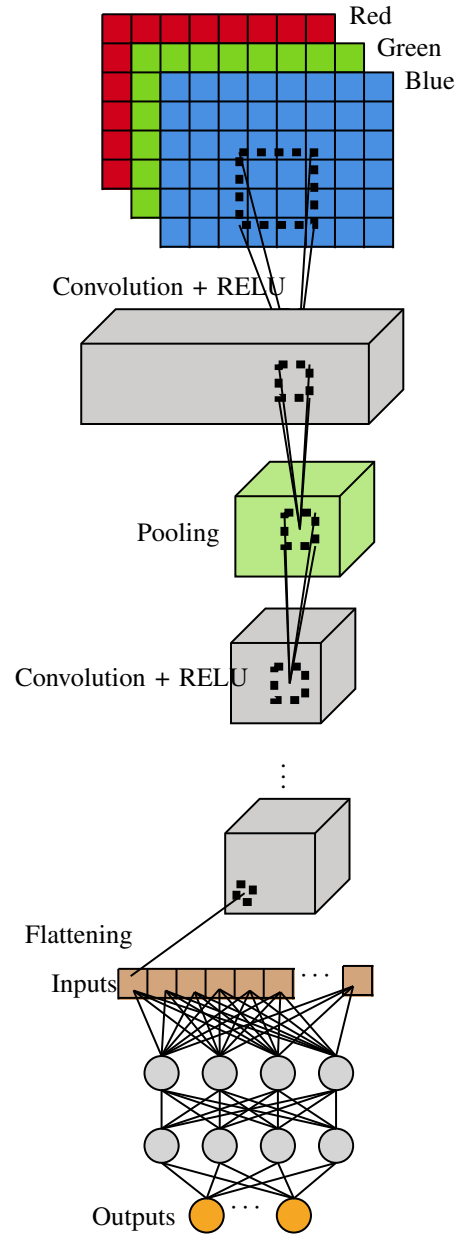

Figure~\ref{Fig:CNN} depicts a typical CNN architecture, which  mimics the neurons' connectivity patterns in the human brain. It consists of neurons, which are arranged in a three dimensional (3D) space, i.e., width, height, and depth. Each neuron receives several inputs and performs an element-wise multiplication, which is usually followed by a non-linear operation. Note that, in most cases, CNN architectures are not fully-connected. This means that the neurons in a layer will only be connected to a small region of the previous layer. Each layer of a CNN transforms its input to a 3D output of neuron activations. In more detail, it consists of the following layers:
\begin{itemize}
	\item \textit{Input}: This layer represents the input image into the CNN. Input layer holds the raw pixels of the image in the three color channels, namely red, green, and blue.
	\item \textit{Convolution}:  layers are the pillars of CNN. They contain the weights that are used to extract the distinguished features of the images. As illustrated in~Fig.~\ref{Fig:CNN}, they  evaluate the output of neurons, which are connected to local regions in the input. 
	\item \textit{Rectified linear unit (RELU)}: applies an element-wise activation function, such as thresholding at zero. This allows the generation of non-linear decision boundaries.
	\item \textit{Pooling}: conducts downsampling  along the spatial dimensions. 
	\item \textit{Flattening}: reorganizes the values of the 3D matrix into a vector. 
	\item \textit{Hidden layers}: returns the classification scores.   
\end{itemize}	  

\subsubsection{Recurrent Neural Networks}\label{SSS:RNN}
Most ML networks rely to the assumption of independence among the training and test data. Thus, after processing each data point, the entire state of the network is lost. Apparently, this is not a problem,  if the data points are independently generated. However, if they are in time or space related, the aforementioned assumption becomes unacceptable. Moreover, conventional networks usually rely on data points, which can be organized in vectors of fixed length. However, in practice, there exist several problems, which require  modeling data with temporal or sequential structure and varying length inputs and outputs.

In order to overcome the aforementioned limitations, \textit{RNNs} have been proposed in~\cite{Lipton2015}. RNNs are connectionist models capable of selectively passing information across sequence steps, while processing sequential data. From the nano-scale applications point of view, RNNs have been used for nano-structure design and data sequence detection in MCs. Specifically, in~\cite{Hegde2020}, Hedge described the role that RNNs are expected to play in the design of nano-structures, while, in~\cite{Farsad2017} and in~\cite{Farsal2018}, the authors employed a RNN in order to train a maximum likelihood detector in MCs~systems.  

\begin{figure*}
	\centering
	\includegraphics[width=0.75\linewidth]{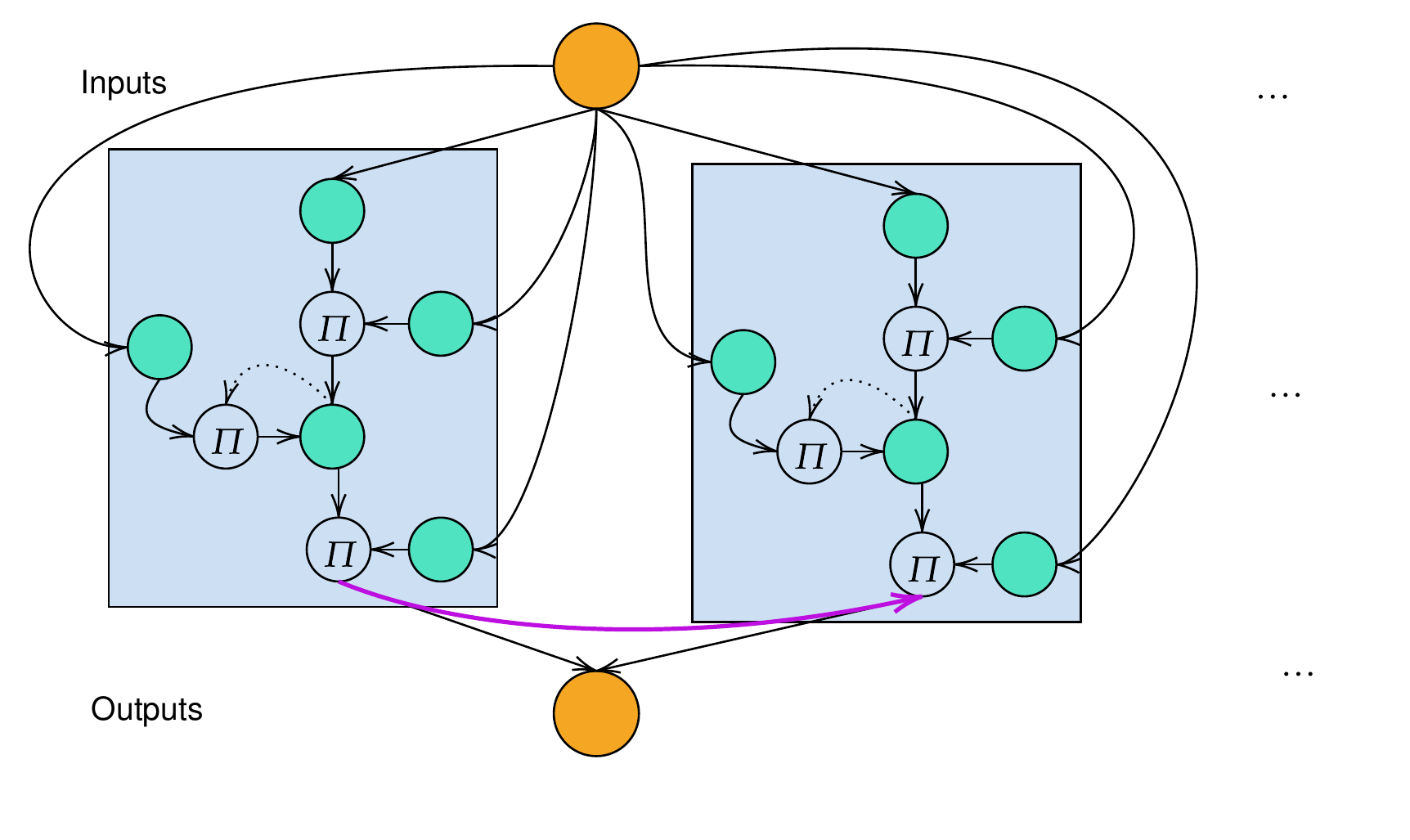}
	\caption{An RNN with a hidden layer consisting of two memory cells.}
	\label{Fig:RNN}
	\hrulefill
\end{figure*}

Figure~\ref{Fig:RNN} depicts the most successful RNN architecture,  introduced by Hochreiter and Schmidhuber~\cite{Hochreiter1997}. From this figure, it is evident that the only difference between RNN and CNN is the fact that the hidden layers of the latter are replaced with memory cells with self-connected recurrent fix-weighted edges. The memory cells store the internal state of the RNN and allow processing sequences of inputs of varying length. Likewise, the recurrent edges guarantee that the gradient can pass across several steps without vanishing. The weights change during training in a slowing rate in order to create a long-term memory. Finally, RNNs support short-term memory through ephemeral activations, which pass from each node to successive nodes. This allows RNNs to exploit the dynamic temporal information hidden in time sequences. 

\subsubsection{Deep Neural Networks}\label{SSS:DNN}

Deep learning was suggested in~\cite{Farsal2018} as an efficient method to detect the information at the receiver in MCs. Specifically, based on the similarities between speech recognition and molecular channels, techniques from DL can be utilized to train a detection algorithm from samples of transmitted and received signals. In the same work, it was proposed that  well-known NNs such as an RNN, can train a detector even if the underlying system model is not known. Furthermore, a real-time NN-based sequence detector was proposed, and it was shown that the suggested DL-based algorithms could eliminate the need for  instantaneous channel state information~estimation.

In another research work, \cite{Lee2017}, a NN-based modeling of  the molecular multiple-input multiple-output (MIMO) channel, was presented.  This is a remarkable contribution, since the proposed model can be used to investigate the possibility of increasing the low rates in MCs. Specifically, in this paper a $2\times 2$ molecular MIMO channel was modeled  through two ML-based techniques and the developed model was used to evaluate the bit error rate (BER). 

\subsubsection{Diffractive Deep Neural Networks}\label{SSS:DDNN}
In \cite{Lin2018}, a diffractive deep NN ($\text{D}^2$NN) framework was proposed. The $\text{D}^2$NN is an all-optical deep learning framework, where multiple layers of diffractive surfaces physically form the NN. These layers collaborate to optically perform an arbitrary function, which can be learned statistically by the network. The learning part is performed through a computer, whereas the prediction of the physical network follows an all-optical approach.

Several transmissive and/or reflective layers create the $\text{D}^2$NN. More specifically, each point on a specific layer can either transmit or reflect the incoming wave.  To this end, an artificial neuron  is formed, which is connected to other neurons of the following layers through optical diffraction.  Following Huygens' principle, each point on a specific layer acts as a secondary source of a wave, whose amplitude and phase are expressed as the product of the complex valued transmission or reflection coefficient and the input wave at that point. Consequently, the input interference pattern, due to the earlier layers and the local transmission/reflection coefficient at a specific point, modulate the amplitude and phase a secondary wave, through which an artificial neuron in the D$^2$NN is connected to the neurons of the following layer. 
 The transmission/reflection coefficient of each neuron can be considered as a multiplicative bias term, which is an repetitively adjusted parameter during the training process of the diffractive network, using an error back-propagation method.
Generally, the amplitude and the phase of each neuron can be a learnable parameter, providing a complex-valued modulation at each layer and, thus, enhancing the inference performance of the network.

\subsubsection{Generalized Regression Neural Networks}\label{SSS:GRNN}
\textit{GRNN} belongs to the instance-based learning methods and it is a variation of radial basis NNs \cite{Specht1991}. Instance-based learning methods, that construct hypotheses directly from the training instances, have tractable computational cost in general, compared to the not instance-based like MLP with backpropagation. GRNN consists of an input layer, a pattern layer, and the output layer and can be expressed~as
\begin{equation}
\hat{y}(x) = \hat{f}(x) = \frac{\sum_{k=1}^N y_k K(x,x_k)}{\sum_{k=1}^N K(x,x_k)},
\end{equation}
where $y(x)$ is the prediction value of the $N+1$-th input $x$, $y_k$ is the activation of $k$-th neuron of the pattern layer and $K(x,x_k)$ is the radial basis function kernel, which is a Gaussian kernel given by
\begin{equation}
K(x,x_k) = e^{-d_k / 2\sigma^2}, \quad d_k = (x-x_k)^T(x-x_k),
\end{equation}
where $d$ is the Euclidean distance and $\sigma$ is a smoothing parameter.
Due to the presence of $K(x,x_k)$, the value $y_k$ of training data instances that are closer to $x$, according to the $\sigma$ parameter, has more significant contribution to the predicted~value. 

GRNN is used in~\cite{Park2014} in order to characterize psychological wellness from survey results that measure stress, depression, anger, and~fatigue. Moreover, it was employed in~\cite{Seela2018} for investigating the effect of zinc oxide (ZnO) NPs suspensions in diesel and Mahua biodiesel blended fuel on single cylinder diesel engine performance characteristics. Finally, in~\cite{Zarei2019}, it was employed for predict the pool boiling heat transfer coefficient  of refrigerant-based nano-fluids.  

\subsubsection{Multi-layer Perceptrons}\label{SSS:MLP}
\textit{MLP} is a type of feed-forward ANN that consists of at least three layers of nodes:  input layer,  output layer, and one or more hidden layers \cite{Hastie2009}. Apart from the input nodes $a_n^{(0)}$, each node is a neuron that takes as input a weighted sum of the node values as well as a bias of the previous layer and gives an output depending on a usually sigmoid activation function, $\sigma(\dot)$. Therefore, the input of the $k$-th neuron in the $L$-th layer can be expressed~as
\begin{equation}
z_k^{(L)} = w_{k,0}a_0^{(L-1)} + \dots w_{k,n}a_n^{(L-1)} + b_k,
\end{equation}
where $w_i$ are the weights associated to each node at the previous layer and $b_i^{(L)}$ is the bias at the $i$-th node of the $L$-th hidden layer. The activation of that neuron then can be written as 
\begin{equation}
a_i^{(L)}=\sigma(z_i^{(L)}).
\end{equation}

The number of nodes in the input layer is equivalent to the number of input features, whereas the number of output neurons corresponds to the output features. A cost function $C$, which is usually the sum squared errors between prediction and target, is calculated and it is fed in a backward fashion in order to update the weights in each neuron via a GD algorithm, and thus, to minimize the cost function. This learning method of updating the weights in such manner is called back-propagation \cite{Wythoff1993}. More specifically, the degree of error in an output node $j$ for the $n$-th training example is $e_j(n) = y_j(n) - \hat{y}_j(n)$, where $y$ is the target value and $\hat{y}$ is the predicted value by the perceptron. The error, for example $n,$ over all output nodes can be obtained~as
\begin{equation}
C(n) = \sum_j e_j^2(n).
\end{equation}
GD dictates a change in weights proportional to the negative gradient of the cost function, $-\nabla\mathbf{C}(\mathbf{w})$. However, this method with the entirety of training data can be computationally expensive, so methods like stochastic GD for every step can increase~efficiency. 

MLP was used in~\cite{Park2014} in order to characterize psychological wellness from survey results that measure stress, depression, anger, and~fatigue. Likewise, in~\cite{Uddin2012}, MLP found an application in analyzing the crystalline structure of magnesium oxide films grown over 6H silicon carbide (SiC) substrates. 

\subsubsection{Generative Adversarial Networks}\label{SSS:GAN}
A \textit{GAN}~\cite{Brown2019} is an unsupervised learning strategy, which was introduced in \cite{Goodfellow2014}. A GAN consists of two networks, a \textit{generator} that estimates the distributions of the parameters and a \textit{discriminator} that evaluates each estimation by comparing it to the available unlabeled data. This strategy can exploit specific training algorithms for different models and optimization algorithms. Specifically, a MLP can be utilized in a twofold way, i.e., the generative model generates samples by passing random noise through it, while it is also used as the discriminative model.  Both networks can be trained using only the highly successful backpropagation and dropout algorithms, while approximate prediction or Markov chains are not necessary.

The generator's distribution $p_g$ over data $x$ can be learned by defining a prior on input noise variables $p_z(z)$ and representing a mapping to data space as $G(z; \theta_g)$, where $G$ is a differentiable function which corresponds to a MLP with parameter $\theta_g$.  A second MLP $D(x; \theta_d)$ with parameter $\theta_d$ and a single scalar number as output, denotes the probability that $x$ is derived from the data rather than $p_g$. The $D$ is trained in order to maximize the probability that the training examples and samples from $G$ are labeled correctly, while $G$ is simultaneously trained to minimize the term $\log(1 - D(G(z)))$.
More specifically, a two-player min-max game is performed with value function $V (G ; D)$ as follows:
\begin{equation}
\begin{split}
\underset{G}{\mathrm{min}} \ \underset{D}{\mathrm{max}} \ V(D,G) & = \mathbb{E}_{x \sim p_g(x)}[\log D(x)] \\
& + \mathbb{E}_{x \sim p_z(z)}[\log(1 - D(G(z)))].
\end{split}
\end{equation}
In practice, the game must be performed by using an iterative numerical approach. Optimizing $D$ in the inner loop of training is computationally prohibitive and on finite data sets would result in over-fitting. A better solution is to alternate between $k$ steps of optimizing $D$ and one step of optimizing $G$. To this end, $D$ is maintained near its optimal solution, while $G$ is modified slowly enough.
In nano-scale biomedical engineering GAN has found application in nanophotonics structure design~\cite{So2019} as well as in metasurface inverse design~\cite{Liu2018b}.

\subsubsection{Behler-Parrinello Networks}\label{SSS:BPN}
\textit{BPNs} are traditionally used in molecular sciences in order to learn and predict the energy surfaces from QM data, by combining all the relevant physical symmetries and properties as well as sharing parameters between atoms~\cite{Han2018}. Another use of BPN lies in the self-learning Monte Carlo simulation development for many-body interactions~\cite{Nagai2020}. Specifically, in~\cite{Nagai2020}, the authors employed BPNs to make trainable effective Hamiltonians that were used to extract the potential-energy surfaces in interacting many particle systems. Finally, in~\cite{Liu2020a}, BPNs were used to predict the atomic energy for different elements. 

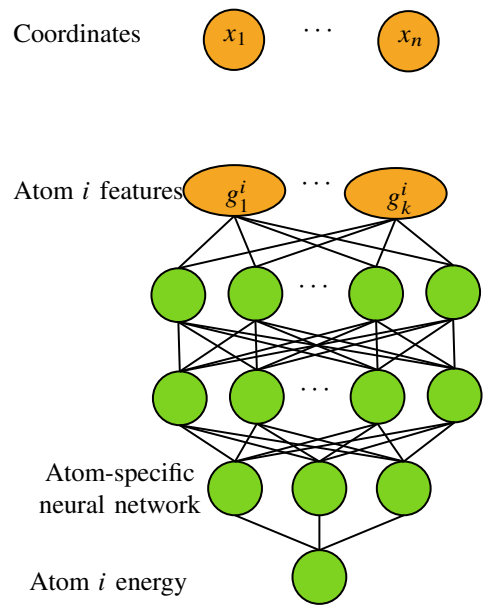
\begin{figure}
	\centering
	
	
	\tikzset {_z5tl059z2/.code = {\pgfsetadditionalshadetransform{ \pgftransformshift{\pgfpoint{0 bp } { 0 bp }  }  \pgftransformrotate{-90 }  \pgftransformscale{2 }  }}}
	\pgfdeclarehorizontalshading{_pkflmvacl}{150bp}{rgb(0bp)=(1,1,1);
		rgb(41.607142857142854bp)=(1,1,1);
		rgb(62.5bp)=(0,0,0);
		rgb(100bp)=(0,0,0)}
	\tikzset{_bxlr7gp08/.code = {\pgfsetadditionalshadetransform{\pgftransformshift{\pgfpoint{0 bp } { 0 bp }  }  \pgftransformrotate{-90 }  \pgftransformscale{2 } }}}
	\pgfdeclarehorizontalshading{_um57r55e1} {150bp} {color(0bp)=(transparent!0);
		color(41.607142857142854bp)=(transparent!0);
		color(62.5bp)=(transparent!10);
		color(100bp)=(transparent!10) } 
	\pgfdeclarefading{_qavp90prk}{\tikz \fill[shading=_um57r55e1,_bxlr7gp08] (0,0) rectangle (50bp,50bp); } 
	\tikzset{every picture/.style={line width=0.75pt}} 
	
	\begin{tikzpicture}[x=0.55pt,y=0.55pt,yscale=-1,xscale=1]
	
	\draw  [fill={rgb, 255:red, 245; green, 166; blue, 35 }  ,fill opacity=1 ] (156,37.75) .. controls (156,26.29) and (165.29,17) .. (176.75,17) .. controls (188.21,17) and (197.5,26.29) .. (197.5,37.75) .. controls (197.5,49.21) and (188.21,58.5) .. (176.75,58.5) .. controls (165.29,58.5) and (156,49.21) .. (156,37.75) -- cycle ;
	\draw  [draw opacity=0][shading=_pkflmvacl,_z5tl059z2,path fading= _qavp90prk ,fading transform={xshift=2}][line width=0.75]  (209.5,91.4) -- (222.88,91.4) -- (222.88,59) -- (249.63,59) -- (249.63,91.4) -- (263,91.4) -- (236.25,113) -- cycle ;
	\draw  [fill={rgb, 255:red, 245; green, 166; blue, 35 }  ,fill opacity=1 ] (276,38.75) .. controls (276,27.29) and (285.29,18) .. (296.75,18) .. controls (308.21,18) and (317.5,27.29) .. (317.5,38.75) .. controls (317.5,50.21) and (308.21,59.5) .. (296.75,59.5) .. controls (285.29,59.5) and (276,50.21) .. (276,38.75) -- cycle ;
	\draw  [fill={rgb, 255:red, 245; green, 166; blue, 35 }  ,fill opacity=1 ] (142,141.5) .. controls (142,131.84) and (157.67,124) .. (177,124) .. controls (196.33,124) and (212,131.84) .. (212,141.5) .. controls (212,151.16) and (196.33,159) .. (177,159) .. controls (157.67,159) and (142,151.16) .. (142,141.5) -- cycle ;
	\draw  [fill={rgb, 255:red, 245; green, 166; blue, 35 }  ,fill opacity=1 ] (253,143.5) .. controls (253,133.84) and (268.67,126) .. (288,126) .. controls (307.33,126) and (323,133.84) .. (323,143.5) .. controls (323,153.16) and (307.33,161) .. (288,161) .. controls (268.67,161) and (253,153.16) .. (253,143.5) -- cycle ;
	\draw    (177,159) -- (138.5,195) ;
	\draw  [fill={rgb, 255:red, 126; green, 211; blue, 33 }  ,fill opacity=1 ] (120,213.5) .. controls (120,203.28) and (128.28,195) .. (138.5,195) .. controls (148.72,195) and (157,203.28) .. (157,213.5) .. controls (157,223.72) and (148.72,232) .. (138.5,232) .. controls (128.28,232) and (120,223.72) .. (120,213.5) -- cycle ;
	\draw  [fill={rgb, 255:red, 126; green, 211; blue, 33 }  ,fill opacity=1 ] (173,212.5) .. controls (173,202.28) and (181.28,194) .. (191.5,194) .. controls (201.72,194) and (210,202.28) .. (210,212.5) .. controls (210,222.72) and (201.72,231) .. (191.5,231) .. controls (181.28,231) and (173,222.72) .. (173,212.5) -- cycle ;
	\draw  [fill={rgb, 255:red, 126; green, 211; blue, 33 }  ,fill opacity=1 ] (256,212.5) .. controls (256,202.28) and (264.28,194) .. (274.5,194) .. controls (284.72,194) and (293,202.28) .. (293,212.5) .. controls (293,222.72) and (284.72,231) .. (274.5,231) .. controls (264.28,231) and (256,222.72) .. (256,212.5) -- cycle ;
	\draw  [fill={rgb, 255:red, 126; green, 211; blue, 33 }  ,fill opacity=1 ] (309,211.5) .. controls (309,201.28) and (317.28,193) .. (327.5,193) .. controls (337.72,193) and (346,201.28) .. (346,211.5) .. controls (346,221.72) and (337.72,230) .. (327.5,230) .. controls (317.28,230) and (309,221.72) .. (309,211.5) -- cycle ;
	\draw    (177,159) -- (191.5,194) ;
	\draw    (177,159) -- (274.5,194) ;
	\draw    (177,159) -- (327.5,193) ;
	\draw    (138.5,195) -- (288,161) ;
	\draw    (288,161) -- (191.5,194) ;
	\draw    (288,161) -- (274.5,194) ;
	\draw    (288,161) -- (327.5,193) ;
	\draw  [fill={rgb, 255:red, 126; green, 211; blue, 33 }  ,fill opacity=1 ] (121,284.5) .. controls (121,274.28) and (129.28,266) .. (139.5,266) .. controls (149.72,266) and (158,274.28) .. (158,284.5) .. controls (158,294.72) and (149.72,303) .. (139.5,303) .. controls (129.28,303) and (121,294.72) .. (121,284.5) -- cycle ;
	\draw  [fill={rgb, 255:red, 126; green, 211; blue, 33 }  ,fill opacity=1 ] (174,283.5) .. controls (174,273.28) and (182.28,265) .. (192.5,265) .. controls (202.72,265) and (211,273.28) .. (211,283.5) .. controls (211,293.72) and (202.72,302) .. (192.5,302) .. controls (182.28,302) and (174,293.72) .. (174,283.5) -- cycle ;
	\draw  [fill={rgb, 255:red, 126; green, 211; blue, 33 }  ,fill opacity=1 ] (257,283.5) .. controls (257,273.28) and (265.28,265) .. (275.5,265) .. controls (285.72,265) and (294,273.28) .. (294,283.5) .. controls (294,293.72) and (285.72,302) .. (275.5,302) .. controls (265.28,302) and (257,293.72) .. (257,283.5) -- cycle ;
	\draw  [fill={rgb, 255:red, 126; green, 211; blue, 33 }  ,fill opacity=1 ] (310,282.5) .. controls (310,272.28) and (318.28,264) .. (328.5,264) .. controls (338.72,264) and (347,272.28) .. (347,282.5) .. controls (347,292.72) and (338.72,301) .. (328.5,301) .. controls (318.28,301) and (310,292.72) .. (310,282.5) -- cycle ;
	\draw    (138.5,232) -- (139.5,266) ;
	\draw    (138.5,232) -- (192.5,265) ;
	\draw    (191.5,231) -- (139.5,266) ;
	\draw    (138.5,232) -- (275.5,265) ;
	\draw    (138.5,232) -- (328.5,264) ;
	\draw    (191.5,231) -- (192.5,265) ;
	\draw    (191.5,231) -- (275.5,265) ;
	\draw    (274.5,231) -- (275.5,265) ;
	\draw    (274.5,231) -- (328.5,264) ;
	\draw    (191.5,231) -- (328.5,264) ;
	\draw    (274.5,231) -- (192.5,265) ;
	\draw    (274.5,231) -- (139.5,266) ;
	\draw    (327.5,230) -- (139.5,266) ;
	\draw    (327.5,230) -- (192.5,265) ;
	\draw    (327.5,230) -- (275.5,265) ;
	\draw    (327.5,230) -- (328.5,264) ;
	\draw  [fill={rgb, 255:red, 126; green, 211; blue, 33 }  ,fill opacity=1 ] (159,346.5) .. controls (159,336.28) and (167.28,328) .. (177.5,328) .. controls (187.72,328) and (196,336.28) .. (196,346.5) .. controls (196,356.72) and (187.72,365) .. (177.5,365) .. controls (167.28,365) and (159,356.72) .. (159,346.5) -- cycle ;
	\draw  [fill={rgb, 255:red, 126; green, 211; blue, 33 }  ,fill opacity=1 ] (217,346.5) .. controls (217,336.28) and (225.28,328) .. (235.5,328) .. controls (245.72,328) and (254,336.28) .. (254,346.5) .. controls (254,356.72) and (245.72,365) .. (235.5,365) .. controls (225.28,365) and (217,356.72) .. (217,346.5) -- cycle ;
	\draw  [fill={rgb, 255:red, 126; green, 211; blue, 33 }  ,fill opacity=1 ] (275,346.5) .. controls (275,336.28) and (283.28,328) .. (293.5,328) .. controls (303.72,328) and (312,336.28) .. (312,346.5) .. controls (312,356.72) and (303.72,365) .. (293.5,365) .. controls (283.28,365) and (275,356.72) .. (275,346.5) -- cycle ;
	\draw    (139.5,303) -- (177.5,328) ;
	\draw    (139.5,303) -- (235.5,328) ;
	\draw    (139.5,303) -- (293.5,328) ;
	\draw    (192.5,302) -- (177.5,328) ;
	\draw    (192.5,302) -- (235.5,328) ;
	\draw    (192.5,302) -- (293.5,328) ;
	\draw    (275.5,302) -- (177.5,328) ;
	\draw    (275.5,302) -- (235.5,328) ;
	\draw    (275.5,302) -- (293.5,328) ;
	\draw    (328.5,301) -- (177.5,328) ;
	\draw    (328.5,301) -- (235.5,328) ;
	\draw    (328.5,301) -- (293.5,328) ;
	\draw  [fill={rgb, 255:red, 126; green, 211; blue, 33 }  ,fill opacity=1 ] (217,407.5) .. controls (217,397.28) and (225.28,389) .. (235.5,389) .. controls (245.72,389) and (254,397.28) .. (254,407.5) .. controls (254,417.72) and (245.72,426) .. (235.5,426) .. controls (225.28,426) and (217,417.72) .. (217,407.5) -- cycle ;
	\draw    (177.5,365) -- (235.5,389) ;
	\draw    (235.5,365) -- (235.5,389) ;
	\draw    (293.5,365) -- (235.5,389) ;
	
	\draw (168,28.4) node [anchor=north west][inner sep=0.75pt]    {$x_{1}$};
	\draw (23,25) node [anchor=north west][inner sep=0.75pt]   [align=left] {Coordinates};
	\draw (222,28.4) node [anchor=north west][inner sep=0.75pt]    {$\cdots $};
	\draw (288,29.4) node [anchor=north west][inner sep=0.75pt]    {$x_{n}$};
	\draw (169,132.4) node [anchor=north west][inner sep=0.75pt]    {$g^{i}_{1}$};
	\draw (221,133.4) node [anchor=north west][inner sep=0.75pt]    {$\cdots $};
	\draw (23,133) node [anchor=north west][inner sep=0.75pt]   [align=left] {Atom $\displaystyle i$ features};
	\draw (280,134.4) node [anchor=north west][inner sep=0.75pt]    {$g^{i}_{k}$};
	\draw (219,204.4) node [anchor=north west][inner sep=0.75pt]    {$\cdots $};
	\draw (220,275.4) node [anchor=north west][inner sep=0.75pt]    {$\cdots $};
	\draw (33,327) node [anchor=north west][inner sep=0.75pt]   [align=left] {\begin{minipage}[lt]{69.62588000000001pt}\setlength\topsep{0pt}
		\begin{center}
		Atom-specific \\neural network
		\end{center}
		
		\end{minipage}};
	\draw (34,402) node [anchor=north west][inner sep=0.75pt]   [align=left] {Atom $\displaystyle i$ energy};

	\end{tikzpicture}
	\caption{Behler-Parrinello network architecture.}
	\label{Fig:BPN}
\end{figure}

The fundamental BPN architecture is depicted in Fig.~\ref{Fig:BPN}. For each atom $i$, the molecular coordinates are mapped to invariant features. A set of correlation functions, which describe the chemical environment of each atom, is employed in order to map the distances of neighboring atoms of a certain type and the angle between two neighbors of specific types. The aforementioned features are inputted into a dense NN, which returns the energy of atom $i$ in its environment. Input feature functions are designed taken into account that the energy is rototranslationally invariant, while equivalent atoms share their parameters. In the final step, all the atoms of a molecule are dentified and their atomic energies are summed. This guarantees permutation invariance. Parameter sharing combined with the summation principle offers also scalability, since it allows growing or shrinking the molecules network to any size, including ones that were never seen in the training data. The main limitation of BPNs is that they cannot accurately predict the energy surfaces in complex chemical environments.   

\subsubsection{Deep Potential Networks}\label{SSS:DPN}
\textit{DPNs} aim at providing an end-to-end representation of PESs, which employ atomic configuration directly at the input data, without decompositioning the contributions of different number of bodies~\cite{Zhang2018b}. Similarly to BPNs, the main challenge is to design a DNN, that takes into account both the rotational and permutational symmetries as well as the chemically equivalent atom. 

\begin{figure*}	
	\centering
	\includegraphics[width=0.9\linewidth]{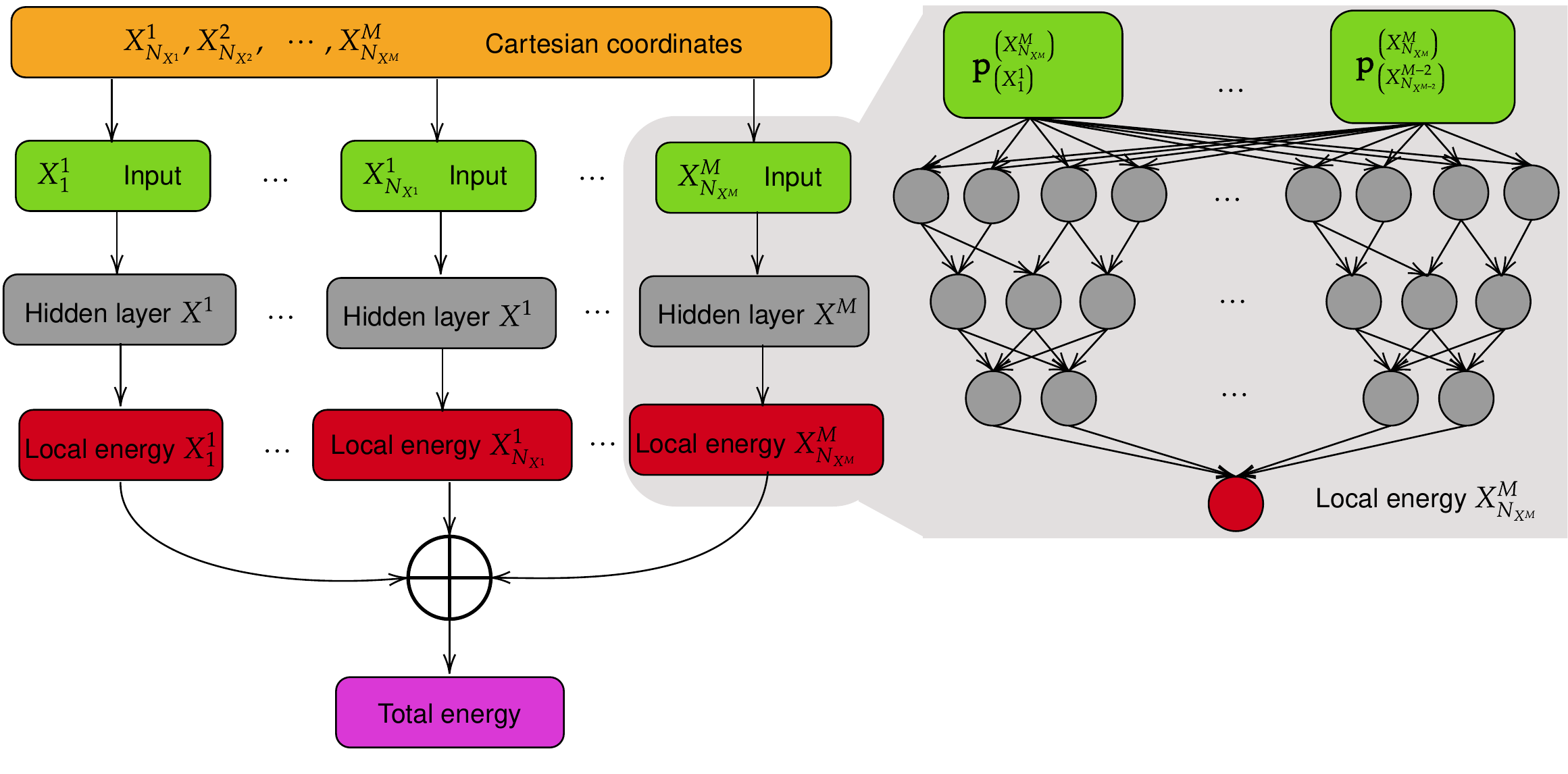}
	\caption{Deep potential net architecture.}
	\label{Fig:DeepPontentialNets}
	\hrulefill	
\end{figure*}

Let us consider a molecule that consists of $N_{X^i}$ atoms of type $X^i$, with $i=\{1, 2,\cdots, M\}$. As demonstrated in Fig.~\ref{Fig:DeepPontentialNets}, the DPN takes as inputs the Cartesian coordinates of each atom and feeds them in $\sum_{i=1}^{M}N_{X^i}$ almost independent sub-networks. Each of them provides a scalar output that corresponds to the local energy contribution to the PES, and maps a different atom in the system. Furthermore, they are coupled only through summation in the last step of this method, when the total energy of the molecule is computed. In order to ensure the permutational symmetry of the input, in each sub-network, the atoms are fed into different groups that corresponds to different atomic species. Within each group, the atoms are sorted in order to increase the distance to the origin. To further guarantee global permutation symmetry, the same parameters are assigned to all the sub-networks.         

\subsubsection{Deep Tensor Neural Networks}\label{SSS:DTNN}
Recently, several researchers have exploited the \textit{DTNN} capability to learn a multi-scale representation of the properties of molecules and materials from large-scale data in order to develop molecular and material simulators~\cite{Schuett2017,Rupp2012,Bartok2013}. In more detail, DTNN initially recognizes and constructs a representation vector for each one of the atoms within the chemical environment, and then it employs a tensor construction algorithm that iteratively learns higher-order representations, after interacting with all pairwise neighbors. 

\begin{figure*}
	\centering
	\includegraphics[width=0.95\linewidth]{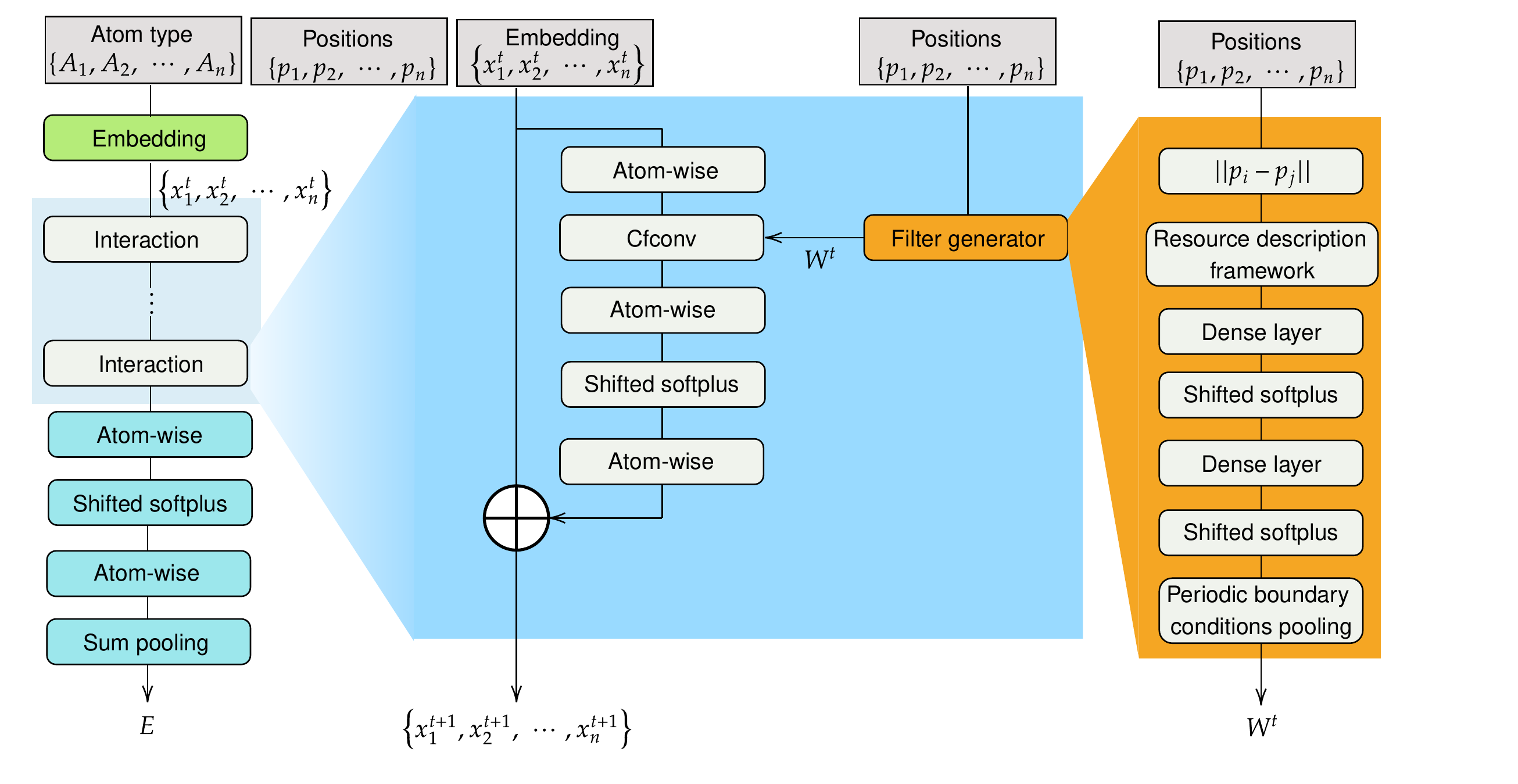}
	\caption{DTNN architecture.}
	\label{Fig:DTNN}
	\hrulefill
\end{figure*}

Figure~\ref{Fig:DTNN} presents a comprehensive example of DTNN architecture. The input, which consists of atom types and positions, is processed through several layers to produce atom-wise energies that are summed to a total energy. In the interaction layer, which is the most important one, atoms interact via continuous convolution functions. The variable $W^t$ stands for convolution weights that are returned from a filter generator function. Continuous convolutions are generated by DNNs that operate on interatomic distances, ensuring rototranslational invariance of the energy.        

DTNNs can accurately model a general QM molecular potential by training them in a diverse set of molecular energies~\cite{Schuett2017}. Their main disadvantage is that they are unable to perform energy predictions for systems larger than those included in the training set~\cite{Smith2017}.   

\subsubsection{SchNet}\label{SSS:SchNet}
SchNets can be considered as a special case of DTNN, since they both share atom embedding, interaction refinements and atom-wise energy contribution. Their main difference is that interactions in DTNNs are modeled by tensor layers, which provide atom representations. Parameter tensors are also used in order to combine the atom representations with inter-atomic distances~\cite{Schuett2018}.  On the other side, to model the interactions, SchNet employs  filter convolutions, which are interpreted as a special case of computational-efficient low-rank factorized tensor layers~\cite{Schuett2018a,Schutt2020}. 

Conventional SchNets use discrete convolution filters (DCFs), which are designed for pixelated image processing in computer vision~\cite{Jeong2011}. QM properties, like energy, are highly sensitive to position ambiguity. As a consequence, the accuracy of  a model that discretize the particles position in a grid is questionable. To solve this problem, in \cite{KristofT.Schuett2017}, the authors employed continuous convolutions in order to map the rototranslationally invariant inter-atomic distances to filter values, which are used in the convolution.      

\subsubsection{Accurate Neural Network Engine for Molecular Energies}\label{SSS:ANI}
\textit{Accurate neural network engine for molecular energies (ANAKIN-ME)}, or ANI for short, are networks that have been developed to break the walls built by DTNNs. The principle behind ANI is to develop modified symmetry functions (SmFs), which were introduced by BPNs, in order to develop NN potentials (NNPs). NNPs output single-atom atomic environments vectors (AEVs), as a molecular representation. AEVs allow energy prediction in complex chemical environments; thus, ANI solves the transferability problem of BPNs. By employing AEVs, the problem, which needs to be solved by ANI, is simplified into sampling statistically diverse set of molecular interactions within a predefined region of interest. To successfully solve this problem, a considerably large data set that spans molecular conformational and configurational space, is required. A trained ANI is capable of accurately predicting energies for molecules within the training set region~\cite{Gao2020}.    

\begin{figure}
	\centering
	\includegraphics[width=1.0\columnwidth]{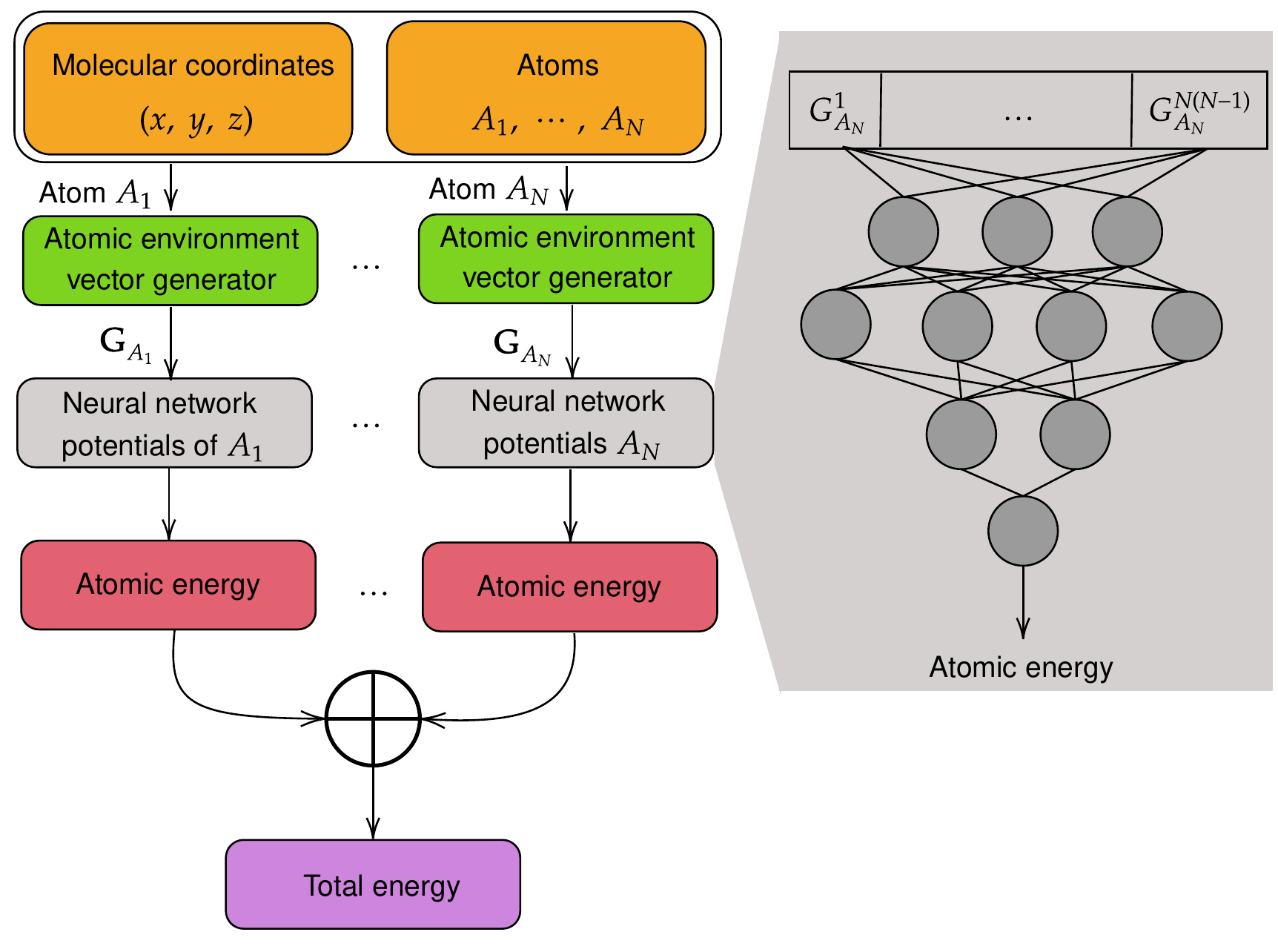}
	\caption{ANI architecture.}
	\label{Fig:ANI}
\end{figure}

As presented in Fig.~\ref{Fig:ANI}, ANI uses the molecular coordinates and the atoms in order to compute the AEV of each atom. The AEV of atom $A_i$ (with $i=1,\cdots, N$), $\mathbf{G}_{A_i}$, scrutinizes specific regions of $A_i$'s radial and angular chemical environment. Each $\mathbf{G}_{A_i}$ is inputted in a single NPP, which returns the energy of atom $i$. Finally, the total energy of a molecule is evaluated as the sum of the energies of each one of the atoms.    

\subsubsection{Coarse Graining Networks}\label{SSS:CGN}
A common approach in order to go beyond the time and length scales, accessible with computational expensive molecular dynamics simulations, is the \textit{coarse-graining (CG)} models. Towards this direction, several research works, including~\cite{Lyubartsev1995,Clementi2000,MuellerPlathe2002,Nielsen2003,Matysiak2004,Marrink2004,Matysiak2006,Wang2009,Chen2018,Wang2019}, developed CG energy functions for large molecular systems, which take into account either the macroscopic properties or the structural features of atomistic models. All the aforementioned contributions agreed on the importance of incorporating the physical constraints of the system in order to develop a successful model. The training data are usually obtained through atomistic molecular dynamics simulations. Values within physically forbidden regions are not sampled and not included in the training. As a result, the machine is unable to perform predictions far away the training data, without additional constraints. 

\begin{figure}
	\centering
	\includegraphics[width=0.55\columnwidth]{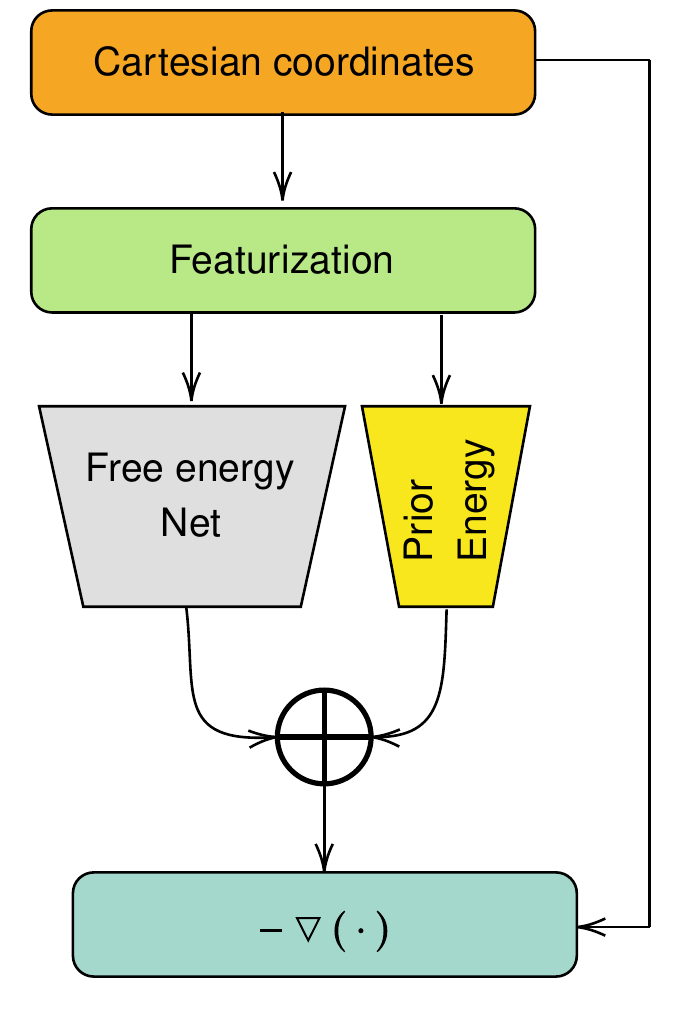}
	\caption{CG network architecture.}
	\label{Fig:CG}
\end{figure}

To countermeasure the aforementioned problem, CG networks employ regularization methods in order to enforce the correct asymptotic behavior of the energy when a nonphysical limit is violated. Similarly to BPNs and SchNets, CG networks initially translate the cartesian into internal coordinates, and use them to predict the rototranslationally invariant energy. Next, as illustrated in Fig.~\ref{Fig:CG}, the network learns the difference from a simple prior energy, which has been defined to have the correct asymptotic behavior~\cite{Wang2019}. Note that due to the fact that CG networks are capable of using available training data in order to correct the prior energy, its exact form is not required.  Likewise, CG networks compute the gradient of the total free energy with respect to the input configuration in order to predict the conservative and rotation-equivariant force fields. The force-matching loss minimization of this prediction is used as a training rule of the CG network. 

In practice, CGNs are used  to predict the thermodynamic of  chemical systems that are considerably larger than what is possible to simulate with atomistic resolution.  Moreover, there have been recently presented some indications that they can also used to approximate the system kinetics, through the addition of fictitious particles~\cite{Davtyan2016} or by employing spectral matching to train the CGN~\cite{Nueske2019}. 

\subsubsection{Neuromorphic Computing}\label{SSS:NC}
Neuromorphic computing \cite{Brown2019} is an emerging field, where the architecture of the brain is closely represented by the designed hardware-level system. The fundamental unit of neuromorphic computation is a memristor, which is a two-terminal device in which conductance is a function of the prior voltages in the device. Memristors were realized experimentally considering that many nanoscale materials exhibit memristive properties through ionic motion \cite{Strukov2008}. Nanophotonic systems are also utilized for neuromorphic computing and especially for the realization of deep learning networks \cite{Shen2017} and adsorption-based photonic NNs~\cite{George2019}.

Although neuromorphic computing and memristors tend to be a scalable practical technology, large area uniformity, reproducibility of the components, switching speed/efficiency and total lifetime in terms of cycles remain quite challenging aspects \cite{Zidan2018}, which require either the development of novel memristive systems or improvements to existing systems. To this end, integration with existing complementary metal-oxide-semiconductor (CMOS) platforms and competitive performance advantage over CMOS neurons must be explored. These analog networks, after they are trained, can be highly efficient, however their training does not utilize digital logic and, thus, lacks flexibility \cite{Brown2019}. 

\subsection{Regression}\label{SS:Regression}

In this section, we discuss the regression methods that are commonly-used in the field of nano-scale biomedical engineering. Regression aims at characterizing the relationships among different variables. Three types of variables are identified in regression problems, namely predictors, objective, and distortion. A predictor, $x_i$, with $i\in[1, N]$, is an independent variable, while the objective, $Y$, is the dependent one. Moreover, let $d$ stand for the distortion parameter that model unknown parameters of the problem under investigation and affect the estimated value of the dependent parameter. Mathematically speaking, the objective of regression methods is to find the regression function $f\left(x_1, \cdots, x_N, d\right)$ that satisfies 
	\begin{align}
		Y=f\left(x_1, \cdots, x_N, d\right).
	\end{align} 
An important step for regression methods is to specify the form of the regression function. Based on the selected regression function, different regression methods can be identified. The rest of this section presents the regression methods that are commonly used in nano-scale biomedical engineering. In more detail, Section~\ref{SSS:LR} provides a brief overview of logistic regression (LR), whereas Sections~\ref{SSS:MvLR} and~\ref{SSS:CvR} respectively discuss multivariate linear regression (MvLR) and classification via regression. Finally, Sections~\ref{SSS:LWL} and~\ref{SSS:SF} respectively report the operating principles of local weighted learning (LWL) and scoring functions (SFs). Table~\ref{T:Reg_approaches_ML_challenges} summarizes the applications of regression methodologies in nano-scale biomedical engineering.

\begin{table*}[]
	\centering
	\caption{Regression applications in nano-scale biomedical engineering. }
	\label{T:Reg_approaches_ML_challenges}
	\begin{tabular}{|c||l|c|r|}
		\hline
		\textbf{Paper} & \textbf{Application} & \textbf{Method} & \textbf{Description} \\
		\hline 	\hline
		\cite{Yamankurt2019} & Nanomedicine design & LR & Structure-activity relationships and design rules for spherical nucleic acids\\
		\hline
		\cite{PerezEspinoza2019} & Treatment design & LR & Classification of clinical trials based on an unsupervised ML algorithm \\
		\hline
		\cite{Sayes2010} & Chemical properties modeling & MvLR & Comparison of predictive computational models for nanoparticle \\ & & & induced cytotoxicity \\
		\hline
		\cite{Jones2015} & Chemical properties modeling & Classification via Regression & Elimination of silico materials from potential human applications \\
		\hline
		\cite{Jones2015} & Chemical properties modeling & LWL, SVM & Cytotoxicity prediction of NPs in biological systems \\
		\hline
		\cite{Ain2015} & Chemical properties modeling & SF & Binding affinity and virtual screening prediction for nano-structures \\
		\hline
		\cite{Li2018} & Chemical properties modeling & SF & Quantification of the impact of protein structure on binding affinity \\
		\hline
	\end{tabular}
\end{table*}

\subsubsection{Logistic Regression}\label{SSS:LR}
LR is a supervised learning classification algorithm used to predict the probability of a target variable. The concept behind the target or the dependent variable is \textit{dichotomous}, which means that there would be only two possible classes. LR can fit trends that are more complex than linear regression, but it still treats multiple properties as linearly related and is still a linear model. LR is named after the function used at the core of the method, the logistic function, which can take any real-valued number and map it into a value between $0$ and $1$. To provide a better understanding of LR, let us consider the binary classification problem in which $z$ is the dependent variable and $\mathbf{x}=[x_1, x_2,\cdots, x_N]$ are the $N$ independent variables. Since, for a fixed $\mathbf{x}$, $z$ follows a Bernoulli distribution, the probabilities $\Pr\left(z=1\left|\mathbf{x}\right.\right)$ and $\Pr\left(z=0\left|\mathbf{x}\right.\right)$ can be respectively obtained~as
\begin{align}
	\Pr\left(z=1\left|\mathbf{x}\right.\right)&= \frac{1}{1+\exp\left(-f(\mathbf{x})\right)} \label{Eq:P_bern},\\
	\Pr\left(z=0\left|\mathbf{x}\right.\right)&= 1- 	\Pr\left(z=1\left|\mathbf{x}\right.\right) \nonumber \\ &= \frac{1}{1+\exp\left(f(\mathbf{x})\right)},
\end{align}
where 
\begin{align}
	f(\mathbf{x})=c_0+\sum_{i=1}^{N} c_i x_i,
\end{align}
with $c_0, c_1, \cdots, c_N$ being the regression coefficients. From~\eqref{Eq:P_bern}, we can straightforwardly obtain $f(\mathbf{x})$ as
\begin{align}
	f(\mathbf{x}) = \ln\left( \frac{\Pr\left(z=1\left|\mathbf{x}\right.\right)}{1-\Pr\left(z=1\left|\mathbf{x}\right.\right)} \right).
\end{align}
For a given training-set of length $N$, $\{z_i, x_{i,1}, \cdots, x_{i,M}\}$ with $i\in[1, N]$,  the regression coefficients can be estimated by employing the maximum likelihood approach.

 LR has been used extensively in biomedical applications, such as disease detection. Indicatively, in~\cite{Yamankurt2019},  LR was used to determine structure-activity relationships and design rules for spherical nucleic acids functioning as cancer-vaccine candidates. Moreover, in~\cite{PerezEspinoza2019}, it has been used for nano-medicine-based clinical trials classification and treatment~development.

\subsubsection{Multivariate Linear Regression}\label{SSS:MvLR}
Following the previous analysis, when multiple correlated dependent variables are predicted rather than a single scalar variable, the method is called MvLR. This method is a generalization of  multiple linear regression and incorporates a number of different statistical models, such as analysis of variance (ANOVA), $t$-test, $F$-test, and more. MvLR has been used in ML for several nano-scale biomedical applications. Among the most successful ones is the prediction of cytotoxicity in NPs~\cite{Sayes2010}.

The MvLR model can be expressed in the~form
\begin{align}
y_{ik} = b_{0k} + \sum^p_{j=1}b_{jk} x_{ij} + e_{ik} , 
\end{align}
where $y_{ik}$ is the $k$-th response for the $i$-th observation, $b_{0k}$ is the regression intercept for the $k$-th response, $b_{jk}$ is the $j$-th predictor's regression slope for the $k$-th response, $x_{ij}$ is the $j$-th predictor for the $i$-th observation, $e_{ik}$ is a Gaussian error term for the $k$-th response, $k \in \left[1,m\right]$ and $i \in \left[1,n\right]$.

\subsubsection{Classification via Regression}\label{SSS:CvR}
Conventionally, when dealing with discrete classes in ML,  a classification method is used, while a regression method is applied, when dealing with continuous outputs. However, it is possible to perform classification through a regression method. The class is binarized and one regression model is built for each class value. In~\cite{Jones2015}, in order to predict cytotoxicity of certain NPs, \textit{classification via regression} is among the methods that were evaluated, in order to eliminate in silico materials from potential human applications. 

\subsubsection{Local Weighted Learning}\label{SSS:LWL}
In the majority of learning methods, a global solution can be reached using the entirety of the training data. \textit{LWL} offers an alternative approach at a much lower cost, by creating a local model, based on the neighboring data of a point of interest. In general, data points in the neighborhood of the point of interest, called query point, are assigned a weight based on a kernel function and their respective distance from the query point. The goal of the method is to find the regression coefficient that minimizes a cost function, similar to most regression methods. Due to its nature as a local approximation, LWL allows for easy addition of new training data. Depending on whether LWL stores in memory or not the entirely of the training data, LWL-based methods can be divided into memory-based and purely incremental, respectively~\cite{Atkeson1997}.

Recently, LWL was used in \cite{Jones2015}, in order to predict the cytotoxicity of NPs in biological systems given an ensemble of attributes. It is found that when the data were further validated, the LWL classifier was the only one out of a set of classifiers that could offer predictions with high~accuracy.

\subsubsection{Machine Learning Scoring Functions}\label{SSS:SF}
SFs can be used to assess the docking performance, i.e. to predict how a small molecule binds to a target can be applied if a structural model of such target is available. However, despite the notable research efforts dedicated in the last years to improve the accuracy of SFs for structure-based binding affinity prediction, the achieved progress seems to be limited. ML-SFs  have recently proposed to fill this performance gap. These are based on ML regression models without a predetermined functional form, and thus,  are able to efficiently exploit a much larger amount of experimental data~\cite{Ain2015}. The concept behind ML-SFs  is that the classical approach of using linear regression with a small number of expert-selected structural features can be strongly improved by using ML on nonlinear regression together with comprehensive data-driven feature selection (FS). Also, in \cite{Li2018} investigated whether the superiority of ML-SFs over classical SFs on average across targets, is exclusively due to the presence of training  with highly similar proteins to those in the test~set. 

\begin{figure}
	\centering
	\includegraphics[width=1.0\columnwidth]{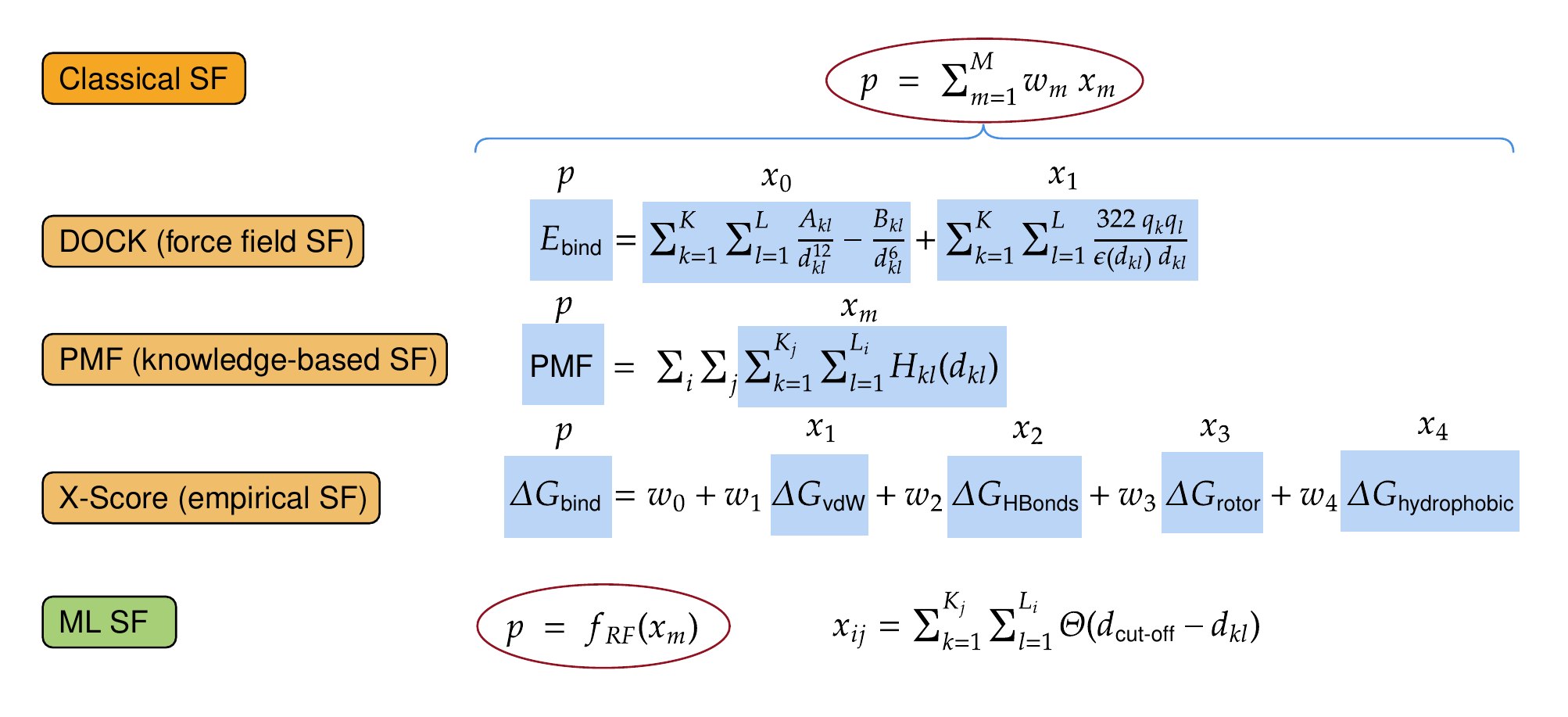}
	\caption{Examples of classical and ML-SFs (from~\cite{Ain2015})} \label{fig_SF}
\end{figure}

In Fig.~\ref{fig_SF} examples of classical and ML-SFs are depicted~\cite{Ain2015}. The first three (DOCK, PMF and X-SCORE) are classical SFs, which are distinguished by the employed structural descriptors. As it is evident, they all assume an additive functional form. On the other side, ML-SFs do not make assumptions about their functional form, which is inferred from the training~data.

\subsection{Support Vector Machine}\label{SS:SVM}
NNs can be efficiently used in classification, when a huge number of data is available for training. However, in many cases this method outputs a local optimal solution instead of a global one. SVM is a supervised learning technique, which can overcome the shortcomings of NNs in classification and regression. For a brief but useful description of the SVM please see \cite{gholami2017} and references therein. Next, for the help of the reader the SVM is summarized by using~\cite{gholami2017}. 

\begin{figure}
	\centering
	\includegraphics[width=0.7\columnwidth]{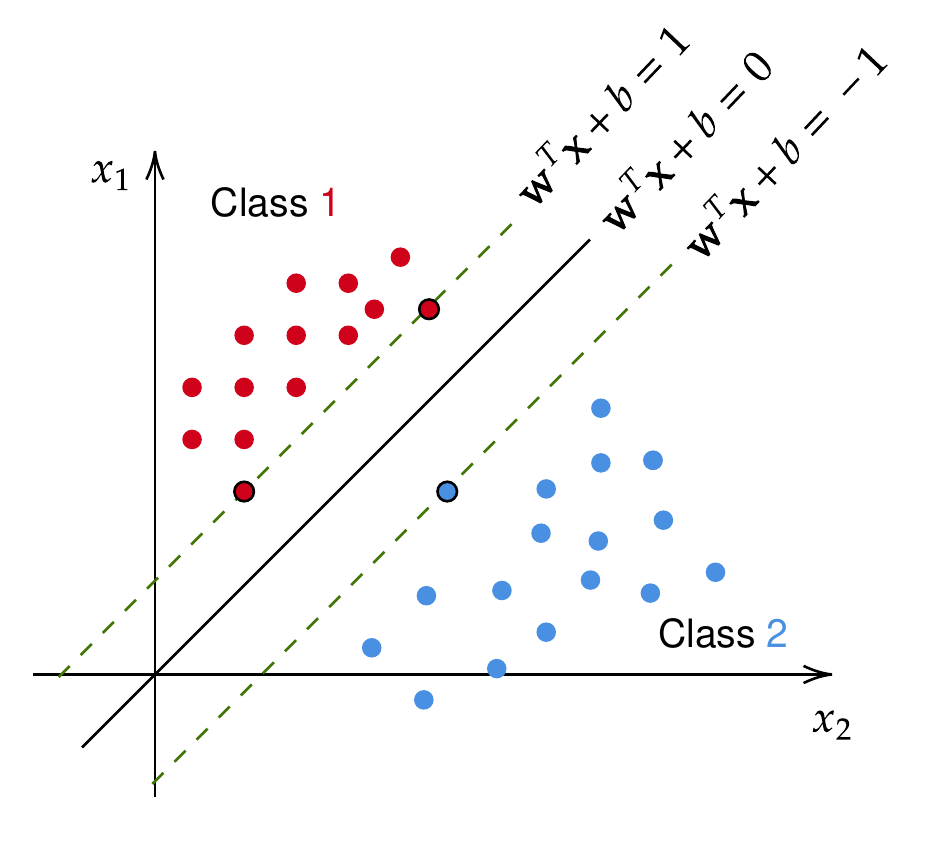}
	\caption{The SVM method~\cite{gholami2017}} \label{SVM}
\end{figure}

The aim of SVM is to find a classification criterion, which can effectively distinguish data at the testing stage. This criterion can be a line for two classes data, with a maximum distance of each class. This linear classifier is also known as an \textit{optimal hyperplane}. In Fig. \ref{SVM}, the linear hyperplane is described  for a set of training data, $\mathbf{x}= (1, 2, 3,...,n)$, as 
\begin{equation}\label{eq:criterion}
\mathbf{w}^{T} \mathbf{x}+b=0,
\end{equation}
where $\mathbf{w}$ is an \textit{n}-dimensional vector and $b$ is a bias (error)~term. 

This hyperplane should satisfy two specific properties: (1) the least possible error in data separation, and  (2) the distance from the closest data of each class must be the maximum one. 
Under these conditions, data of each class can only belong to the left  of the hyperplane. Therefore, two margins can be defined to ensure the separability of data as
\begin{equation}\label{eq:margins}
\mathbf{w}^{T} \mathbf{x}+b\left\{\begin{array}{ll}
\geq 1 & \text { for } y_{i}=1 \\
\leq-1 & \text { for } y_{i}=-1
\end{array}\right.
\end{equation}
The general equation of the SVM for a linearly separable case, which would be subjected to two constraints as
\begin{equation}\label{eq:general}
\begin{array}{l}
\operatorname{max} { L_{d}(\alpha)=\sum_{i=1}^{N} \alpha_{i}-\frac{1}{2} \sum_{i, j=1}^{N} y_{i} y_{j} \alpha_{i} \alpha_{j} x_{i}^{T} x_{j} }\\
\text { s.t. }\left\{\begin{array}{l}
\alpha_{i} \geq 0 \\
\sum_{i=1}^{N} \alpha_{i} y_{i}=0
\end{array}\right.
\end{array}
\end{equation}
where $\alpha$ is a Lagrange multiplier. 

Eq. \eqref{eq:general} is used in order to  find the support vectors and their corresponding input data. The parameter $w$ of the hyperplane (decision function) can then be obtained~as
\begin{equation}\label{eq:w}
w_{0}=\sum_{i=1}^{N} \alpha_{i} x_{i} y_{i}
\end{equation}
and the bias parameter $b$ can be calculated as
\begin{equation}\label{eq:b}
b_{0}=\frac{1}{N} \sum_{S=1}^{N}\left(y_{S}-w^{T} x_{S}\right)
\end{equation}
More details about the use of the linear  as well as the non-linear SVM  methods, can be found in~\cite{gholami2017}.

An indicative training algorithm for SVM is the sequential minimal optimization (SMO). \textit{SMO} is a training algorithm for SVMs. The training of an SVM requires the solution of a large quadratic programming (QP) optimization problem. Conventionally, the QP problem is solved by complex numerical methods, however SMO breaks down the problem into the smallest possible and solves it analytically, thus reducing significantly the amount of required time. SMO chooses two Lagrange multipliers to optimize, which can be done analytically, and updates the SVM accordingly. Interestingly, the smallest amount of Lagrange multipliers to solve the dual problem is two, one from a box constraint and one from linear constraint, meaning the minimum lies in a diagonal line segment. If only one multiplier was used in SMO, it would not be able to guarantee that the linear constraint is fulfilled at every step \cite{Platt1998}. Moreover, SMO ensures convergence using Osuna's theorem, since it is a special case of the Osuna algorithm, that is guaranteed to converge \cite{Osuna97}. Recently, in \cite{Jones2015}, SMO was one of the classifiers used to predict cytotoxicity of Polyamidoamine (PAMAM) dendrimers, well documented NPs that have been proposed as suitable carriers of various bioactive~agents. 

SVM have been applied in many significant applications in bioinformatics and bioemedical engineering. Examples include: protein classification, detection of the splice sites,  analysis of the gene expression, including gene selection for microarray data, where a special type of SVM called  Potential SVM has been successfully used for analysis of brain tumor data set, lymphoma data set, and breast cancer data set (\cite{Cyran2013} and references therein).

Recently, SVM was considered in MCs. Specifically, in \cite{Li2020} the authors proposed injection velocity as a very promising  modulation method in turbulent diffusion channels, which can be applied in several practical applications as in pollution monitoring, where inferring the pollutant ejection velocity may give an indication to the rate of underlying activities. In order to increase the reliability of inference, a time difference SVM technique was proposed to identify the initial velocities. It was shown that this can be achieved with very high accuracy.

In~\cite{Mohamed2020} a diffused molecular communication system model was proposed with the use a spherical transceiver and a trapezoidal container. The model was developed through SVM-Regression and other ML techniques, and it was shown that it performs with high accuracy, especially if long distance is~assumed. 


\subsection{$k-$Nearest Neighbors}\label{SS:KNN}
KNN is a supervised ML classifier and  regressor. It is based on the evaluation of  the distance between the test data and the input and gives the prediction accordingly. The concept behind  KNN  is the classification of a class of data, based on the k nearest neighbors. Other names of this ML algorithm are   memory-based classification and example-based classification or case-based classification. 

KNN classification consists of two stages: the determination of the nearest neighbors and the  class using those neighbors. A brief description of the KNN algorithms is as follows \cite{kNN}: Let us considered a training data set $D$ consisted of $\left(\mathbf{x}_{i}\right)_{i \in[1,|D|]}$   training samples. The examples are described by a set of features $F$, which are normalized in the range$[0,1]$. Each training example is labelled with a class label $y_{j} \in Y$. The aim is to classify an unknown example $q$. To achieve this, for each $\mathbf{x}_{i} \in D$, we evaluate the distance between $q$ and $x_i$ as
\begin{equation}
d\left(\mathbf{q}, \mathbf{x}_{i}\right)=\sum_{f \in F} w_{f} \delta\left(\mathbf{q}_{f}, \mathbf{x}_{i f}\right)
\end{equation}

There are many  choices for this distance metric; a fundamental metric, based on the Euclidian distance,  for continuous and discrete attributes is
\begin{equation}
\delta\left(\mathbf{q}_{f}, \mathbf{x}_{i f}\right)=\left\{\begin{array}{ll}
0 & f \text { discrete and } \mathbf{q}_{f}=\mathbf{x}_{i f} \\
1 & f \text { discrete and } \mathbf{q}_{f} \neq \mathbf{x}_{i f} \\
\left|\mathbf{q}_{f}-\mathbf{x}_{i f}\right| f & \text { continuous }
\end{array}\right.
\end{equation}

The KNNs are selected based on this distance metric. There are a variety of ways in which the KNN can be used to determine the class of $q$. The most straightforward approach is to assign the majority class among the nearest neighbors to the query.

\begin{figure}
	\centering
	\includegraphics[width=0.7\columnwidth]{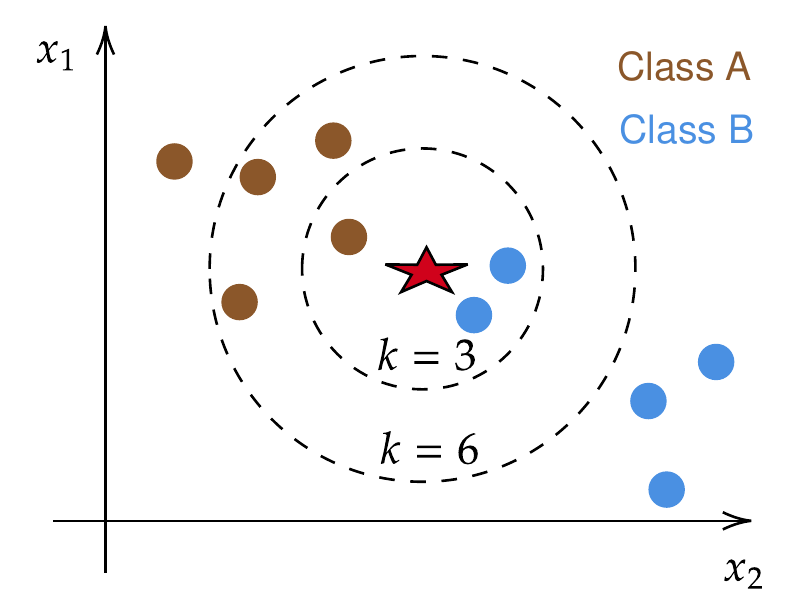}
	\caption{The KNN ML method.}
	\label{Fig:kNN}
\end{figure}

Figure~\ref{Fig:kNN} depicts a  $3$ and $6$ KNN on a two-class problem in a two-dimensional space~\cite{kNN}. The red star represents the test data point whose value is $( 2 , 1 , 3 )$. The test point is surrounded by yellow and blue dots which represent the two classes. The distance from our test point to each of the dots present on the graph. Since there are 10 dots, we get 10 distances. We determine the lowest distance and predict that it belongs to the same class of its nearest neighbor. If a yellow dot is the closest, we predict that our test data point is also a yellow dot. In some cases, you can also get two distances which exactly equal. Here, we take into consideration a third data point and calculate its distance from the test data. In Fig.~\ref{Fig:kNN}  the test data lies in between the yellow and the blue dot. We considered the distance from the third data point and predicted that the test data is of the blue~class. 

The advantages of KNN are simple implementation and no need for prior assumption of the data. The disadvantage of KNN is the high prediction time.

\subsection{Dimentionality Reduction}\label{SS:DR}
This section is devoted to discussing dimentionality reduction methods. Dimentionality reduction constitutes the preparatory phase of ML, because the initially acquired raw data may contain some irrelevant or redundant
features. Next, a comprehensive description of FS is provided in Section~\ref{SSS:FS}. Likewise, principal component analysis (PCA) and linear discriminant analysis (LDA) are respectively discussed in Sections~\ref{SSS:PCA} and~\ref{SSS:LDA}. Finally, Section~\ref{SSS:ICA} presents the fundamentals of independent component analysis (ICA). Table~\ref{T:Dim_approaches_ML_challenges} report the dimentionality reduction methodologies applications in nano-scale biomedical engineering.

\begin{table*}[]
	\centering
	\caption{Dimentionality reduction applications in nano-scale biomedical engineering. }
	\label{T:Dim_approaches_ML_challenges}
	\begin{tabular}{|c||l|c|r|}
		\hline
		\textbf{Paper} & \textbf{Application} & \textbf{Method} & \textbf{Description} \\
		\hline 	\hline
		\cite{Kourou2015} & Disease detection & FS & Cancer prognosis and prediction \\
		\hline
		\cite{JIE2014} & Disease detection & FS & Breast cancer detection \\
		\hline
		\cite{Ren2012} & Disease detection & FS & Metastatic cancer detection \\
		\hline
		\cite{Kim2016} & Disease detection & FS & Improved diagnoses based on composite biomarker signature \\
		\hline
		\cite{Jesse2009} & Image analysis & PCA & Spectroscopic image analysis \\
		\hline
		\cite{Subasi2010} & Signal analysis & PCA, LDA & Classification of EEG signals \\
		\hline
	\end{tabular}
\end{table*}

\subsubsection{Feature Selection}\label{SSS:FS}
\textit{FS} reduces the complexity of a problem by detecting the subset of features that contribute most to the results. FS is one of the core concepts in ML, which hugely impacts the achievable performance. It is important to point out that FS is different from dimensionality reduction. Both methods seek to reduce the number of attributes in the data set, but a dimensionality reduction method do so by creating new combinations of attributes, whereas FS methods include and exclude attributes present in the data without changing~them.

Combining ML algorithms with FS has been proven to be very useful for disease detection~\cite{Kourou2015,JIE2014}. It highlights the features associated with a specific target from a larger pool. For instance, in~\cite{Ren2012}, a classification algorithm was used to analyze $10000$ genes from $200$ cancer patients, while FS was used to associate $50$ of them with metastatic prostate cancer. The selected features were then utilized as biomarker signature criteria in a ML algorithm for classification and diagnostics. Furthermore, recent research efforts provided evidence that combining data from multiple sources, such as transcriptomics and metabolomics to create composite signatures can improve the accuracy of biomarker signatures and disease diagnoses~\cite{Kim2016}.

\subsubsection{Principal Component Analysis}\label{SSS:PCA}
\textit{PCA} \cite{Jesse2009,Subasi2010,Brown2019,Cao2003} is an approach to solve the problem of blind source separation (BSS), which aims at the separation of a set of source signals from a set of mixed signals, with little information about the source signals or the mixing process. PCA utilizes the eigenvectors of the covariance matrix to determine which linear combinations of input variables contain the most information. It can also be used for feature extraction and dimensionality reduction. For cases with strong response variations, PCA allows an effective approach to rapidly process, de-noise, and compress data, however it cannot explicitly classify data.

More specifically, in PCA, the dimensional data are represented in a lower-dimensional space, reducing the degrees of freedom, the space and time complexities.  PCA aims to represent the data in a space that best expresses the variation in a sum-squared error sense and is utilized for segmenting signals from multiple sources. As in standard clustering methods, it is useful if the number of the independent components is determined. 
Using the covariance matrix $\mathbf{C} = \mathbf{A} \mathbf{A}^\mathrm{T}$, where $\mathbf{A}$ denotes the matrix of all experimental data points, the eigenvectors $w_k$ and the corresponding eigenvalues $\lambda_k$ can be calculated. 
The eigenvectors are orthogonal and are chosen in order for the corresponding eigenvalues to be placed in descending order, i.e, $\lambda_1 > \lambda_2 > ...$. To this end, the first eigenvector $w_1$ contains the most information and the amount of information decreases in the following eigenvectors. Therefore, the majority of the information is contained in a number of eigenvectors, whereas the remaining ones are dominated by noise.

\subsubsection{Linear Discriminant Analysis}\label{SSS:LDA}
\textit{LDA} is another method for the solution of the BSS problem~\cite{Subasi2010,Brown2019}.
In LDA, linear combinations of parameters that optimally classify data are identified and the main goal is to reduce the dimension of data.
LDA has been used with a nanofluidic system to interpret gene expression data from exosomes and thus, to classify the disease state of patients. More specifically, LDA aims to create a new variable that is a combination of the original predictors, by maximizing the differences between the predefined groups with respect to the new variable. 
The predictor scores are utilized in order to form the discriminant score, which constitutes a single new composite variable. 
Therefore, the use of LDA results in an significant data dimension reduction technique that compresses the p-dimensional predictors into a one-dimensional line.
Although at the end of the process the desired result is that each class will have a normal distribution of discriminant scores with the largest possible difference in mean scores between the classes, some overlap between the discriminant score distributions exists. The degree of this overlap represent a measure of the success of LDA. The discriminant function which is used to calculate the discriminant scores can be expressed as
\begin{equation} \label{LDA}
D = w_1 Z_1 + w_2 Z_2 + ... + w_p Z_p,
\end{equation}
where $w_k$ and $Z_k$ with $k=1,...p$ denote the weights and predictors, respectively. From~\eqref{LDA}, it can be observed that the discriminant score is a weighted linear combination of the predictors. The estimation of the weights aims to maximize the difference between each class mean discriminant scores. To this end, the predictors which are not similar with respect to the class mean discriminant scores will have larger weights, whereas the weights will reduce the more similar the class means are \cite{Fielding2006}.

\subsubsection{Independent Component Analysis}\label{SSS:ICA}
\textit{ICA }\cite{Subasi2010,Brown2019,Cao2003} was introduced in \cite{Comon1994} and is another approach to the solution of the BSS problem. According to ICA, the original inputs are transformed into features, which are mutually  independent and the non-orthogonal basis vectors that correspond to the correlations of the data are identified through higher order statistics.  The use of the last one is needed, since the components are statistically independent, i.e., the joint  PDF of the components is obtained as the product of the PDFs of all components.

Let consider $c$ independent scalar source signals $x_k(t)$, with $K = 1, ... , c$ and $t$ being a time index. The $c$ signals can be grouped into a  zero mean.vector $x(t)$. Assuming that there is no noise and considering the independence of the components, the joint PDF can be expressed as
\begin{equation}
f_{x}(x) = \prod_{k=1}^{c} f_{x_k}(x_k).
\end{equation}
An d-dimensional data vector, $y(t)$, can be observed at each moment through,
\begin{equation}
y(t) = A x(t)
\end{equation}
where $A$ is a $c \times d$ scalar matrix with $d \geq c$.

ICA aims to recover the source signals from the sensed signals, thus the real matrix $W = A^{-1}$ has to be determined. To this end, the determination of $A$ is performed by maximum-likelihood techniques. An estimate of the density, termed as $\hat{f}_y(y;a)$, is used and the parameter vector $a$, that minimizes the difference between the source distribution and the estimate has to be determined. It should be highlighted that $a$ is the basis vector of $A$ and, thus, $\hat{f}_y(y;a)$ is an estimate of $f_y(y)$.

\subsection{Gradient Descent Method} \label{SS:GD}
When there are one or more inputs the optimization of the coefficients by iteratively minimizing the error of the model on the training data becomes a very important  procedure. This operation is called GD and initiates with random values for each coefficient. The sum of the squared errors is calculated for each pair of input and output values. A learning rate is used as a scale factor and the coefficients are updated to minimize the error. The process is repeated until a minimum sum squared error is achieved or no further improvement is possible. In practice, GD is taught using a linear regression model due to its straightforward nature and it proves to be useful for very large datasets~\cite{ruder2016overview}. 

GD is one of the most popular algorithms to optimize in NNs and has been extensively used in nano-scale biomedical engineering. For example, in~\cite{Peurifoy2018}, the authors proposed a method to use ANNs to approximate light scattering by multi-layer NPs and used the GD for optimizing the input parameters of the NN.

\subsection{Active Learning}\label{SS:AL}
In AL, also known as the optimal design of experiments, a surrogate model is created from a given data set, and then the model is used to select which data should be obtained next~\cite{Settles2012}. The selected data are added to the original data set and then used to create an updated surrogate model. The process is repeated iteratively such that the surrogate model is improved continuously. In contrast to classic ML methods, the identifier of an AL system is that it develops and tests new hypotheses as part of a continuing, interactive learning process. This method of iterative surrogate model screening has already been used in other fields, such as drug discovery and molecular property prediction~\cite{Warmuth2003,Gubaev2018}.

\subsection{Bayesian Machine Learning}\label{SS:BML}
In addition to the Bayes Theorem is a powerful tool in statistics, it is also widely used in ML to develop models for classification, such as the \textit{Optimal Bayes} classifier and \textit{Naive Bayes}. The optimal Bayes classifier selects the class that presents the largest a posteriori probability of occurrence. It can be shown, that among all classifiers, the Optimal Bayes classifier has the lowest error probability. In most real-life applications the posterior distribution is unknown but can rather be estimated. In this case, the Naive Bayes classifier approximates the optimal Bayes classifier by looking at the empirical distribution and assuming independence of predictors. So, the Naive Bayes classifier is a simple but suboptimal solution. It should be mentioned that Naive Bayes can be coupled with a variety of methods to improve the accuracy \cite{Witten2011}. Furthermore, since it relies on the computation of closed-form expressions of a posteriori probabilities, it takes linear time to compute, in contrast to expensive iterative approximations that are commonly used in other methods. 

Assuming an instance that is represented by the observation of $n$ features, $\mathbf{x} = (x_1, \dots, x_n)$, Naive Bayes assigns a probability $p(C_k| \mathbf{x})$ for each possible class $C_k$ among $K$ possible outcomes. According to Bayes' theorem, the posterior probability is given by the prior times the likelihood over the evidence, i.e. 
\begin{equation}
p(C_k| \mathbf{x}) = \frac{p(C_k)p(\mathbf{x}|C_k)}{p(\mathbf{x})}.
\end{equation}
The evidence is not dependent on $C$ so it is of no interest. Naive Bayes is a naive classifier because it assumes that all features in $\mathbf{x}$ are mutually independent conditioned on $C_k$. Therefore, it assigns a class label as
\begin{equation}
\hat{y} = \underset{k\in\{1,\dots,K\}}{\mathrm{argmax}} p(C_k) \prod_{i=1}^np(x_i|C_k).
\end{equation}
Bayesian analysis and ML are playing an important role in various aspects of nanotechnology and related molecular-scale research. Recently it has been shown that an atomic version of Green's function and Bayesian optimization is capable of optimizing the interfacial thermal conductance of Si-Si and Si-Ge nano-structures~\cite{Ju2017}. This method was able to identify the optimal structures between $60000$ candidate structures. Furthermore, more recent works have relaxed the data requirement limitations by adapting output parameters to unsupervised learning methods such as Bayesian statistical methods that do not rely on an external reference~\cite{Bryan2013,Huang2013,Bauer2015}. Naive Bayes has been applied to predict cytotoxicity of PAMAM dendrimers, which are well documented NPs that have been proposed as suitable carriers of various bioactive agents, in~\cite{Jones2015}. By pre-processing the data, Naive Bayes presented substantial improvement in the accuracy despite its simplicity, thus, outperforming the classification methods used in~\cite{Jones2015}. 



\subsection{Decision Tree Learning}\label{SS:DTL}
\textit{DTL} is a predictive modeling technique used in ML, which uses a decision tree to draw conclusions about the target value based on observations. In the tree paradigm, the target values are represented as leaves, while the observations are denoted by branches. There are two types of DTL, namely classification and regression trees. In the former, the target variable belongs in a discrete set of values, while in the latter the target variable is continuous. Furthermore, some techniques, such as bagged trees and bootstrap aggregated decision trees, use multiple decision trees. In more detail, the bagged trees method builds an ensemble incrementally by training each new instance to emphasize the training instances that were previously mis-modeled. The bootstrap aggregated decision trees is an early ensemble method that creates multiple decision trees by resampling training data and voting the trees for a consensus~prediction.

DTL has been used extensively in nano-medicine by optimizing material properties according to predicted interactions with the target drug, biological fluids, immune system, vasculature, and cell membranes, all affecting therapeutic efficacy~\cite{Adir2019}. Specifically, in~\cite{JAGLA2005}, decision trees were used for classification of effective and ineffective sequences for Ribonucleic acid interference (RNAi) in order to recognize key features in their design. In addition, several algorithms have been developed over the years that improve the accuracy and efficiency of DTL. For instance, the J48 algorithm is considered among the best algorithms with regards to accuracy and has been used in various biomedical tasks, such as predicting cytotoxicity, measured as cell viability~\cite{HorevAzaria2013,Jones2015}.

Next, we present the most commonly used DTL methods. In this direction, Bootstrap aggregating  (bagging) is revisited in Section~\ref{SSS:Bagging}, while the operating principles of bagged trees are highlighted in Section~\ref{SSS:BTs}. Moreover, the fundamentals of bagged Bayes trees are discussed in Section~\ref{SSS:NBT}, whereas the adaptive boosting (AdaBoost) approach is reported in Section~\ref{SSS:AdaBoost}.  Finally, descriptions of random forest (RForest) and M5P approaches are respectively delivered in Sections~\ref{SSS:RF} and~\ref{SSS:M5P}

\begin{table*}[]
	\centering
	\caption{Decision tree learning applications in nano-scale biomedical engineering. }
	\label{T:Tree_approaches_ML_challenges}
	\begin{tabular}{|c||l|c|r|}
		\hline
		\textbf{Paper} & \textbf{Application} & \textbf{Method} & \textbf{Description} \\
		\hline 	\hline
		\cite{Adir2019} & Disease treatment & DTL & Cancer treatment \\
		\hline
		\cite{JAGLA2005} & Chemical properties modeling & DTL & Feature recognition in the design of RNA sequences \\
		\hline
		\cite{HorevAzaria2013} & Chemical properties modeling & DTL & Prediction of cytotoxicity \\
		\hline
		\cite{Jones2015} & Chemical properties modeling & DTL, NBTree & Prediction of cytotoxicity \\
		\hline
		\cite{Liu2013} & Chemical properties modeling & Bagging, M5P & Prediction of cytotoxicity \\
		\hline
		\cite{Yekkala2017} & Disease prediction & Bagged tree, AdaBoost & Heart disease prediction \\
		\hline
		\cite{Mallick2004} & Disease detection & AdaBoost & Particle detection in cryo-electon micrographs \\
		\hline
		\cite{John2016} & Chemical properties modeling & AdaBoost & Characterization nanomaterial properties \\
		\hline
	\end{tabular}
\end{table*}

\subsubsection{Bagging}\label{SSS:Bagging}
Bootstrapping methods have been used extensively to minimize statistical errors of predictors by utilizing random sampling with replacement. In supervised learning, a training dataset is utilized to train a predictor. Bootstrap replicas of the training dataset can be employed to generate new predictors. \textit{bagging} is a meta-learning algorithm that uses this idea to develop an aggregated predictor, either by averaging the predictors over the learning sets when the exit is numerical or by voting, when the exit is a class label \cite{Breiman1996}. More specifically, assuming a learning set $\mathcal{L}$ consists of data $\{(y_n,\mathbf{x}_n),n=1,\dots,N\}$ and a predictor $\phi(\mathbf{x}, \mathcal{L})$, $y$ is predicted by $\phi(\mathbf{x},\mathcal{L})$ if the input is $\mathbf{x}$. The learning set consists of $N$ observations and since it is hard or in many cases impossible to obtain more observations to improve the learning set, we turn to bootstrapping, creating different learning sets using the sample $N$ as the population, which effectively leads to new predictors ($\{\phi(\mathbf{x},\mathcal{L})\}$). The aggregated predictor's accuracy is determined by the stability of the procedure for constructing each $\phi$ predictor, i.e., the accuracy will be improved with bagging in unstable procedures, where small variation in the learning set leads to large changes in the~predictor.

Recently, bagging has been used to predict possible toxic effects caused by the exposure to nanomaterials in biological systems~\cite{Liu2013}. As a base predictor $\phi$, REPTree was used, which is a fast decision tree-based learning algorithm. It should be mentioned that the bagging algorithm offered the highest accuracy, in terms of correlation, between actual and predicted~results. 

\subsubsection{Bagged Tree}\label{SSS:BTs}
Bagging can be applied to any kind of model. By using bagged decision trees, it is possible to lower the bias by leaving the trees un-pruned. High variance and low bias is essential for bagging classifiers. The aggregate classifier can capitalize on this and provide an increase in accuracy. In \cite{Yekkala2017}, a bagged tree was used with great success in a ensemble classifier with particle swarm optimization (PSO) in order to predict heart disease.


\subsubsection{Naive Bayes Tree}\label{SSS:NBT}
A hybrid approach to learning, when many attributes are deemed relevant for a classification task, yet they are not sufficiently independent, is the \textit{NBTree}. NBTree consists in practice of a decision tree with Naive Bayes classifiers at the leaf nodes~\cite{Kohavi96}. Firstly, according to a utility function an attribute is split in the decision tree making process. If the utility is not sufficiently high, the node becomes a leaf and a Naive Bayes classifier is created at the node. NBTree can deal both with discrete data, by multi-way splits for all values, and with continuous data, by using a threshold~split. 

In~\cite{Jones2015}, NBTree was used among other learning methods as a way to predict the cytotoxicity of nanomaterials in biological systems. When leave-one-out cross validation was performed, NBTree achieved the best performence and achieved an accuracy of~$77.7\%$.

\subsubsection{Adaptive Boosting}\label{SSS:AdaBoost}
AdaBoost is a learning method that uses an ensemble of classifiers in order to improve accuracy~\cite{Freund1995,Freund1997}. Boosting is a technique that takes a set of weak learners --usually a decision tree classifier-- and combines them into a strong one. The procedure can be summarized as follows. A set of labeled training examples $\{(x_i,y_i)\}$, where $x_i$ is an observable quality and $y_i$ is the outcome, are given into a set of classifiers that are each assigned a weight. After every weak classifier has reached to a prediction, the boosting method combines all the weak hypotheses into a single prediction. AdaBoost does not need prior knowledge of the accuracies of the weak classifiers, instead, it adapts to the errors of the weak classifiers.  In essence, the weak classifiers are tweaked to better handle data that were mishandled by previous classifiers. In some cases, AdaBoost has shown to be less susceptible to over-fitting than other learning methods, however it is prone to noisy data and outliers due to its adaptive~nature. 

AdaBoost was one of the methods used in~\cite{Yekkala2017} in an ensemble classifier together with PSO to predict heart~disease. Moreover, AdaBoost was used in~\cite{Mallick2004} as a learning approach for particle detection in cryo-electron micrographs. Similarly, in~\cite{John2016}, it was used for characterizing and analyzing unique features and properties of nanomaterials and nanostructures.

\subsubsection{Random Forest}\label{SSS:RF}
\begin{figure}
	\centering
	\scalebox{.6}{
		
		\tikzset{every picture/.style={line width=0.75pt}} 
		
		\begin{tikzpicture}[x=0.75pt,y=0.75pt,yscale=-1,xscale=1]
		
		\draw    (141.5,133.5) -- (180.5,169.5) -- (202.5,211) ;
		\draw    (180.5,169.5) -- (174.74,186.59) -- (166.5,211) ;
		\draw    (141.5,133.5) -- (106.5,170.5) -- (126.5,210) ;
		\draw    (106.5,170.5) -- (85.5,210.5) ;
		\draw    (312.5,132.5) -- (351.5,168.5) -- (373.5,210) ;
		\draw    (351.5,168.5) -- (345.74,185.59) -- (337.5,210) ;
		\draw    (312.5,132.5) -- (277.5,169.5) -- (297.5,209) ;
		\draw    (277.5,169.5) -- (256.5,209.5) ;
		\draw    (520.5,129.5) -- (559.5,165.5) -- (581.5,207) ;
		\draw    (559.5,165.5) -- (553.74,182.59) -- (545.5,207) ;
		\draw    (520.5,129.5) -- (485.5,166.5) -- (505.5,206) ;
		\draw    (485.5,166.5) -- (464.5,206.5) ;
		\draw    (354.5,38) .. controls (402.02,36.02) and (432.39,99.71) .. (483.45,115.54) ;
		\draw [shift={(485,116)}, rotate = 196.09] [color={rgb, 255:red, 0; green, 0; blue, 0 }  ][line width=0.75]    (10.93,-3.29) .. controls (6.95,-1.4) and (3.31,-0.3) .. (0,0) .. controls (3.31,0.3) and (6.95,1.4) .. (10.93,3.29)   ;
		\draw    (268,41) .. controls (224.44,44.96) and (226.94,100.87) .. (168.78,115.57) ;
		\draw [shift={(167,116)}, rotate = 346.87] [color={rgb, 255:red, 0; green, 0; blue, 0 }  ][line width=0.75]    (10.93,-3.29) .. controls (6.95,-1.4) and (3.31,-0.3) .. (0,0) .. controls (3.31,0.3) and (6.95,1.4) .. (10.93,3.29)   ;
		\draw    (313,49) -- (313,109) ;
		\draw [shift={(313,111)}, rotate = 270] [color={rgb, 255:red, 0; green, 0; blue, 0 }  ][line width=0.75]    (10.93,-3.29) .. controls (6.95,-1.4) and (3.31,-0.3) .. (0,0) .. controls (3.31,0.3) and (6.95,1.4) .. (10.93,3.29)   ;
		\draw    (143,250) -- (143,287) -- (527,287) -- (527,245) ;
		\draw    (317,287) -- (317,249) ;
		\draw   (308,320.6) -- (312.25,320.6) -- (312.25,308) -- (320.75,308) -- (320.75,320.6) -- (325,320.6) -- (316.5,329) -- cycle ;
		\draw  [fill={rgb, 255:red, 245; green, 166; blue, 35 }  ,fill opacity=1 ] (133,133.5) .. controls (133,128.81) and (136.81,125) .. (141.5,125) .. controls (146.19,125) and (150,128.81) .. (150,133.5) .. controls (150,138.19) and (146.19,142) .. (141.5,142) .. controls (136.81,142) and (133,138.19) .. (133,133.5) -- cycle ;
		\draw  [fill={rgb, 255:red, 245; green, 166; blue, 35 }  ,fill opacity=1 ] (172,169.5) .. controls (172,164.81) and (175.81,161) .. (180.5,161) .. controls (185.19,161) and (189,164.81) .. (189,169.5) .. controls (189,174.19) and (185.19,178) .. (180.5,178) .. controls (175.81,178) and (172,174.19) .. (172,169.5) -- cycle ;
		\draw  [fill={rgb, 255:red, 245; green, 166; blue, 35 }  ,fill opacity=1 ] (194,211) .. controls (194,206.31) and (197.81,202.5) .. (202.5,202.5) .. controls (207.19,202.5) and (211,206.31) .. (211,211) .. controls (211,215.69) and (207.19,219.5) .. (202.5,219.5) .. controls (197.81,219.5) and (194,215.69) .. (194,211) -- cycle ;
		\draw  [fill={rgb, 255:red, 74; green, 144; blue, 226 }  ,fill opacity=1 ] (79,210.5) .. controls (79,205.81) and (82.81,202) .. (87.5,202) .. controls (92.19,202) and (96,205.81) .. (96,210.5) .. controls (96,215.19) and (92.19,219) .. (87.5,219) .. controls (82.81,219) and (79,215.19) .. (79,210.5) -- cycle ;
		\draw  [fill={rgb, 255:red, 74; green, 144; blue, 226 }  ,fill opacity=1 ] (100,170.5) .. controls (100,165.81) and (103.81,162) .. (108.5,162) .. controls (113.19,162) and (117,165.81) .. (117,170.5) .. controls (117,175.19) and (113.19,179) .. (108.5,179) .. controls (103.81,179) and (100,175.19) .. (100,170.5) -- cycle ;
		\draw  [fill={rgb, 255:red, 74; green, 144; blue, 226 }  ,fill opacity=1 ] (158,211) .. controls (158,206.31) and (161.81,202.5) .. (166.5,202.5) .. controls (171.19,202.5) and (175,206.31) .. (175,211) .. controls (175,215.69) and (171.19,219.5) .. (166.5,219.5) .. controls (161.81,219.5) and (158,215.69) .. (158,211) -- cycle ;
		\draw  [fill={rgb, 255:red, 74; green, 144; blue, 226 }  ,fill opacity=1 ] (120,210) .. controls (120,205.31) and (123.81,201.5) .. (128.5,201.5) .. controls (133.19,201.5) and (137,205.31) .. (137,210) .. controls (137,214.69) and (133.19,218.5) .. (128.5,218.5) .. controls (123.81,218.5) and (120,214.69) .. (120,210) -- cycle ;
		\draw  [fill={rgb, 255:red, 245; green, 166; blue, 35 }  ,fill opacity=1 ] (304,132.5) .. controls (304,127.81) and (307.81,124) .. (312.5,124) .. controls (317.19,124) and (321,127.81) .. (321,132.5) .. controls (321,137.19) and (317.19,141) .. (312.5,141) .. controls (307.81,141) and (304,137.19) .. (304,132.5) -- cycle ;
		\draw  [fill={rgb, 255:red, 245; green, 166; blue, 35 }  ,fill opacity=1 ] (343,168.5) .. controls (343,163.81) and (346.81,160) .. (351.5,160) .. controls (356.19,160) and (360,163.81) .. (360,168.5) .. controls (360,173.19) and (356.19,177) .. (351.5,177) .. controls (346.81,177) and (343,173.19) .. (343,168.5) -- cycle ;
		\draw  [fill={rgb, 255:red, 74; green, 144; blue, 226 }  ,fill opacity=1 ] (366,210) .. controls (366,205.31) and (369.81,201.5) .. (374.5,201.5) .. controls (379.19,201.5) and (383,205.31) .. (383,210) .. controls (383,214.69) and (379.19,218.5) .. (374.5,218.5) .. controls (369.81,218.5) and (366,214.69) .. (366,210) -- cycle ;
		\draw  [fill={rgb, 255:red, 74; green, 144; blue, 226 }  ,fill opacity=1 ] (249,209.5) .. controls (249,204.81) and (252.81,201) .. (257.5,201) .. controls (262.19,201) and (266,204.81) .. (266,209.5) .. controls (266,214.19) and (262.19,218) .. (257.5,218) .. controls (252.81,218) and (249,214.19) .. (249,209.5) -- cycle ;
		\draw  [fill={rgb, 255:red, 74; green, 144; blue, 226 }  ,fill opacity=1 ] (270,169.5) .. controls (270,164.81) and (273.81,161) .. (278.5,161) .. controls (283.19,161) and (287,164.81) .. (287,169.5) .. controls (287,174.19) and (283.19,178) .. (278.5,178) .. controls (273.81,178) and (270,174.19) .. (270,169.5) -- cycle ;
		\draw  [fill={rgb, 255:red, 245; green, 166; blue, 35 }  ,fill opacity=1 ] (329,210) .. controls (329,205.31) and (332.81,201.5) .. (337.5,201.5) .. controls (342.19,201.5) and (346,205.31) .. (346,210) .. controls (346,214.69) and (342.19,218.5) .. (337.5,218.5) .. controls (332.81,218.5) and (329,214.69) .. (329,210) -- cycle ;
		\draw  [fill={rgb, 255:red, 74; green, 144; blue, 226 }  ,fill opacity=1 ] (290,209) .. controls (290,204.31) and (293.81,200.5) .. (298.5,200.5) .. controls (303.19,200.5) and (307,204.31) .. (307,209) .. controls (307,213.69) and (303.19,217.5) .. (298.5,217.5) .. controls (293.81,217.5) and (290,213.69) .. (290,209) -- cycle ;
		\draw  [fill={rgb, 255:red, 245; green, 166; blue, 35 }  ,fill opacity=1 ] (512,129.5) .. controls (512,124.81) and (515.81,121) .. (520.5,121) .. controls (525.19,121) and (529,124.81) .. (529,129.5) .. controls (529,134.19) and (525.19,138) .. (520.5,138) .. controls (515.81,138) and (512,134.19) .. (512,129.5) -- cycle ;
		\draw  [fill={rgb, 255:red, 74; green, 144; blue, 226 }  ,fill opacity=1 ] (551,165.5) .. controls (551,160.81) and (554.81,157) .. (559.5,157) .. controls (564.19,157) and (568,160.81) .. (568,165.5) .. controls (568,170.19) and (564.19,174) .. (559.5,174) .. controls (554.81,174) and (551,170.19) .. (551,165.5) -- cycle ;
		\draw  [fill={rgb, 255:red, 74; green, 144; blue, 226 }  ,fill opacity=1 ] (573,207) .. controls (573,202.31) and (576.81,198.5) .. (581.5,198.5) .. controls (586.19,198.5) and (590,202.31) .. (590,207) .. controls (590,211.69) and (586.19,215.5) .. (581.5,215.5) .. controls (576.81,215.5) and (573,211.69) .. (573,207) -- cycle ;
		\draw  [fill={rgb, 255:red, 74; green, 144; blue, 226 }  ,fill opacity=1 ] (457,206.5) .. controls (457,201.81) and (460.81,198) .. (465.5,198) .. controls (470.19,198) and (474,201.81) .. (474,206.5) .. controls (474,211.19) and (470.19,215) .. (465.5,215) .. controls (460.81,215) and (457,211.19) .. (457,206.5) -- cycle ;
		\draw  [fill={rgb, 255:red, 245; green, 166; blue, 35 }  ,fill opacity=1 ] (477,166.5) .. controls (477,161.81) and (480.81,158) .. (485.5,158) .. controls (490.19,158) and (494,161.81) .. (494,166.5) .. controls (494,171.19) and (490.19,175) .. (485.5,175) .. controls (480.81,175) and (477,171.19) .. (477,166.5) -- cycle ;
		\draw  [fill={rgb, 255:red, 74; green, 144; blue, 226 }  ,fill opacity=1 ] (538,207) .. controls (538,202.31) and (541.81,198.5) .. (546.5,198.5) .. controls (551.19,198.5) and (555,202.31) .. (555,207) .. controls (555,211.69) and (551.19,215.5) .. (546.5,215.5) .. controls (541.81,215.5) and (538,211.69) .. (538,207) -- cycle ;
		\draw  [fill={rgb, 255:red, 245; green, 166; blue, 35 }  ,fill opacity=1 ] (497,206) .. controls (497,201.31) and (500.81,197.5) .. (505.5,197.5) .. controls (510.19,197.5) and (514,201.31) .. (514,206) .. controls (514,210.69) and (510.19,214.5) .. (505.5,214.5) .. controls (500.81,214.5) and (497,210.69) .. (497,206) -- cycle ;
		
		\draw (288,25) node [anchor=north west][inner sep=0.75pt]   [align=left] {{\fontfamily{ptm}\selectfont Instance}};
		\draw (121,228) node [anchor=north west][inner sep=0.75pt]   [align=left] {Tree \#1};
		\draw (290,226) node [anchor=north west][inner sep=0.75pt]   [align=left] {Tree \#2};
		\draw (501,224) node [anchor=north west][inner sep=0.75pt]   [align=left] {Tree \#N};
		\draw (404,148) node [anchor=north west][inner sep=0.75pt]   [align=left] {\textbf{{\Large ...}}};
		\draw  [fill={rgb, 255:red, 255; green, 255; blue, 255 }  ,fill opacity=1 ]  (265.5,271.5) -- (368.5,271.5) -- (368.5,302.5) -- (265.5,302.5) -- cycle  ;
		\draw (317,287) node   [align=left] {\begin{minipage}[lt]{67.405pt}\setlength\topsep{0pt}
			{\fontfamily{ptm}\selectfont Majority Voting}
			\end{minipage}};
		\draw  [fill={rgb, 255:red, 255; green, 255; blue, 255 }  ,fill opacity=1 ]  (278.5,331.5) -- (355.5,331.5) -- (355.5,362.5) -- (278.5,362.5) -- cycle  ;
		\draw (317,347) node   [align=left] {\begin{minipage}[lt]{49.64pt}\setlength\topsep{0pt}
			{\fontfamily{ptm}\selectfont Final Result}
			\end{minipage}};

		\end{tikzpicture}
	}
	\caption{Random forest diagram.}
	\label{fig:Random_forest}
\end{figure}
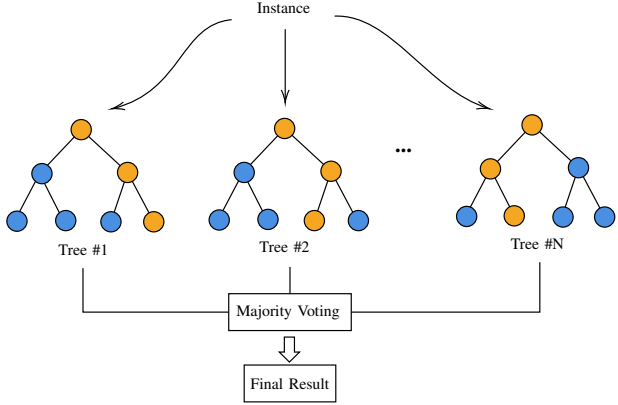

\textit{RForest} is one of the one of the most used ML algorithms, due to its simplicity and diversity, since it can be used for both classification and regression. As the name suggests, a RForest is a tree-based ensemble, where  each tree is connected to a collection of random variables~\cite{EnsembleML}. In Fig.~\ref{fig:Random_forest}, RForest average multiple decision trees are presented, that have been trained on different parts of the same training set, in order to reduce the variance. The different decision trees are trained based on the bagging technique, thus they exploit the random subsets of the training data. An advantage of RForest is that it decreases the variance of the model and, thus, combines uncorrelated individual trees with bagging, making them more robust without increasing the bias to overfitting. Another technique for combining individual trees is boosting, where the samples are weighted for sampling so that samples, which were predicted incorrectly, get a higher weight and are therefore, sampled more often. The concept behind this is that difficult cases should be emphasized during learning, compared to easy ones. Because of this difference, bagging can be easily paralleled, while boosting is performed sequentially. Next, we provide briefly the mathematical concept behind the RForest method. 

We assume an unknown joint distribution $P_{XY (X,Y)}$, where $X=\left(X_{1}, \ldots, X_{p}\right)^{T}$ is a $p$-dimensional random vector, which represents the  predictor variables and $Y$ is  the real-valued response.  The aim of the RForest algorithm is to find a prediction function $f(X)$ in order to predict $Y$. The prediction function is that which minimizes the expected value of the loss function $L(Y, f(X))$, i.e.
$E_{X Y}(L(Y, f(X))),$
where the subscripts denote expectation with respect to the joint distribution of $X$ and $Y$.

Note that $L(Y, f(X))$ is a measure of how close $f(X)$ is to $Y$ and it penalizes values of $f(X)$ that are far from $Y$. Typical choices of $L$ are squared error loss $L(Y, f(X))=(Y-f(X))^{2}$ for regression and zero-one loss for classification:
\begin{equation}
L(Y, f(X))=I(Y \neq f(X))=\left\{\begin{array}{l}
0 \text { if } Y=f(X) \\
1 \text { otherwise. }
\end{array}\right.
\end{equation}

It turns out that minimizing $E_{X Y}(L(Y, f(X)))$ for squared error loss gives the conditional expectation $f(x)=E(Y \mid X=x)$, which is known as the regression function. When classification is considered, if the set of possible values of $Y$ is denoted by $\mathcal{Y}$, then minimizing $E_{X Y}(L(Y, f(X)))$ for zero-one loss results to 
\begin{equation}
f(x)=\arg \max _{y \in \mathcal{Y}} P(Y=y \mid X=x)
\end{equation}
which is the Bayes rule.

Ensembles construct $f$ in terms of the so-called ``base learners'' $h_1 (x),...,h_J (x)$ and these  are combined to give the ``ensemble predictor'' $f(x)$. In regression, the base learners are averaged
\begin{equation}
f(x)=\frac{1}{J} \sum_{j=1}^{J} h_{j}(x)
\end{equation}
while in classification, $f(x)$ is the most frequently predicted class
\begin{equation}
f(x)=\arg \max _{y \in \mathcal{Y}} \sum_{j=1}^{J} I\left(y=h_{j}(x)\right)
\end{equation}

In RForests the $j_{th}$ base learner is a tree denoted as $h_{j}\left(X, \Theta_{j}\right)$, where $\Theta_{j}, j = 1,...,J.$ is a collection of independent random variables.  To deeply understand the RForest algorithm, a fundamental knowledge of the type of trees used as base learners is needed.

\subsubsection{M5P}\label{SSS:M5P}

The M5 model tree method was introduced by Quinlan in $1992$~\cite{Quinlan1992}.   Wang and Witten later presented an improved public-domain scheme \cite{Wang97}, called M5P, that generates more compact and comprehensible models with slightly better accuracy. M5P combines conventional binary decision tree models with regression planes at the leaves, to provide a way to deal with continuous-class problems. The initial tree split is based on a standard deviation criterion, called \textit{standard deviation reduction (SDR)} and  given by
\begin{equation}
\mathrm{SDR} = \mathrm{SD}(A) - \sum_i \frac{|T_i|}{|T|} \mathrm{SD}(T),
\end{equation}
where $\mathrm{SD}(A)$ is the standard deviation of  the set $A$, $T$ is the set of learning examples that reach the node, and $\{T_i\}$ are the subsets that result from splitting $T$ according to a chosen attribute. The attribute that maximizes SDR is the chosen for the split. 
However, this process can lead to large tree structures that are prone to over-fitting. Therefore, pruning the tree is necessary to improve accuracy. For every interior node of the tree, a regression model is calculated with the examples that reach that node, if the subtree error is greater than the respective error of the regression model in that node, the tree is pruned and that particular node is turned into a leaf node. 

Recently, M5P was used in~\cite{Liu2013} to built a simulator that can dynamically predict the mortality rate of cells in biological systems in order to test possible toxic effects from exposure to nano-materials. The simulator's user can change the attribute values dynamically and obtain the predicted value of the used metric.

\subsection{Decision Table}\label{SS:DT}
A~\textit{DT} is a simple tabular representation of conditions and actions~\cite{Hall2008}. It is very similar to the  popular decision trees. A key difference between among them is that the former can include more than one ``OR'' condition. However, DTs are usually preferred when a small number of features is available, whereas decision trees can be used for more complex~models.

\subsubsection*{Decision Table Naive Bayes}
Combined learning models is an efficient way to improve the accuracy of stand-alone models. \textit{DT Naive Bayes (DTNB) } is such a hybrid model, where a DT classifier is combined with a naive Bayes network, to produce a table with conditional probabilities. The learning process for DTNB splits the training data into two disjoint subsets and utilizes one set for training the DT and the other for training the NB~\cite{Hall2008}. The goal is to use NB on the attributes that are somewhat independent, since NB already assumed independence of attributes. Cross validation methods are suitable in this hybrid model since it is effective in both DTs, due to the structure of the table remaining the same, and the NB as the frequency counts can be updated in constant~time. 

Assuming that $\mathbf{x}^\mathrm{DT}$ is the set of attributes used in DT, and $\mathbf{x}^\mathrm{NB}$ is the respective set of attributes for NB, the class $k$ probability can be computed as
\begin{equation}
P(C_k|\mathbf{x}) =  \frac{a P(C_k|\mathbf{x}^\mathrm{DT})P(C_k|\mathbf{x}^\mathrm{NB})}{P(C_k)},
\end{equation}
where $a$ is a normalization constant and $P(C_k)$ is the prior probability of the class. DTNB is shown to achieve significant gains over both DTs and NB. More specifically, in~\cite{Jones2015}, DTNB was used among other methods to predict cytotoxicity values of nanomaterials in biological~systems.

\subsection{Surrogate-Based Optimization}\label{SS:SBO}
\textit{Surrogate-based optimization }\cite{Queipo2005,Zhang2012} refers to a class of optimization methodologies, that calculate the local or global optima by utilizing surrogate modeling techniques. This framework utilizes conventional optimization algorithms, such as gradient-based or evolutionary algorithms, for sub-optimization. Surrogate modeling techniques can significantly improve the design efficiency and facilitate finding global optima, filtering numerical noise, accomplishing parallel design optimization and integrating simulation codes of different disciplines into a process chain. 

In optimization problems, surrogate models can approximate the cost functions and the state functions, constructed from sampled data which are obtained by randomly exploring the design space. After this step, a new design based on the surrogate models, which is most likely to be the optimum, is searched by applying an optimization algorithm such as Genetic Algorithms. Utilizing a surrogate model for the estimation of the optimum is more effective than using a numerical analysis code, thus, the computational cost of the search based on the surrogate models is negligible. Surrogate models are built from the sampled data, thus the way the sample points are chosen and the way the accuracy of surrogate models is evaluated are important issues for surrogate modeling.

In \cite{Tran2018}, surrogate-based optimization is used to search the space of intermetallics for potentially selective catalysts for $\text{CO}_2$ reduction reaction and hydrogen evolution reaction.

\subsection{Quantitative Structure-Activity Relationships}\label{SS:QSAR}
ML techniques have been combined with QSARs models over the past decade~\cite{Winkler2014}. One of the most successful applications of such models is the development of new drugs faster and with lower cost. QSAR methods are data-driven and based on supervised learning. They capture the complex relationships between the properties of nanomaterials without requiring detailed knowledge of the mechanisms of interaction. In more detail, every biological activity of organic molecules is a function of their structural properties that depend on their chemical structures. These relationships can be expressed as in~\cite{Winkler2014}
\begin{align}
\text{Activity} = f\left(\sum\left(\text{Properties}\right)\right) ,
\end{align}
and
\begin{align}
\text{Property} = f\left(\text{Structure}\right) .
\end{align}

Due to the complexity of the materials the predictivity of the applied methods must be optimized, thus various different techniques have been used in the literature. Specifically, in~\cite{Burden2009}, QSAR models were developed based on sparse linear FS and regression in conjunction with a minimization algorithm, while, in~\cite{Burden1999,Winkler2000,Burden2009a}, nonlinear FS was used with Bayesian regularized NNs that used Gaussian or Laplacian priors. Also, ANNs have been recently employed to forecast the biological activity of compounds under investigation, while the ANN-classification model categorizes the compounds for a specific biological response~\cite{Gupta2016}.

\subsection{Boltzmann Generator}\label{SS:BG}
The aim of statistical mechanics is to assess the average behavior of physical systems based on their microscopic constituents and  interactions, in order not only to understand the molecules and materials functionalities, but also provide the principles for devising drug molecules and materials with novel properties. In this direction, the statistics of the equilibrium states of many-body systems needs to be evaluated. To conceive the complexity of this, let us try to evaluate the probability that, at a given temperature, a protein will be folded. In order to solve this problem, we need to examine each one of the huge number of ways to place all the proteins in a predetermined space and for each one of them extract the corresponding probability. However, since the enumeration of all configurations is extremely difficult or even infeasible, the necessity to sample them from their equilibrium distribution has been identified in ~\cite{Noe2019}. In this work, the authors proposed the Boltzmann generator, which  combines deep ML and statistical mechanics in order to learn sample equilibrium distributions. In contrast to conventional generative learning, the Boltzmann generator is not trained to learn the probability density from data, but to directly produce independent samples of low-energy structures for condensed-matter systems and protein molecules. 

\begin{figure}
	\centering\includegraphics[width=1\linewidth,trim=0 0 0 0,clip=false]{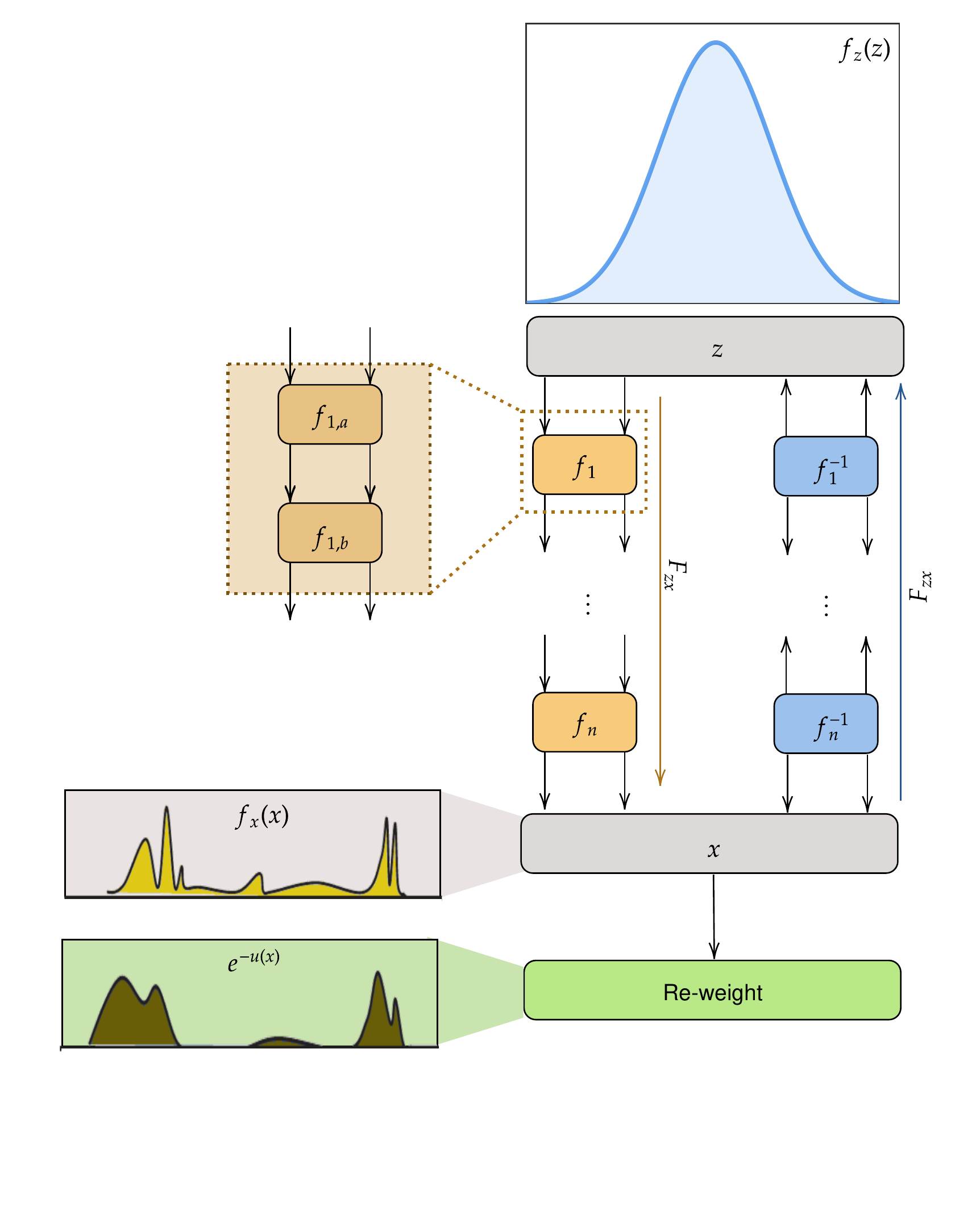}
	\vspace{-1.5cm}
	\caption{Boltzmann generator.}
	\label{Fig:BG}
\end{figure}

As presented in Fig.~\ref{Fig:BG}, the operation principle of Boltzmann generator consists of two parts: 
\begin{enumerate}
	\item A generative model, $F_{zx}$, is trained capable of providing samples from a stochastic distribution, which is described by the probability density function (PDF), $f_{x}(x)$, when sampling $z$ from a simple prior, such as a Gaussian distribution with PDF $f_{z}(z)$.
	\item A re-weighting process that transforms the generated distribution, $f_x(x)$, into the Boltzmann distribution, and produces unbiased samples from the $e^{-u(x)}$, with $u(x)$ being the dimensionless energy.
\end{enumerate}  
Note that both training and re-weighting require $f_{x}(x)$ knowledge. This can be ensured by adopting an invertible $F_{zx}$ transformation, which allows us to transform $f_{z}(z)$ to $f_{x}(x)$.  

\subsection{Feedback System Control}\label{SS:FSC}
\textit{FSC} \cite{Nowak2016} is a recently proposed method for the optimization of drug combinations. FSC is a  phenotypically driven optimization process, which does not require any mechanistic knowledge for the system. This is the reason that FSC can be successfully applied in various complex biological systems (see \cite{Liu2018} and references therein)

\begin{figure}
	\centering
	\includegraphics[width=1.0\columnwidth]{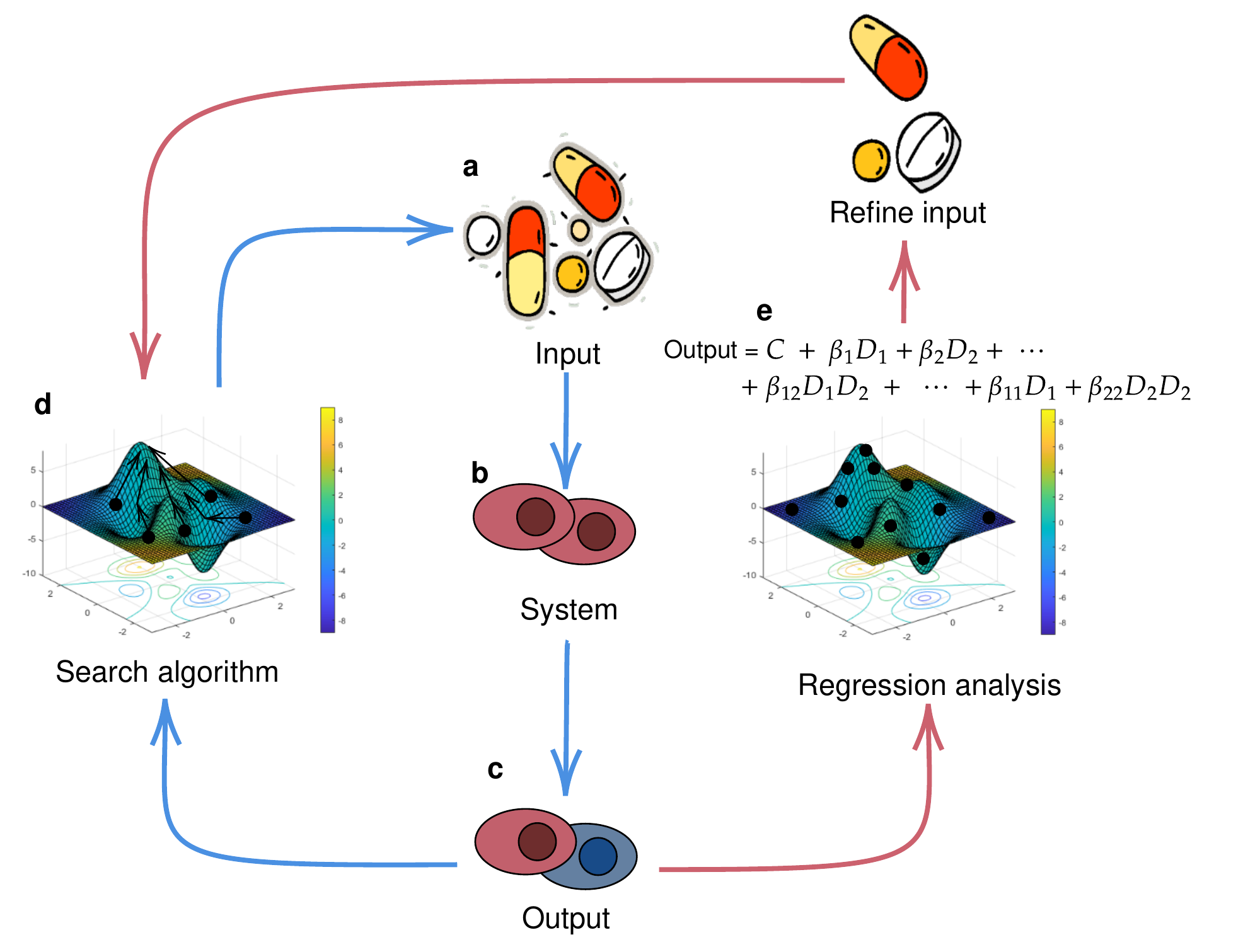}
	\caption{Examples of classical and ML-SFs.} \label{FSC}
\end{figure}

The FSC method is based on the closed-loop feedback control process outlined in Fig.~\ref{FSC} \cite{Nowak2016}. It mainly consists of two steps: the first step is the definition of an initial set of compounds to be tested. The second step refers to the generation of broad dose-response curves for each compound in the selected cellular bioassay, which is selected to provide a phenotypic output response, that is used to evaluate the efficacy of the drugs and drug combinations on overall cell activity. 

A schematic representation of the FSC technique is presented in Fig. \ref{FSC}. The five main components of the optimization process are depicted as: 

(a) The input, i.e., the drug combinations with defined drug doses. 

(b) The system, i.e., the selected cell type representation of the disease to be studied 

(c) The system output, i.e., the cellular response to the defined drug combination input in the selected cell bioassay. 

(d) The search algorithm that iteratively drives the system output toward the desired response. 

(e) The statistical analysis used to guide drug elimination.

\subsection{Quadratic Phenotypic Optimization Platform}\label{SS:QPOP}
Methods based on ML, like FSC, aim to overcome the disadvantages of the traditional methods, as for example the high-throughput screening. 
Recently, a powerful AI platform called Quadratic Phenotypic Optimization Platform (QPOP) was proposed,  to interrogate a large pool of potential drugs  and to design a novel combination therapy against multiple myeloma \cite{Rashid2018}. This platform can 
efficiently and iteratively outputs effective drug combinations and can optimize the drug doses. 

The main concept of QPOP lies in recognizing the relationship between inputs (e.g., drugs) and desired phenotypic outputs (e.g., cell viabilities) to a smooth, second-order quadratic surface representative of the biological system of interest. Since QPOP utilizes only the controllable inputs and measurable phenotypic outputs of the biological system, it is able to identify optimal drug combinations and doses independently of predetermined drug synergy information and pharmacokinetic properties. Furthermore, QPOP utilized ML  in order to  preclinically re-optimize the combination and successfully translate the multi-drug regimen through in vivo validation. It is important to mention that both the in vitro and preclinical re-optimization processes were able to simultaneously take into account both efficacy and safety, and this is an important aspect of the QPOP  platform. 

QPOP can also be used as an actionable platform to design patient-specific regimens. This multi-parametric global optimization methodology can  overcome many of the drug development process difficulties, and can result in efficient and safe therapies. This will revisit the drug development, translating into improved and effective treatment choices.

More details about the use of the QPQP platform in biomedicine applications can be found in \cite{Ho2019} and \cite{Khong2020} and references therein.

\section{Discussion \& The Road Ahead}\label{S:Discussion}
In this section, we clarify how the ML methodologies presented in Section~\ref{S:Approaches} can be efficiently used to solve the problems discussed in Section~\ref{S:Problems}  and elaborate on some major open research problems, which are of great importance for unveiling the potential benefits, advantages, and limitations of employing ML in nano-scale biomedical engineering. In this direction, Table~\ref{T:ML_approaches_ML_challenges}, which is given in the top of the next page, connects the ML challenges with the ML methodologies, that have been used in nano-scale biomedical~engineering. 

\begin{table*}[]
	\centering
	\caption{ML problems and solutions. }
	\label{T:ML_approaches_ML_challenges}
	\begin{tabular}{|p{5.5cm}||c|c|c|c|}
		\hline
		\multicolumn{1}{|c||}{\multirow{2}{*}{\textbf{ML approaches}}} & \multicolumn{3}{c|}{\textbf{ML challenge categories}}                                                                                                                                                                 \\ \cline{2-4} 
		\multicolumn{1}{|c||}{} & \multicolumn{1}{p{2.5cm}|}{\centering Structure and material design and simulation} & \multicolumn{1}{p{2.5cm}|}{\centering Communications and signal processing} &  \multicolumn{1}{p{2.5cm}|}{\centering Applications} \\ \hline \hline
		 \cline{1-4}  \multicolumn{4}{|c|}{\textbf{ANNs}}  \\ \hline \hline
		 Convolution neural networks	&  \checkmark & \checkmark &   \checkmark \\ \hline
		 Recurrent neural networks & \checkmark & \checkmark &  \checkmark \\ \hline 
		 Deep neural networks	&  \checkmark & \checkmark &   \checkmark \\ \hline
		 Diffractive deep neural networks  & \checkmark & -  & - \\ \hline  
		 Generalized regression neural networks  & \checkmark & -  & \checkmark\\ \hline 
		 Multi-layer perceptron  & \checkmark & \checkmark  & - \\ \hline 
		 Generative adversarial networks & \checkmark & -  & \checkmark \\ \hline
		 Behler-Parrinello networks  & \checkmark & -  & - \\ \hline
		 Deep potential networks & \checkmark & -  & - \\ \hline 
		 Deep tensor neural networks & \checkmark & -  & - \\ \hline 
		 SchNet & \checkmark & -  & - \\ \hline  
		 Accurate neural network engine for molecular energies & \checkmark & - &  - \\ \hline  
		 Coarse graining & \checkmark & -  & - \\ \hline 
		 Neuromorphic computing & \checkmark & -  & - \\ \hline \hline
		 \cline{1-4}  \multicolumn{4}{|c|}{\textbf{Regression}}  \\ \hline \hline
		 Logistic regression & \checkmark & \checkmark &  \checkmark \\ \hline 
		 Multivariate linear regression & \checkmark  & -  & - \\ \hline 
		 Classification via regression & \checkmark & -  & - \\ \hline 
		 Local weighted learning &\checkmark & -  & -\\ \hline
		 Machine learning scoring functions & \checkmark & -  & - \\ \hline
		 \hline
		 \cline{1-4}  \multicolumn{4}{|c|}{\textbf{Support vector machine}}  \\ \hline \hline
		Support vector machine	& \checkmark & \checkmark    & \checkmark  \\ \hline \hline
		\cline{1-4}  \multicolumn{4}{|c|}{\textbf{$k$-nearest neighbors}}  \\ \hline \hline
		$k$-nearest neighbors	& \checkmark & \checkmark &   -   \\ \hline
		\hline 
		\cline{1-4}  \multicolumn{4}{|c|}{\textbf{Dimentionality reduction}}  \\ \hline \hline
		Feature	selection & \checkmark & -   &  - \\ \hline
		Principle component analysis & \checkmark & -  & \checkmark   \\ \hline
		Linear discriminant analysis & \checkmark & -  & \checkmark \\ \hline
		Independent component analysis &  \checkmark & -  & \checkmark \\ \hline
		\hline \hline 
		\cline{1-4}  \multicolumn{4}{|c|}{\textbf{Gradient descent}}  \\ \hline \hline
		Gradient descent & \checkmark & -  & - \\ \hline \hline 
		\cline{1-4}  \multicolumn{4}{|c|}{\textbf{Active learning}}  \\ \hline \hline
		Active learning & \checkmark & -  & - \\ \hline \hline 
		Bayesian ML & \checkmark &  \checkmark & \checkmark \\ \hline \hline 
		\cline{1-4}  \multicolumn{4}{|c|}{\textbf{Decision tree learning}}  \\ \hline \hline
		Bagging &  \checkmark & \checkmark & - \\ \hline
		Bagged tree &  - & -  & \checkmark \\ \hline
		Naive Bayer tree &  \checkmark & -  & \checkmark  \\ \hline
		Adaptive boosting &  -  & - & \checkmark \\ \hline
		Random forest & -   & - & \checkmark \\ \hline
		M5P & \checkmark & -  & - \\ \hline \hline 
		\cline{1-4}  \multicolumn{4}{|c|}{\textbf{Decision table}}  \\ \hline \hline
		Decision table naive Bayes &  \checkmark  & \checkmark & -  \\ \hline\hline 
		\cline{1-4}  \multicolumn{4}{|c|}{\textbf{Surrogate-based optimization}}  \\ \hline \hline
		Surrogate-based optimization & \checkmark & -  & - \\ \hline
		\cline{1-4}  \multicolumn{4}{|c|}{\textbf{QSAR}}  \\ \hline \hline
		QSAR & \checkmark & -  & \checkmark  \\ \hline\hline
		\cline{1-4}  \multicolumn{4}{|c|}{\textbf{Boltzmann generator}}  \\ \hline \hline
		Boltzmann generator& \checkmark & -  & - \\ \hline  \hline
		\cline{1-4}  \multicolumn{4}{|c|}{\textbf{Feedback system control}}  \\ \hline \hline
		Feedback system control & -  & - & \checkmark \\ \hline  
		\hline
		\cline{1-4}  \multicolumn{4}{|c|}{\textbf{Quadratic phenotypic optimization platform}}  \\ \hline \hline
		Quadratic phenotypic optimization platform & \checkmark & -  & \checkmark  \\ \hline 	
	\end{tabular}
\end{table*}

From Table~\ref{T:ML_approaches_ML_challenges}, it becomes evident that ANNs can be employed to solve a large variety of ML problems in nano-scale biomedical engineering. The ML methods CNNs, RNNs, and DNNs are capable of identifying patterns, locate and classify target objects in an image, and detect events~\cite{Liu2017}. As a result, they can excel in the development of ARES, which contributes to the discovery, design, and performance optimization of nano-structures and nano-materials. Furthermore, they can be used for the detection of received symbols in molecular and electromagnetic nano-networks, for the classification of observations that may provide a better understanding of biological and chemical processes, and for the identification of specific patterns. On the other hand,  D$^2$NNs  can efficiently execute identification and classification tasks, after being trained by large datasets. Therefore, they have been successfully used in lens imaging at THz spectrum, while they are expected to find application in image analysis, feature detection, and object classification. In other words, D$^2$NNs may be employed for heterogeneous nano-structures discovery, channel estimation and symbol detection in nano-scale molecular and THz networks, as well as disease detection and therapy development.

By inducing the algorithm to learn complex relationships within a training dataset and making judgments on test  datasets with high fidelity, GRNNs are capable of providing a systematic methodology to map inputs to predictive outputs. As a consequence, they have been applied in several fields, including optical character recognition, pattern recognition, and manufacturing for predicting the output classification~\cite{Xiang2014,Almakaeel2018}. In nano-scale biomedical engineering, they have been extensively used in discovering the properties of and designing heterogeneous nano-structures~\cite{Almakaeel2018,Akter2018} as well as analyzing the data collected from them~\cite{RosarioMartinezBlanco2016}. However, their applicability in molecular and electromagnetic nano-scale networks specific problems needs to be assessed. 

Based on Cybernko's theorem~\cite{Cybenko1989}, MLPs are proven to be universal function approximators. In other words, they return low-complexity approximating solutions  from extremely complex problems. As a result, MLPs have been a popular ML method in 80s in several fields including speech and image recognition (see e.g.,~\cite{Kuang1992,Tang2016} and references therein). In nano-scale biomedical engineering, MLPs have been applied for nano-structure properties discovery~\cite{Bayat2018,Guo2017} and data analysis~\cite{Park2014}. However, it is expected to be replaced by much simpler SVMs, which are considered their main competitors. 

GANs have been recently used to inversely design metasurfaces in order to provide arbitrary patterns of the unit cell structure~\cite{Radford2015}. However, they experience high instability. To solve this problem conditional deep convolutional  GANs are usually employed. These networks return very stable Nash equilibrium solutions that can be used for inversely designing nanophotonic structures~\cite{So2019,GayonLombardo2020}.  Another application of GANs lies in the statistical characterization of psychological wellness states~\cite{Park2014}. In general, for applications in which the data have a non-linear behavior, GANs achieve similar performance as SVMs and $k-$nearest neighbor, and outperform~MLPs.  

Classical force field theory can neither easily scale into large molecules nor become transferable to different environments. To break these limitations, BPMs,  DPNs, DTNNs, SchNets, and CGNs have been traditionally used to model the PESs and atomic forces in large molecules, like proteins and provide transferability to different covalent and non-covalent environments. However, these approaches are incapable of reaching the required accuracy with lower than classical force field evaluation complexity. Motivated by this, symmetrized gradient-domain ML have been very recently presented as a possible solution to the aforementioned problem~\cite{Chmiela2018,Chmiela2017,Chmiela2019,sauceda2020}. The limitation of this ML approach is that it cannot support molecules that consists of more than $20$ atoms. In other words, it lacks scalability and transferability. To countermeasure this, researchers should turn their eye in combining  BPMs,  DPNs, DTNNs, SchNets, and CGNs with gradient-domain ML in order to provide high-accuracy in configuration and chemical space simulations. A plethora of new insights awaits as a result of such simulations.        

Regression approaches have been used to extract the relationship between several independent variables and one dependent variable. Therefore, they have supported the solution of a large variety of problems that range from the area of nano-materials and nano-structure design to data-driven applications in biomedical engineering~\cite{Yamankurt2019,Mohamed2020}. Moreover,  they usually  require no input features or tuning for scaling and they are easy to regularize. However, it is incapable of solving non-linear problems. Another disadvantage of regression approaches is that they require the identification of all the important independent attributes before inserting the data into the machine. Moreover, most of them return discrete outputs, i.e., they only provide categorical outcomes. Finally, they are sensitive to overfitting~\cite{Menard2010}.  

Similarly to regression, SVMs are efficient methods for problems with high-dimensional spaces. Taking this into account, several researchers have adopted them in order to provide solutions to a large range of problems from heterogeneous structure design to signal detection in molecular communication systems and data-driven applications. However, as the data set size increases, SVMs may underperform. Another limitation that should be highlighted is that they are not suitable for problems with overlapping targeting classes~\cite{Mirjalili2019}. 

KNN has been employed in structure and material design~\cite{Nigsch2006}, MCs for symbol detection~\cite{Qian2019}, and disease detection~\cite{Junejo2019,Rani2020}. {It is a low-complexity approach suitable for classifying data without training.} However, it suffers from performance degradation when applied to large data sets, due to increased cost of computing the distance between the new point and each of the existing points. A similar performance degradation is observed as the dimensions of the data increase. This indicates that the application of KNN approach in heterogeneous nano-structure design is questionable. On the other hand, it excels in data sequence detection in MC systems, where the dimension of the data is no higher than $2$. 

Dimensionality reduction methods have been applied in the nano-structure and material design~\cite{Chen2018a,Lamoureux2020} as well as in therapy development~\cite{Chen2009}. Their objective is to remove dimensions, i.e. redundant features, in order to identify the more suitable variable for the problem under investigation. As a result, they contribute to  data compression and to computation time reduction. Moreover, they are capable of transforming multi-dimensional problems into two dimensional (2D) or 3D ones allowing their visualization. This property has been extensively used in nano-structure properties discovery. Likewise, dimensionality reduction methods can aid at noise removal; thus, they can significantly improve the model's performance. However, they come with some disadvantages. In particular, they cause data loss. Moreover, PCA tends to extract linear correlations between variables. In practice, most of the nano-structure properties have a non-linear behavior. As a result, PCA may return unrealistic results. This highlights the need of designing new dimensionality reduction methods that take into accounts the chemical and biological properties of the nano-structure components. Finally, dimensionality reduction methods traditionally fail in cases where the datasets cannot be fully defined by their mean and covariance. 

GD is an iterative ML optimization algorithm that aims at reducing the cost function in order to make accurate predictions; therefore, it has been employed in predicting the properties of heterogeneous nano-structures. Its main disadvantage is that the solution returned by this method is not guaranteed to be a global minimum. As a result, every time that  the search-space is expanded, due to  the incorporation of an additional parameter into the objective function, the surface of optimal solutions may exhibit numerous locally optimal solutions. Thus, conventional GD algorithms may return a non-global local optimum. In this context, examination of more sophisticated GD algorithms needs to be performed. Finally, GD may be seen as an attractive optimization tool for finding Pareto-optimal solutions of multi-objective optimization problems in nano-scale networks. Such problems would aim at minimizing the outage probability, power consumption and/or maximizing throughput, network lifetime and other parameters that improve the network's quality of~experience. 

DTL algorithms are able to solve both regression and classification problems. As a result, they have been extensively used in several fields including structure and material design and simulation as well as analyzing data acquired from nano-scale systems. Compared to other ML algorithms, decision tree and table learning algorithms simplify data preparation processes, since they demand neither data normalization nor scaling. Moreover, they perform well even when with incomplete data sets and their models are very intuitive and easy to explain. Therefore, several researchers have used them to provide comprehensive understanding of the properties of nano-structures and the relationship with their design parameters. However, DTL algorithms are sensitive to even small changes in the data. In more detail, a small change in the data may result in a significant change in the structure of the decision tree, which in turn may cause instability. Another disadvantage of decision trees and tables is that they require higher time to train the models and to perform after-training calculations. Finally, they are incapable for applying regression and predicting continuous values. These disadvantages render them unsuitable for use in real-time applications in the fields of communications and signal processing as well as in nano-scale networks. 

QSARs are mathematical models, which relate a pharmacological or biological activity with the physicochemical characteristics (termed molecular descriptors) of molecule sets. Indicative examples of QSAR applications are the study of enzyme activity~\cite{Liu2020}, the minimum effective dose of a drug estimation~\cite{Tomalia2020}, and toxicity prediction of nano-structures~\cite{Mukherjee2010}. The main advantage of QSAR models lies with their ability to predict activities of a large number of compounds with little to no prior experimental data. However, they are incapable of providing in-depth insights on the mechanism behind biological actions.

Boltzmann generators have been employed to create physically realistic one-shot samples of model systems and proteins in implicit solvent~\cite{Noe2019a,Ulanov2019}. Scaling to large systems, such as those investigated in MCs and nano-scale networks, needs to build the invariances of the energy, as the exchange of molecules, into the transformation to include parameter sharing. In other words, researchers need to develop equivariant networks with parameter sharing. These networks are expected to provide a better understanding of molecular channel modeling and eventually contribute to the design of new transmission~schemes. 

\section{Conclusion}\label{S:Conclusion}
In summary, in this article, we  have reviewed how ML algorithms bear fruit in nano-scale biomedical engineering. In more detail, we presented the main challenges and problems in this field, which, due to their high complexity, require the use of ML in order to be solved, and classified them, based on their discipline, into three distinctive categories. For each category, we have provided insightful discussions that revealed its particularities as well as existing research gaps.  Moreover, we have surveyed a variate of SOTA ML methodologies and models, which have been used as countermeasures to the aforementioned challenges. Special attention was payed to the ML methodologies architecture, operating principle, advantages and limitations. Finally, future research directions have been provided, which highlight the need of thorough interdisciplinary research efforts for the successful realization of hitherto uncharted scenarios and applications in the nano-scale biomedical engineering field.   

%
%
%
%
%

\balance
\bibliographystyle{IEEEtran}
\bibliography{IEEEabrv,refs}

\begin{IEEEbiography}[{\includegraphics[width=1in,height=1.25in,clip,keepaspectratio]{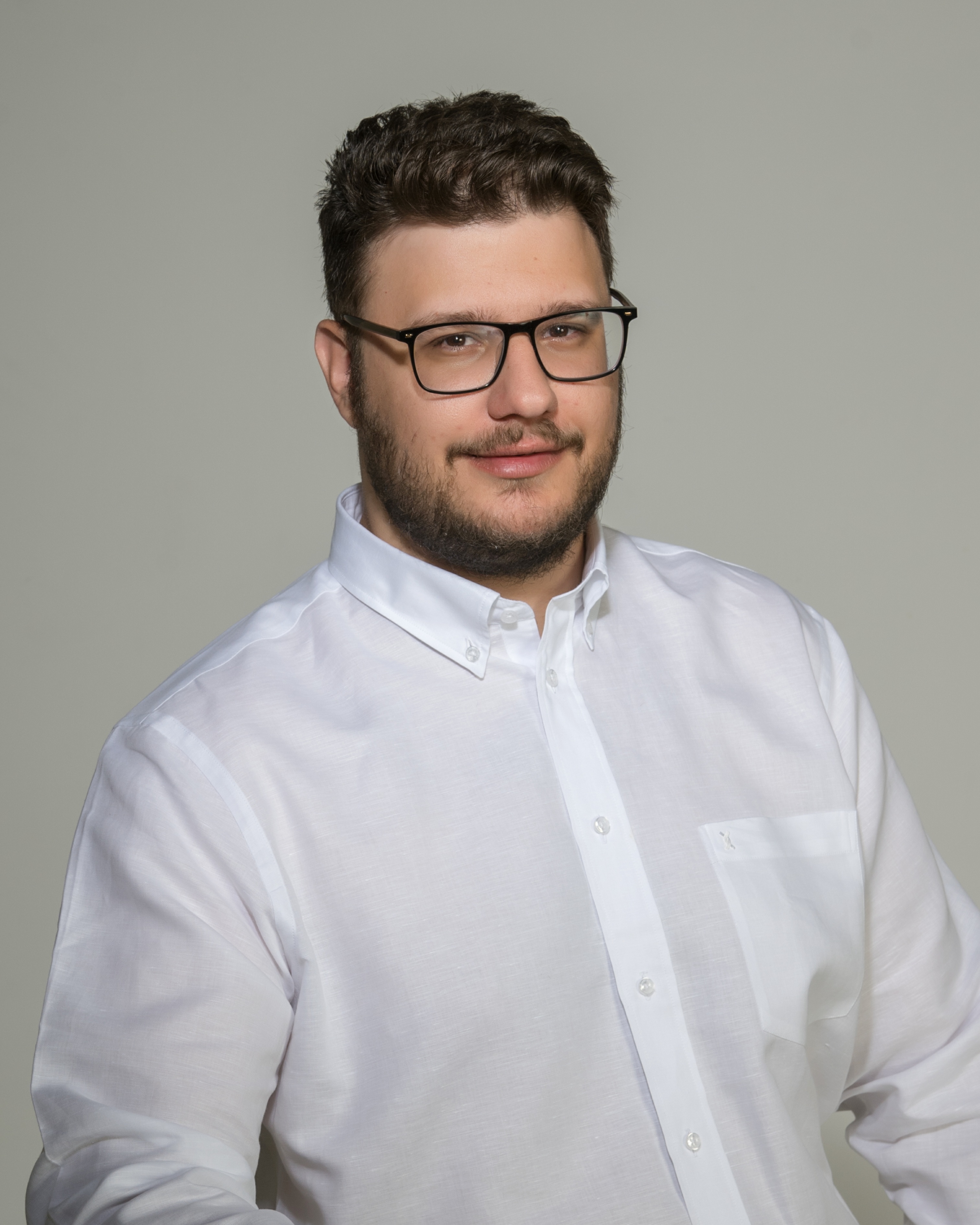}}]{Alexandros-Apostolos A. Boulogeorgos} (S'11, M'16, SM'19) was born in Trikala, Greece in 
	1988. He received the Electrical and Computer Engineering (ECE) diploma degree and Ph.D. degree in Wireless Communications from the Aristotle University of Thessaloniki (AUTh) in 2012 and 2016, respectively. 
	
	From November 2012, he has been a member of the wireless communications division of AUTh, working as a research assistant/project engineer in various telecommunication and networks projects. During 2017, he joined the information technologies institute, while from November 2017, he has joined the Department of Digital Systems, University of Piraeus, where he conducts research  in the area of wireless communications. Moreover, from October 2012 until September 2016, he was a teaching assistant at the department of ECE of AUTh, whereas, from February 2017, he serves as an adjunct professor at the Department of ECE of the University of Western Macedonia and as an visiting lecturer at the Department of Computer Science and Biomedical Informatics of the University of Thessaly.  
	
	Dr. Boulogeorgos  has authored and co-authored more than 50 technical papers, which were published in scientific journals and presented at prestigious international conferences. Furthermore, he has submitted two (one national and one European) patents. Likewise, he has been involved as member of Technical Program Committees in several IEEE and non-IEEE conferences and served as a reviewer in various IEEE journals and conferences. Dr. Boulogeorgos was awarded with the ``Distinction Scholarship Award'' of the Research Committee of AUTh for the year 2014 and was recognized as an exemplary reviewer for IEEE Communication Letters for 2016 (top $3\%$ of reviewers). Moreover, he was named a top peer reviewer (top $1\%$ of reviewers) in Cross-Field and Computer Science in the Global Peer Review Awards 2019, which was presented by the Web of Science and Publons.  His current research interests spans in the area of wireless communications and networks with emphasis in high frequency communications, optical wireless communications and communications for biomedical applications. He is a \textit{Senior Member} of the IEEE and a member of the Technical Chamber of Greece. He is currently an  Editor for \textit{IEEE Communications Letters}, and an Associate Editor for the \textit{Frontier In Communications And Networks}.
\end{IEEEbiography}

\begin{IEEEbiography}[{\includegraphics[width=1in,height=1.25in,clip,keepaspectratio]{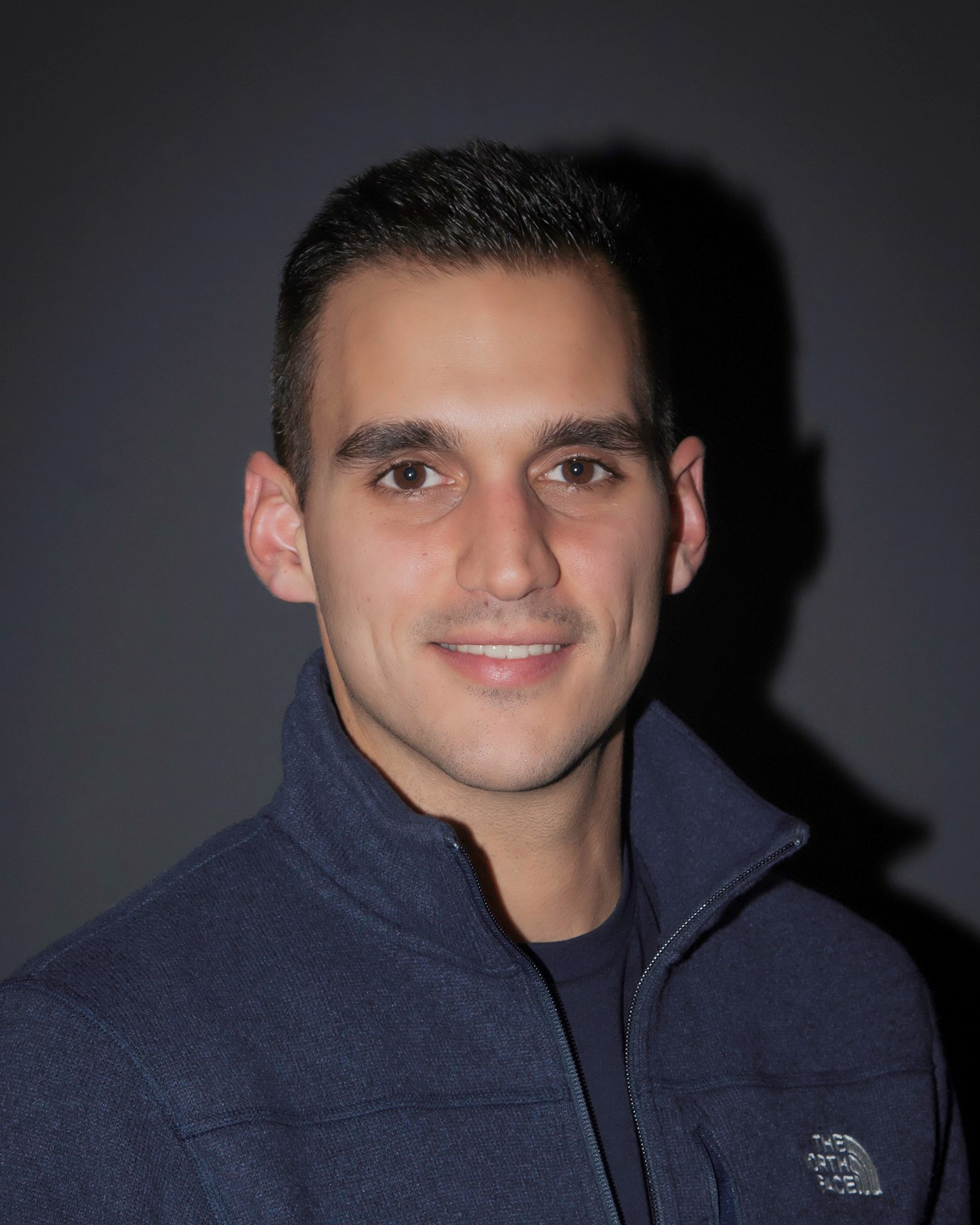}}]{Stylianos E. Trevlakis} was born in Thessaloniki, Greece in 1991. He received the Electrical and Computer Engineering (ECE) diploma (5 year) from the Aristotle University of Thessaloniki (AUTh) in 2016. Afterwards he served in the Hellenic Army in for nine months in the Research Office as well as at the Office of Research and Informatics of the School of Management and Officers. During 2017, he joined the Information Technologies Institute, while from October 2017, he is pursuing his PhD at the department of ECE of AUTh. Also, he is a member of the Wireless Communications \& Information Processing (WCIP) group. In 2018, he was a visitor researcher at the department of Electrical and Computer Engineering at Khalifa University, Abu Dhabi, UAE.
	
His research interests are in the area of Wireless Communications, with emphasis on Optical Wireless Communications, and Communications \& Signal Processing for Biomedical Engineering.  
		
\end{IEEEbiography}

\begin{IEEEbiography}[{\includegraphics[width=1in,height=1.25in,clip,keepaspectratio]{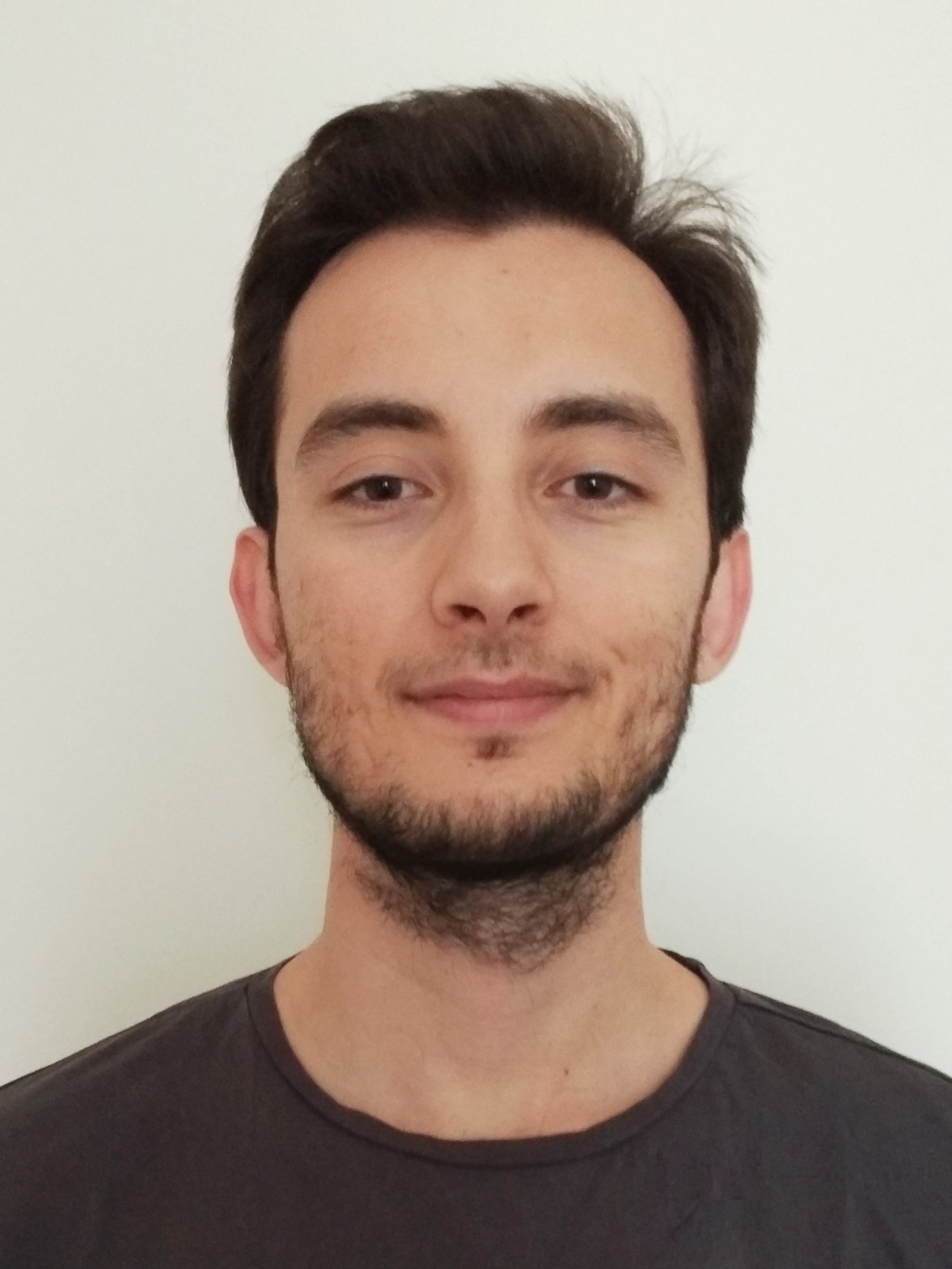}}]{Sotiris A. Tegos } was born in Serres, Greece. He received the Diploma Degree (5 years) in Electrical and Computer Engineering from the Aristotle University of Thessaloniki (AUTH), Greece, in 2017, where he is currently pursuing his PhD with the Department of Electrical and Computer Engineering. Also, he is a member of the Wireless Communications \& Information Processing (WCIP) group. In 2018, he was a visitor researcher at the department of Electrical and Computer Engineering at Khalifa University, Abu Dhabi, UAE. His current research interests include resource allocation in wireless communications, wireless power transfer, optimization theory and applications, and probability theory. He was an exemplary reviewer in IEEE Communication Letters for 2019 (top $3\%$ of reviewers).

\end{IEEEbiography}

\begin{IEEEbiography}[{\includegraphics[width=1in,height=1.25in,clip,keepaspectratio]{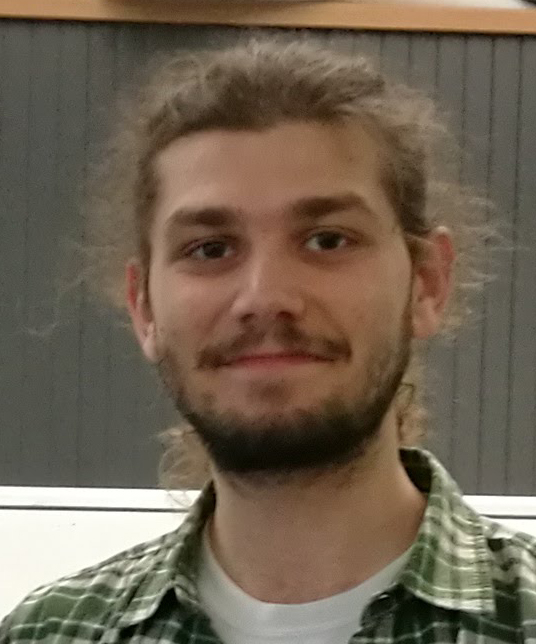}}]{Vasilis K. Papanikolaou} was born in Kavala, Greece in 1995. He received the Diploma Degree (5 years) in Electrical and Computer Engineering from the Aristotle University of Thessaloniki (AUTH), Greece, in 2018, where is currently pursuing his PhD with the Department of Electrical and Computer Engineering. Also, he is a member of the Wireless Communications \& Information Processing (WCIP) group. He was a visitor researcher at Lancaster University, UK and at Khalifa University, Abu Dhabi, UAE. In 2018, he received the IEEE Student Travel Grant Award for IEEE WCNC 2018. His research interests include visible light communications (VLC), non-orthogonal multiple access (NOMA), optimization theory, and game theory. He has served as a reviewer in various IEEE journals and conferences and he was an exemplary reviewer in IEEE Communication Letters for 2019 (top $3\%$ of reviewers).
	
\end{IEEEbiography}

\begin{IEEEbiography}
	[{\includegraphics[width=1in,height=1.25in,clip,keepaspectratio]{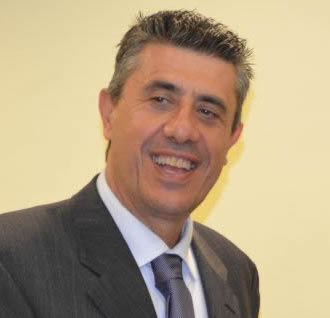}}]
	{George K. Karagiannidis} (M'96-SM'03-F'14) was born in Pithagorion, Samos Island, Greece. He received the University Diploma (5 years) and PhD degree, both in electrical and computer engineering from the University of Patras, in 1987 and 1999, respectively. From 2000 to 2004, he was a Senior Researcher at the Institute for Space Applications and Remote Sensing, National Observatory of Athens, Greece. In June 2004, he joined the faculty of Aristotle University of Thessaloniki, Greece where he is currently Professor in the Electrical \& Computer Engineering Dept. and Head of Wireless Communications \& Information Processing Systems Group (WCIP).  He is also Honorary Professor at South West Jiaotong University, Chengdu, China.
	
	His research interests are in the broad area of Digital Communications Systems and Signal processing, with emphasis on Wireless Communications, Optical Wireless Communications, Wireless Power Transfer and Applications and Communications \& Signal Processing for Biomedical Engineering.
	
	Dr. Karagiannidis has been involved as General Chair, Technical Program Chair and member of Technical Program Committees in several IEEE and non-IEEE conferences. In the past, he was Editor in several IEEE journals and from 2012 to 2015 he was the Editor-in Chief of IEEE Communications Letters. Currently, he serves as Associate Editor-in Chief of IEEE Open Journal of Communications Society.
	
	Dr. Karagiannidis is one of the highly-cited authors across all areas of Electrical Engineering, recognized from Clarivate Analytics as Web-of-Science Highly-Cited Researcher in the five consecutive years 2015-2019.
	
\end{IEEEbiography}

\end{document}